\documentclass[twoside,11pt]{article}

\usepackage[nohyperref,preprint]{jmlr2e}

\usepackage[utf8]{inputenc}
\usepackage{nicefrac}
\usepackage{array}
\usepackage{microtype}
\usepackage{graphicx}
\usepackage[aboveskip=2pt]{subcaption}
\usepackage{amsmath,amsfonts,amssymb,mathrsfs,bm}
\usepackage{mathtools}
\usepackage{comment}
\usepackage{pbox}

\usepackage[breaklinks,colorlinks,citecolor=black,linkcolor=black,urlcolor=black]{hyperref}
\usepackage[noabbrev,capitalize,nameinlink]{cleveref}
\usepackage{booktabs}
\usepackage{tabularx}
\usepackage{xcolor}
\usepackage{url}
\usepackage{colortbl}
\usepackage{tcolorbox}

\newcommand{\N}{\mathrm{N}}
\newcommand{\vb}{\mathbf{b}}
\newcommand{\vg}{\mathbf{g}}

\newcommand{\vf}{\mathbf{f}}
\newcommand{\vm}{\mathbf{m}}
\newcommand{\vs}{\mathbf{s}}
\newcommand{\vq}{\mathbf{q}}
\newcommand{\vr}{\mathbf{r}}
\newcommand{\vu}{\mathbf{u}}

\newcommand{\vx}{\mathbf{x}}
\newcommand{\vy}{\mathbf{y}}
\newcommand{\vlambda}{\bm{\lambda}}
\newcommand{\vmu}{\bm{\mu}}
\newcommand{\vnu}{\bm{\nu}}
\newcommand{\vtheta}{\bm{\theta}}
\newcommand{\veta}{\bm{\eta}}

\newcommand{\vomega}{\bm{\omega}}
\newcommand{\vbeta}{\bm{\beta}}

\newcommand{\vpsi}{\bm{\psi}}

\newcommand{\MA}{\mathbf{A}}
\newcommand{\MB}{\mathbf{B}}
\newcommand{\MC}{\mathbf{C}}
\newcommand{\MD}{\mathbf{D}}
\newcommand{\MF}{\mathbf{F}}
\newcommand{\MG}{\mathbf{G}}
\newcommand{\MH}{\mathbf{H}}
\newcommand{\MI}{\mathbf{I}}
\newcommand{\MJ}{\mathbf{J}}
\newcommand{\MK}{\mathbf{K}}
\newcommand{\ML}{\mathbf{L}}
\newcommand{\MP}{\mathbf{P}}
\newcommand{\MQ}{\mathbf{Q}}
\newcommand{\MR}{\mathbf{R}}
\newcommand{\MS}{\mathbf{S}}
\newcommand{\MV}{\mathbf{V}}
\newcommand{\MW}{\mathbf{W}}
\newcommand{\MX}{\mathbf{X}}
\newcommand{\MY}{\mathbf{Y}}
\newcommand{\MZ}{\mathbf{Z}}
\newcommand{\MOmega}{\bm{\Omega}}
\newcommand{\MSigma}{\bm{\Sigma}}
\newcommand{\MPhi}{\bm{\Phi}}
\newcommand{\T}{\top}

\newcommand{\R}{\mathbb{R}}
\newcommand{\BO}{\mathcal{O}}
\newcommand{\GP}{\text{GP}}

\newcommand{\overbar}[1]{\mkern 1.5mu\overline{\mkern-1.5mu#1\mkern-1.5mu}\mkern 1.5mu}

\newcommand{\vfunc}{\mathbf{f}}
\newcommand{\MFunc}{\mathbf{F}}
\newcommand{\priormean}{\vmu}
\newcommand{\priorcov}{\MK}
\newcommand{\priornatone}{\vlambda_\text{prior}^{(1)}}
\newcommand{\priornattwo}{\vlambda_\text{prior}^{(2)}}
\newcommand{\postmean}{\vm}
\newcommand{\postcov}{\MC}
\newcommand{\postnatone}{\vlambda^{(1)}}
\newcommand{\postnattwo}{\vlambda^{(2)}}
\newcommand{\likmean}{\overbar{\vm}}
\newcommand{\likcov}{\overbar{\MC}}
\newcommand{\liknatone}{\overbar{\vlambda}^{(1)}}
\newcommand{\liknattwo}{\overbar{\vlambda}^{(2)}}
\newcommand{\natparams}{\vlambda}
\newcommand{\meanparams}{\vomega}
\newcommand{\jacobian}{\MJ}
\newcommand{\hessian}{\MH}
\newcommand{\cavmean}{\postmean^{\cav}_n}
\newcommand{\cavcov}{\postcov^{\cav}_{n,n}}
\newcommand{\cavmeanfull}{\postmean^{\cav}}
\newcommand{\cavcovfull}{\postcov^{\cav}}

\newcommand{\fdim}{D}
\newcommand{\ydim}{D_\vy}

\newcommand{\state}{\bar{\mathbf{f}}}

\usepackage{relsize}
\newcommand{\cav}{\mathsmaller{\backslash}\!}

\newcommand{\LL}{\mathcal{L}}

\newcommand{\LLbar}{\overbar{\mathcal{L}}}

\renewcommand{\mid}{\,|\,}

\usepackage{xspace}
\newcommand{\eg}{\textit{e.g.}\xspace}
\newcommand{\ie}{\textit{i.e.}\xspace}

\newcommand{\E}{\mathbb{E}}
\DeclareMathOperator{\Cov}{Cov}

\DeclarePairedDelimiterX{\infdivx}[2]{\big[}{\big]}{%
	#1\;\delimsize\|\;#2
}
\newcommand{\KL}{\mathrm{D}_\mathrm{KL}\infdivx}
\newcommand{\DivAlpha}{\mathrm{D}_\alpha\infdivx}

\DeclarePairedDelimiterX{\infdivxbig}[2]{\bigg[}{\bigg]}{%
	#1\;\big \|\;#2
}
\newcommand{\KLbig}{\mathrm{D}_\mathrm{KL}\infdivxbig}

\usepackage{xcolor}

\newlength\figureheight
\newlength\figurewidth

\usepackage{tikz,pgfplots}
\usetikzlibrary{plotmarks,shapes,arrows}
\pgfplotsset{compat=newest} 

\pgfplotsset{/pgf/number format/.cd, 1000 sep={}}
\pgfkeys{/pgf/number format/.cd,1000 sep={}}

\pgfplotsset{every axis/.append style={
		grid style={line width=0.6pt,dotted,gray}}}

\pgfplotsset{every axis/.append style={
		legend style={inner xsep=1pt, inner ysep=0.5pt, nodes={inner sep=1pt, text depth=0.1em},draw=none}
}}

\usepackage{color}

\usepackage{algorithm}
\usepackage{algorithmic}

\usepgfplotslibrary{groupplots}

\newcommand\inputpgf[2]{{
		\let\pgfimageWithoutPath\pgfimage
		\renewcommand{\pgfimage}[2][]{\pgfimageWithoutPath[##1]{#1/##2}}
		\input{#1/#2}
}}

\pgfplotsset{ignore legend/.style={every axis legend/.code={\let\addlegendentry\relax}}}

\usepackage{lastpage}
\jmlrheading{1}{2021}{1-\pageref{LastPage}}{4/00}{10/00}{meila00a}{William Wilkinson and Simo S\"arkk\"a and Arno Solin}

\ShortHeadings{Bayes--Newton Methods}{Wilkinson, S\"arkk\"a and Solin}
\firstpageno{1}

\begin{document}

\title{Bayes--Newton Methods for Approximate Bayesian Inference \\ with PSD Guarantees}

\author{\name William J.\ Wilkinson \email william.wilkinson@aalto.fi \\
       \addr Department of Computer Science\\
       Aalto University\\
       Finland
       \AND
       \name Simo S\"arkk\"a \email simo.sarkka@aalto.fi \\
       \addr Department of Electrical Engineering and Automation\\
       Aalto University\\
       Finland
	   \AND
	   \name Arno Solin \email arno.solin@aalto.fi \\
	   \addr Department of Computer Science\\
	   Aalto University\\
	   Finland}

\editor{Pierre Alquier}

\maketitle

\begin{abstract}
We formulate natural gradient variational inference (VI), expectation propagation (EP), and posterior linearisation (PL) as generalisations of Newton's method for optimising the parameters of a Bayesian posterior distribution. This viewpoint explicitly casts inference algorithms under the framework of numerical optimisation. We show that common approximations to Newton's method from the optimisation literature, namely Gauss--Newton and quasi-Newton methods (\eg, the BFGS algorithm), are still valid under this `Bayes--Newton' framework. This leads to a suite of novel algorithms which are guaranteed to result in positive semi-definite (PSD) covariance matrices, unlike standard VI and EP. Our unifying viewpoint provides new insights into the connections between various inference schemes. All the presented methods apply to any model with a Gaussian prior and non-conjugate likelihood, which we demonstrate with (sparse) Gaussian processes and state space models.
\end{abstract}


\section{Introduction}\label{sec:intro}

When performing approximate Bayesian inference in probabilistic models, the need to strike a balance between accuracy and efficiency under varying use cases has led to the development of numerous schemes. Typically these have been derived from different viewpoints, and from various communities of researchers with different priorities. For example, linearisation-based methods have been intensively studied in the signal processing literature due to their intuitive nature when applied to nonlinear dynamical systems \citep{bell1994iterated, sarkka2013bayesian, garcia2016iterated}. Attempts to generalise these methods beyond signal processing motivated the invention of expectation propagation \citep[EP,][]{minka2001family} in the machine learning community as an alternative to variational inference \citep[VI,][]{sato2001online, blei2017variational}. However, the Laplace approximation \citep{tierney1986accurate} arguably remains the most popular approach for performing inference in probabilistic machine learning models due to its simplicity.

Despite their differing backgrounds, we will show here that all of these schemes can be viewed under the framework of numerical optimisation \citep{nocedal2006numerical}, namely as generalisations of Newton's method. Explicitly, we show that they all reduce to either Newton's method or the Gauss--Newton method under certain conditions. By making these links to the optimisation literature explicit, we gain new insights into the type of approximations used and the connections between the methods. For example, we show that natural gradient VI is a limiting case of power EP, we discuss the connection between variational inference and Newton's method, and we show that when approximations are applied to Newton's method the extended Kalman smoother is recovered. We also derive an improved version of the posterior linearisation algorithm \citep{garcia2016iterated} based on our insights.

Furthermore, we show that our optimisation viewpoint provides the means by which to derive new variants of VI and EP by showing that Gauss--Newton \citep{bjorck1996numerical} and quasi-Newton \citep{broyden1967quasi} approximations remain valid in these cases. These methods address stability issues by ensuring that updates to the approximate posterior always result in positive semi-definite (PSD) covariance matrices. Such stability issues have previously hindered the use of approximate inference in cases where the likelihood is not log-concave \citep{challis2013gaussian}. We also present an alternative approach to PSD constraints based on Riemannian gradients \citep{lin2020handling}.

Finally, we demonstrate that all the methods outlined here can be used to perform inference in Gaussian processes \citep{rasmussen2003gaussian} and their variants, as well as state space models. Our experiments show that for some complicated non-conjugate models our methods can significantly improve prediction accuracy relative to a simple heuristic approach.

Our main contributions, which apply generally to cases where the approximate posterior is chosen to be Gaussian, can be summarised as follows:
\begin{itemize}
\item We present natural gradient variational inference, power expectation propagation, and posterior linearisation under a unified `Bayes--Newton' framework based on numerical optimisation. We argue that this presentation makes the connections between methods more explicit than in previous work.
\item We utilise this framework to derive novel Gauss--Newton methods for approximate inference, including a Gauss--Newton approximation to variational inference which guarantees the posterior covariance is PSD.
\item We derive a (damped) quasi-Newton method for approximate inference based on the application of BFGS updates to local likelihood terms. This leads to novel quasi-Newton approximations for variational inference, expectation propagation, and posterior linearisation, all of which guarantee the posterior covariance is PSD.
\item We discuss PSD constraints for variational inference based on Riemannian gradients, and show that similar constraints can be applied to expectation propagation and posterior linearisation. We also discuss heuristic methods for ensuring PSD covariances, and present case studies comparing the proposed methods.
\end{itemize}

\section{Background}\label{sec:background}

Despite their varying motivations and attributes, there has been much work pointing out the connections between different approximate inference schemes. In this section, we review this past work and set out the notation to be used throughout this paper to unify the discussed methods.

\subsection{Related Work} \label{sec:related}

Casting approximate inference as optimisation is a viewpoint explicitly used when performing variational inference \citep{blei2017variational}, in which gradient ascent is typically applied to a lower bound of the model likelihood (see \cref{sec:vi}). The derivation of \emph{natural gradient ascent} for VI \citep{amari1998natural, khan2017conjugate} has made clear further connections to optimisation methods that incorporate second-order information, namely Newton's method. Notably, \citet{khan2021bayesian} show how VI generalises a large class of machine learning algorithms, including Newton's method, the Laplace approximation, and many gradient-based optimisation algorithms such as the Adam optimiser \citep{kingma2014adam, khan2018fast}.

The VI formulation of inference in \citet{khan2021bayesian} is extremely general, so much so that they refer to it as the `Bayesian Learning Rule', but it ignores the various other approaches to approximate inference, namely linearisation-based methods and EP. This is likely because it less clear how such approaches can be cast as optimisation. However, it was shown by \citet{bell1994iterated} that the iterated extended Kalman smoother (EKS) is equivalent to applying the Gauss--Newton algorithm to a nonlinear state space model. \citet{garcia2016iterated} further discuss how posterior linearisation, which is a generalisation of all nonlinear Kalman smoothers, can also be seen as a Gauss--Newton type of method.

Viewing EP as a form of gradient-based optimisation is less straight-forward, but the connections between EP and VI have been discussed at length: VI is in fact a special case of Power EP \citep[PEP,][]{minka2004power, minka2005divergence}, and this relationship remains valid in settings such as sparse Gaussian process models \citep{bui2017unifying} and when using natural gradients \citep{bui2018partitioned}. By considering EP in the large-data limit, \citet{dehaene2018expectation} show that EP does have a deep connection to Newton's method. However, our work shows that these connections are more general than this limiting case (see \cref{sec:pep}).

\citet{nickisch2008approximations} provide a presentation of some of the methods discussed here with a unifying aim but do not discuss linearisation-based methods and their viewpoint does not allow for the derivation of Gauss--Newton and quasi-Newton extensions. \citet{jylanki2011robust} similarly discuss a variety of approaches and explore ways to deal with the instability of inference, particularly EP, when using non-log-concave likelihoods which result in non-PSD covariance updates. They suggest a complicated scheme involving double-loop algorithms, {\it ad hoc} fixes, and adaptive learning rates.

Applying PSD constraints in EP has long been an open research question, with most methods being based on similar double-loop algorithms \citep{opper2005expectation, seeger2011fast}. \citet{lin2020handling} present a method for enforcing PSD constraints in VI based on Riemannian gradients, and we discuss this approach in \cref{sec:psd-constraints}, where we also derive a similar approach for EP. On the other hand, PL and its variants are guaranteed to result in PSD covariance matrices, but are often less accurate than VI and EP since the PL parameter updates are based on first-order derivative information only (see \cref{sec:gauss-newton} for discussion).

We also explore the use of quasi-Newton algorithms for inference. The quasi-Newton approach has been used previously to improve the computational scaling of Gaussian process regression \citep{leithead2007n}. However, as discussed in \cref{sec:quasi-newton}, the application of low-rank updates to a large full-rank covariance matrix can be a very poor approximation, and hence our proposed method is instead based on updates to local terms, with a focus on accurate inference and PSD guarantees. It is also important to distinguish our work from Bayesian interpretations of quasi-Newton methods \citep{hennig2013quasi, hennig2015probabilistic}, which aim to characterise uncertainty about the optimisation procedure itself, rather than use the quasi-Newton algorithm for Bayesian inference as we do.

In this paper, we consider models with a Gaussian prior and non-Gaussian observation model, and all of our case studies in \cref{sec:examples} are based on Gaussian processes \citep{rasmussen2003gaussian}. Approximate inference has been intensively studied for such models, and we outline all the connections between our work and this body of literature in \cref{sec:gp}, \cref{sec:sparse-gp}, and \cref{sec:markov-gp}, where we discuss Gaussian processes, sparse Gaussian processes, and state space models respectively. \citet{challis2013gaussian} list many other modelling scenarios that fit our specification, including Bayesian generalised linear models, binary logistic regression, and independent component analysis. Many of the presented methods could also be extended beyond the Gaussian case to any exponential family distribution, but we do not discuss such extensions here.

\subsection{Model Definition and Notation} \label{sec:notation}

We consider models with a matrix-valued latent variable, $\MFunc\in\R^{N\times \fdim}$, and observed data, $\MY\in\R^{N\times \ydim}$. We use their vectorised form, letting \mbox{$\vfunc=\mathrm{vec}(\MFunc)\in\R^{N\fdim\times 1}$}, \mbox{$\vy=\mathrm{vec}(\MY)\in\R^{N\ydim\times 1}$}, and at all times we abuse notation by indexing them as follows: $\vfunc_n=\MFunc_n^\T\in\R^{\fdim\times 1}$, and $\vy_n=\MY_n^\T\in\R^{\ydim\times 1}$. For a block-diagonal matrix $\likcov$, we also refer to its $n$-th block as $\likcov_{n,n}$. This allows us to use vector notation throughout, whilst recognising that all methods are extendable to matrix-valued data (see \cref{sec:experiments} for an example).

We assume a Gaussian prior for $\vfunc$ with a non-conjugate, \ie, non-Gaussian, observation model for $\vy$ (which we will refer to as the \emph{likelihood}) that factorises as follows,
\begin{gather}\label{eq:model}
	\begin{aligned}
		\vfunc &\sim p(\vfunc) = \N(\vfunc \mid \priormean, \priorcov), & & \text{(prior)} \\
		\vy \mid \vfunc &\sim p(\vy \mid \vfunc) = \prod_{n=1}^N p(\vy_n\mid \vfunc_n). \qquad & & \text{(likelihood)}
	\end{aligned}
\end{gather}
Our main motivation for such a model is Gaussian processes (see \cref{sec:gp}), where the prior is constructed using a \emph{mean function}, $\mu(\cdot)$, and a \emph{kernel}, $\kappa(\cdot,\cdot)$, applied to some input features, $\MX$: $\priormean=\mu(\MX)$, $\priorcov=\kappa(\MX,\MX)$. However, as discussed in \cref{sec:related}, our methods are not limited to this case, so we maintain a more general presentation.

We explore methods for computing a Gaussian approximation, $q(\vfunc)=\N(\vfunc \mid \postmean, \postcov)$, to the non-Gaussian posterior, $p(\vfunc \mid \vy)$:
\begin{equation} \label{eq:approx-post}
q(\vfunc) \approx p(\vfunc\mid\vy) \propto p(\vfunc) \prod_{n=1}^N p(\vy_n\mid \vfunc_n).
\end{equation}
\emph{Without loss of generality} we can assume that $q(\vfunc)$ factorises in the same way as the true posterior by approximating the non-Gaussian likelihood with a factorisable \emph{unnormalised} Gaussian function, $t(\vfunc) = \prod_n t(\vfunc_n)$, resulting in an approximate posterior of the form,
\begin{equation} \label{eq:fact-post}
q(\vfunc) \propto p(\vfunc) \prod_{n=1}^N t(\vfunc_n).
\end{equation}
We emphasise that this parametrisation is not a restriction or limitation on the approximate posterior since, as we will show, updates to $q(\vfunc)$ (\eg, via gradient-based methods) always implicitly contain a contribution from the prior and a factorised contribution from the true likelihood. That is to say, the approximate posterior is \emph{fully characterised} by the prior and an approximate likelihood. We denote $\likmean$ and $\likcov$ as the mean and covariance of $t(\vfunc)$ respectively, $t(\vfunc) = z\N(\vfunc \mid \likmean, \likcov )$, where $\likcov\in\R^{N\fdim\times N\fdim}$ is a block-diagonal matrix with block size $\fdim$, and $z=\prod_{n=1}^N z_n$ is the unnormalised Gaussian constant. The above construction is very general, and we will show that it directly enables all approximate inference methods to be cast as local parameter update rules for $\likmean$ and the block-diagonal elements of $\likcov$.

Letting $\natparams=[\postnatone \in \R^{N\fdim \times 1}, \, \postnattwo \in \R^{N\fdim \times N\fdim}]$ be the \emph{natural parameters}, we parametrise the model densities as follows,
\begin{gather}\label{eq:notation}
	\begin{aligned}
		&\text{Prior:} && p(\vfunc) = \N(\vfunc \mid \priormean, \priorcov), && \quad \priornatone = \priorcov^{-1}\priormean, && \priornattwo = -\frac{1}{2} \priorcov^{-1}, \\
		&\text{Approximate likelihood:} && t(\vfunc) = \N(\vfunc \mid \likmean, \likcov), && \quad \liknatone = \likcov^{-1}\likmean, && \liknattwo = -\frac{1}{2} \likcov^{-1}, \\
		&\text{Approximate posterior:} && q(\vfunc) = \N(\vfunc \mid \postmean, \postcov), && \quad \postnatone = \postcov^{-1}\postmean, && \postnattwo = -\frac{1}{2} \postcov^{-1}. \\
	\end{aligned}
\end{gather}
Due to conjugacy, we have $\postnatone=\priornatone + \liknatone$ and $\postnattwo=\priornattwo + \liknattwo$.

We will now introduce our unifying perspective on approximate inference. It is important to note that in \cref{sec:Newton} and \cref{sec:bayes-newton} we do not propose any new methods: we present a view of \emph{existing} algorithms under a numerical optimisation framework. Later, in \cref{sec:bayes-gauss-newton}, \cref{sec:bayes-quasi-newton}, and \cref{sec:psd-constraints}, we derive entirely \emph{novel} algorithms motivated by this viewpoint.

\section{Newton's Method and the Laplace Approximation as Bayesian Inference} \label{sec:Newton}

Newton's method \citep{nocedal2006numerical} is a very general approach for finding the optimum of a function, or the mode of a distribution. It can be used to perform {\it maximum a posteriori} (MAP) estimation in a Bayesian model by letting the optimisation target be the log-posterior, $\LL(\vfunc)=\log p(\vfunc\mid \vy)$. Using the model and notation from above we have,
\begin{gather}\label{eq:Laplace}
	\begin{aligned}
		\LL(\vfunc) &=\log p(\vfunc \mid \vy) = \log p(\vy \mid \vfunc) + \log p(\vfunc) - \log p(\vy) , \\
		\nabla_\vfunc \LL(\vfunc) &= \nabla_\vfunc \log p(\vy \mid \vfunc) - \priorcov^{-1} (\vfunc-\priormean) , \\
		\nabla_\vfunc^2 \LL(\vfunc) &= \nabla_\vfunc^2\log p(\vy\mid \vfunc) - \priorcov^{-1}. \\
	\end{aligned}
\end{gather}
We then iterate the following \emph{online} Newton updates (see \cref{app:newton-updates}),
\begin{gather}\label{eq:Newton}
\begin{aligned}
	\postcov_{k+1}^{-1} &= (1-\rho) \postcov_{k}^{-1} - \rho \, \nabla_\vfunc^2 \LL(\postmean_k) , \\
	\postmean_{k+1} &= \postmean_k +  \rho \, \postcov_{k+1} \, \nabla_\vfunc \LL(\postmean_k) , \\
\end{aligned}
\end{gather}
where $k$ is the iteration number. The term `online' refers to the fact that $\postcov_{k}$ is updated in a damped fashion using learning rate $\rho$. When $\rho=1$, this reduces to standard Newton's method. The iterates $\postmean_k$ converge to the fixed point, $\postmean^*$, the posterior mode. This MAP estimate can be transformed into a full approximate inference scheme by using the inverse Hessian of the objective, $(-\nabla_\vfunc^2\LL(\postmean^*))^{-1}$, as the posterior covariance estimate, which is known as the \emph{Laplace approximation} \citep{tierney1986accurate}. This is a natural choice since the iterates $\postcov_k$ converge to this quantity. The approximate posterior is then given by $q(\vfunc) = \N(\vfunc \mid \postmean=\postmean^*, \postcov= (-\nabla_\vfunc^2\LL(\postmean^*))^{-1})$.

Examining the form of the Hessian in \cref{eq:Laplace} we can see that the running estimate of $\postcov^{-1}$ contains additive contributions from the prior precision and a block-diagonal term depending on the likelihood: $\nabla_\vfunc^2 \LL(\vfunc)=\nabla_\vfunc^2\log p(\vy\mid \vfunc) - \MK^{-1} = \nabla_\vfunc^2\log p(\vy\mid \vfunc) + 2\priornattwo$. For notational convenience we define $\nabla_\vfunc \log p(\vy \mid \postmean_k) := \nabla_\vfunc \log p(\vy \mid \vfunc)|_{\vfunc=\postmean_{k}}$ to be the Jacobian w.r.t.\ $\vfunc$ evaluated at the posterior mean estimate. Utilising the property that $\postcov^{-1} = \MK^{-1} + \likcov^{-1}$, and similarly $\natparams = \natparams_\text{prior} + \overbar{\natparams}$, 
the Newton updates can be rewritten in terms of the natural parameters as
\begin{gather}\label{eq:Newton-natural}
	\begin{aligned}
		\postnattwo_{k+1} := -\frac{1}{2} \postcov_{k+1}^{-1} &= -(1-\rho)\frac{1}{2} \postcov_{k}^{-1} - \rho \, \frac{1}{2} \left(\priorcov^{-1} - \nabla_\vfunc^2\log p(\vy\mid \postmean_k) \right) \\
		&= \priornattwo + \underbrace{(1-\rho) \liknattwo_k + \rho \, \frac{1}{2}\nabla_\vfunc^2\log p(\vy\mid \postmean_k)}_{\liknattwo_{k+1}} \,, \\
		\postnatone_{k+1} := \postcov^{-1}_{k+1} \postmean_{k+1} &= \postcov^{-1}_{k+1} \postmean_k + \rho \, \nabla_\vfunc \log p(\vy \mid \postmean_k) - \rho \, \MK^{-1}(\postmean_k-\priormean) \\
		&= \priornatone + \underbrace{(1-\rho)\liknatone_k + \rho \, \big(\nabla_\vfunc \log p(\vy \mid \postmean_k) - \nabla_\vfunc^2\log p(\vy\mid \postmean_k) \, \postmean_k \big) }_{\liknatone_{k+1}} \,, \\
	\end{aligned}
\end{gather}
where we define $\postnatone_k$, $\postnattwo_k$ and $\liknatone_k$, $\liknattwo_k$ respectively to be the natural parameters of the approximate posterior and approximate likelihood at iteration $k$. A more detailed derivation is given in \cref{app:newton-updates}. Observing that the prior contribution in \cref{eq:Newton-natural} is fixed, with only the terms depending on the likelihood being updated across iterations, leads to the following remark:
\begin{remark}
	Only the block-diagonal entries of the posterior precision are updated when iterating Newton's method, since $\nabla_\vfunc^2\log p(\vy\mid \postmean_k)$ is block-diagonal and the prior $\priornattwo$ is fixed. Therefore Newton's method / the Laplace approximation can be written as a combination of local (likelihood) and global (posterior) updates.
\end{remark}
	
This fact, that Newton's method can be seen as iterative updates to a likelihood component, can be further clarified by defining the log likelihood to be a \emph{surrogate target} for optimisation. The combination of this with global conjugate updates ensures that the likelihood is always updated using the latest global information. We denote this surrogate target $\LLbar(\vfunc_n)$, and its Jacobian $\MJ_k$ and Hessian $\MH_k$ (which is a block-diagonal matrix) are, for all $n=1,2,\dots,N$,
\begin{gather} \label{eq:newton-target}
	\begin{aligned}
		\hspace{-1.2cm}
		\left.
		\begin{array}{ll}
			\LLbar(\vfunc_n) &= \log p(\vy_n \mid \vfunc_n)  \\
			\jacobian_{k,n} &= \nabla_{\vfunc_n} \LLbar(\postmean_{k,n})  \\
			\hessian_{k,n,n} &= \nabla_{\vfunc_n}^2\LLbar(\postmean_{k,n})  \\
		\end{array} \,\, \right\} \quad \parbox[]{15em}{surrogate target \& gradients \\ (Newton / Laplace)}
	\end{aligned}
\end{gather}
We then iterate the following updates,
\begin{gather} \label{eq:local-update}
	\begin{aligned}
		\left.
		\begin{array}{ll}
			\liknattwo_{k+1} &= (1-\rho) \, \liknattwo_k + \rho \, \frac{1}{2} \hessian_k   \\
			\liknatone_{k+1} &= (1-\rho) \, \liknatone_k + \rho \, \left( \jacobian_k -\hessian_k \, \postmean_k \right)  \\
		\end{array} \,\, \right\} \quad \parbox[]{10em}{local likelihood \\ online Newton update}
	\end{aligned}
\end{gather}
\vspace{-0.3cm}
\begin{gather} \label{eq:global-update}
	\begin{aligned}
		\hspace{-.1cm}
		\left.
		\begin{array}{ll}
		 \postcov_{k+1} &= -\frac{1}{2}(\priornattwo + \liknattwo_{k+1})^{-1}  \\
		\postmean_{k+1} &= \postcov_{k+1} (\priornatone + \liknatone_{k+1}) 
	\end{array}  \quad \hspace*{1.3cm} \right\} \quad \parbox[]{10em}{global posterior \\ update}
	\end{aligned}
\end{gather}
Since $\liknattwo_{k+1}$ is block-diagonal (the likelihood factorises) \cref{eq:local-update} is cheap to compute. That being said, the full algorithm has the same computational complexity as global updates because \cref{eq:global-update} involves inverting a dense matrix. \cref{eq:global-update} is equivalent to a conjugate regression step with the prior and the approximate likelihood. We notice that the Newton updates are now applied to a surrogate target, the log likelihood $\log p(\vy \mid \vfunc)$, rather than the log posterior. However, we have shown that these updates completely characterise the full Newton method, since the contribution from the prior is static.

Whilst \cref{eq:local-update,eq:global-update} seem more complicated than the standard updates in \cref{eq:Newton} (and have the same computational complexity), it turns out that this new form, in which Newton updates are applied to the local likelihoods before performing a conjugate update, will provide us with a unifying perspective that subsumes many approximate Bayesian inference algorithms. We now derive multiple such algorithms, showing how they all result in Newton-like updates of this form.

\section{Bayes--Newton: Approximate Bayesian Inference as Probabilistic Variants of Newton's Method}\label{sec:bayes-newton}

In this section we will present three prominent Bayesian inference methods: variational inference, power expectation propagation, and posterior linearisation. Our presentation will demonstrate how these methods can all be viewed as generalisations of Newton's method, with the parameter updates taking a surprising similar form to those presented in the previous section.

The methods presented here have an important distinction from Newton's method: the target $\LLbar(\cdot)$ is a function of not only the posterior mean, $\postmean$, but also the posterior covariance, $\postcov$, and involves computing expectations with respect to a probability distribution rather than using a single point estimate at the mean.

Due to the incorporation of the full Bayesian posterior into the updates we name this class of inference algorithms \emph{Bayes--Newton methods}. We will show that they too are completely characterised by a set of local likelihood parameter updates. In later sections we will show that common approximations to Newton's method from the optimisation literature (Gauss--Newton and quasi-Newton methods) are still valid under the Bayes--Newton framework.

\subsection{Variational Inference} \label{sec:vi}

Variational inference aims to minimise the KL divergence of the approximation $q(\vfunc)$ from the true posterior,
\begin{equation}
	q(\vfunc) = \mathrm{arg}\,\min_{q^*(\vfunc)}{\KL{q^*(\vfunc)}{p(\vfunc \mid \vy)}}.
\end{equation}
Doing so is equivalent to minimising the variational free energy (VFE), \ie, the negative evidence lower bound (ELBO),
\begin{align}
    \text{VFE}(q(\vfunc)) &= -\E_{q(\vfunc)}[ \log p(\vy \mid \vfunc)] + \KL{q(\vfunc)}{p(\vfunc)} \nonumber \\
		&= \E_{q(\vfunc)}[- \log p(\vy, \vfunc) + \log q(\vfunc) ], \label{eq:var-obj_}
\end{align}
with respect to the parameters of $q(\vfunc)$. It is highly desirable to apply \emph{natural gradient} descent \citep{amari1998natural} to the VFE to obtain the posterior natural parameters, $\natparams$. Following \citet{khan2017conjugate}, we use the property that the gradient with respect to the mean parameters, $\meanparams=[\postmean, \postcov + \postmean \postmean^\T]$, is equivalent to the natural gradient. This results in the following natural gradient update step,
\begin{align} \label{eq:natgrad-vfe}
		\natparams_{k+1} &= \natparams_k - \rho \, \nabla_{\meanparams} \text{VFE}(q(\vfunc)) \nonumber \\
		&= (1 - \rho)\natparams_k + \rho \, \nabla_{\meanparams}\E_{q(\vfunc)}[\log p(\vy, \vfunc)] .
\end{align}
By application of the chain rule to obtain \cref{eq:natgrad-vfe} in terms of gradients with respect to $\postmean$, the individual posterior parameter updates then become
\begin{gather}\label{eq:vi-natural}
	\begin{aligned}
		\postnattwo_{k+1} &= (1-\rho) \postnattwo_k + \rho \, \frac{1}{2} \nabla^2_{\vm} \E_{q(\vfunc)}[ \log p(\vy, \vfunc) ] \\
		&= \priornattwo + (1-\rho) \liknattwo_k + \rho \, \frac{1}{2} \nabla^2_{\vm} \E_{q(\vfunc)}[ \log p(\vy \mid  \vfunc) ] ,  \\
		\postnatone_{k+1} &= (1-\rho) \postnatone_k + \rho \left( \nabla_{\vm} \E_{q(\vfunc)}[ \log p(\vy , \vfunc)] - \nabla^2_{\vm} \E_{q(\vfunc)}[ \log p(\vy, \vfunc) ] \, \postmean_k \right) \\
		&= \priornatone + (1-\rho) \liknatone_k + \rho \left( \nabla_{\postmean} \E_{q(\vfunc)}[ \log p(\vy \mid \vfunc)] - \nabla^2_{\postmean} \E_{q(\vfunc)}[ \log p(\vy \mid \vfunc) ] \, \postmean_k \right) . \\
	\end{aligned}
\end{gather}
A more detailed derivation is given in \cref{app:VI}. The striking similarity between these updates and \cref{eq:Newton-natural} leads to the following remark:
\begin{remark}
	Natural gradient VI is fully characterised by updates to the local approximate likelihoods: only the block-diagonal of the posterior precision is updated iteratively since $\nabla^2_{\postmean} \E_{q(\vfunc)}[ \log p(\vy \mid \vfunc) ]$ is block-diagonal. Furthermore, the updates can be framed in a similar local/global fashion to Newton's method by defining the expected log likelihood, $\E_{q(\vfunc)}[ \log p(\vy \mid \vfunc) ]$, as the target, and calculating its gradients with respect to the posterior mean. 
\end{remark}
We conclude that natural gradient VI can be performed by defining:
\begin{gather} \label{eq:vi-target}
	\begin{aligned}
		\hspace{-1.2cm}
		\left.
		\begin{array}{ll}
			\LLbar(\postmean_n, \postcov_{n,n}) &= \E_{q(\vfunc_n)}[ \log p(\vy_n \mid \vfunc_n) ]  \\
			\jacobian_{k,n} &= \nabla_{\postmean_n} \LLbar(\postmean_{k,n}, \postcov_{k,n,n})  \\
			\hessian_{k,n,n} &= \nabla_{\postmean_n}^2\LLbar(\postmean_{k,n}, \postcov_{k,n,n})  \\
		\end{array} \,\, \right\} \quad \parbox[]{15em}{surrogate target \& gradients \\ (VI)}
	\end{aligned}
\end{gather}
and then iterating the local damped Newton updates, \cref{eq:local-update}, and the posterior updates given in \cref{eq:global-update}. This analysis shows that the approximate posterior does indeed factorise in the way given by \cref{eq:approx-post}, which is perhaps surprising since VI is not usually characterised this way (whereas EP, for example, explicitly uses this parameterisation).

\subsubsection{The Variational Free Energy} \label{sec:vfe}

Whilst the natural gradient updates above have a very convenient form, it is still often useful to be able to compute the variational free energy explicitly, for example when optimising the hyperparameters of a Gaussian process model (\cref{sec:gp}), or when monitoring convergence in line search methods. Due to our parametrisation of the approximate posterior, the VFE can be written,
\begin{align}
		\text{VFE}(q(\vfunc)) &= -\E_{q(\vfunc)} \left[ \log \frac{p(\vy \mid \vfunc) p(\vfunc)}{q(\vfunc)} \right] \nonumber \\
		&= -\E_{q(\vfunc)} \left[ \log \frac{p(\vy \mid \vfunc) \int p(\vfunc) t(\vfunc) \, \mathrm{d} \vfunc }{t(\vfunc)} \right] \nonumber \\
		&= -\E_{q(\vfunc)} \left[ \log \frac{p(\vy \mid \vfunc) \int p(\vfunc) \N(\vfunc \mid \likmean, \likcov) \, \mathrm{d} \vfunc }{\N(\vfunc \mid \likmean, \likcov)} \right] \nonumber \\
		&= -\sum_{n=1}^N \E_{q(\vfunc_n)} [\log p(\vy_n \mid \vfunc_n)] + \sum_{n=1}^N \E_{q(\vfunc_n)} [\log \N(\vfunc_n \mid \likmean_n, \likcov_{n,n})] - \log \mathcal{Z} ,  \label{eq:vfe}
\end{align}
where $\mathcal{Z}=\int p(\vfunc) \N(\vfunc \mid \likmean, \likcov) \, \mathrm{d}\vfunc = \N(\likmean \mid \priormean, \priorcov + \likcov)$. The first term in \cref{eq:vfe} can be computed via quadrature methods, whilst the remaining terms are both Gaussian and can be computed in closed form. Also recall that the approximate likelihood factors are unnormalised, $t(\vfunc_n) = z_n \N(\vfunc_n \mid \likmean_n, \likcov_{n,n})$, but that the constants $z_n$ cancel out in the VFE and therefore can be ignored.

We also propose using an approximation to the VFE as the energy associated with the Laplace/Newton method by replacing the expectations in \cref{eq:vfe} by point estimates at the posterior mean. We call this the Laplace energy (LE),
\begin{equation} \label{eq:laplace-energy}
	\text{LE}(q(\vfunc)) = -\sum_{n=1}^N \log p(\vy_n \mid \postmean_n) + \sum_{n=1}^N \log \N(\postmean_n \mid \likmean_n, \likcov_{n,n}) - \log \mathcal{Z} .
\end{equation}
Alternatively, \citet{rasmussen2003gaussian} propose an approximation to the negative log marginal likelihood based on a first-order Taylor expansion of the exact marginal likelihood evaluated at $\postmean$, giving,
\begin{gather}
	\begin{aligned} \label{eq:laplace-energy-gpml}
	\text{LE}_2(q(\vfunc)) &=  -\sum_{n=1}^N \log p(\vy_n \mid \postmean_n) + \frac{1}{2} \postmean^\T \priorcov^{-1} \postmean + \frac{1}{2} \log|\priorcov| + \frac{1}{2}\log|\priorcov^{-1}+\likcov^{-1}| .
	\end{aligned}
\end{gather}

\subsection{Power Expectation Propagation} \label{sec:pep}

In the previous sections we have seen how updates to the global posterior in both Newton's method and VI are completely characterised by iterative updates to local parameters. Whilst this is not the standard presentation for either of these schemes, we now turn our attention to a method which is typically (and necessarily) defined this way.

Power expectation propagation \citep[PEP,][]{minka2004power} aims to minimise the $\alpha$-divergence of the true posterior from its approximation,
\begin{equation}
	q(\vfunc) = \mathrm{arg}\,\min_{q^*(\vfunc)}{\DivAlpha{p(\vfunc \mid \vy)}{q^*(\vfunc)}}.
\end{equation}
In practice this is intractable, so instead the local approximate likelihood terms, $t(\vfunc_n)$, are updated iteratively. To update a single $t(\vfunc_n)$ the current term is removed from the posterior, and replaced with the true likelihood. This new quantity is termed the \emph{tilted distribution}, and PEP minimises the $\alpha$-divergence of the approximate posterior from the tilted distribution, which can be done by raising the likelihood terms to a power of $\alpha$ and minimising the forward KL divergence:
\begin{align}
		t(\vfunc_n) &= \mathrm{arg}\,\min_{t_*(\vfunc_n)} \KLbig{\frac{1}{Z_n}\frac{p^\alpha(\vy_n \mid \vfunc_n)}{t^\alpha(\vfunc_n)} q(\vfunc)} {\frac{1}{W_n}\frac{t_*^\alpha(\vfunc_n)}{t^\alpha(\vfunc_n)} q(\vfunc)} \nonumber \\
		&= \mathrm{arg}\,\min_{t_*(\vfunc_n)} \KLbig{\frac{1}{Z_n}\frac{p^\alpha(\vy_n \mid \vfunc_n)}{t^\alpha(\vfunc_n)} q(\vfunc_n)} {\frac{1}{W_n}\frac{t^\alpha_*(\vfunc_n)}{t^\alpha(\vfunc_n)} q(\vfunc_n)}, \label{eq:KL}
\end{align}
for $Z_n=\int \frac{p^\alpha(\vy_n \mid \vfunc_n)}{t^\alpha(\vfunc_n)} q(\vfunc) \mathrm{d}\vfunc=\int \frac{p^\alpha(\vy_n \mid \vfunc_n)}{t^\alpha(\vfunc_n)} q(\vfunc_n) \mathrm{d}\vfunc_n$ and $W_n=\int \frac{t^\alpha_*(\vfunc_n)}{t^\alpha(\vfunc_n)} q(\vfunc) \mathrm{d}\vfunc = \int \frac{t^\alpha_*(\vfunc_n)}{t^\alpha(\vfunc_n)} q(\vfunc_n) \mathrm{d}\vfunc_n$.

The above divergence can be minimised by choosing $t_*(\vfunc_n)$ such that the first two moments of the tilted distribution and the approximate posterior are matched, \ie, by computing the moments of both sides, setting them to be equal, and then solving for $t_*(\vfunc_n)$. In practice, a useful shortcut involves differentiating $\log Z_n$ with respect to the mean of the cavity, $q^{\cav}_n(\vfunc_n) = \N(\vfunc_n \mid \cavmean, \cavcov) \propto q(\vfunc_n)/t^\alpha(\vfunc_n)$ \citep{seeger2005expectation, rasmussen2003gaussian}. The derivatives of $\log Z_n$ turn out to be a function of the required moments, and so after some rearranging we obtain the following update algorithm for a single factor $t(\vfunc_n)$ (see \cref{app:EP} for the derivation),
\begin{align}
	&\MR_n = {\cavcovfull}_{n,n}^{-1} \left( \nabla_{\cavmean}^2\log \E_{q_{\cav n}(\vfunc_n)} [ p^\alpha(\vy_n \mid \vfunc_n) ] + {\cavcovfull}_{n,n}^{-1} \, \right)^{-1} , \nonumber \\
	&\liknattwo_{k+1,n,n} =  (1-\rho) \, \liknattwo_{k,n,n} + \rho \, \frac{1}{2\alpha} \MR_n \, \nabla_{\cavmean}^2 \log \E_{q_{\cav n}(\vfunc_n)} [ p^\alpha(\vy_n \mid \vfunc_n) ] , \nonumber \\
	&\liknatone_{k+1,n} \!= \! (1\!-\!\rho) \, \liknatone_{k,n} \!+ \!\rho \, \frac{1}{\alpha} \MR_n \!\left(\! \nabla_{\cavmean} \log \E_{q_{\cav n}(\vfunc_n)} [ p^\alpha(\vy_n \mid \vfunc_n) ] -\nabla_{\cavmean}^2 \log \E_{q_{\cav n}(\vfunc_n)} [ p^\alpha(\vy_n \mid \vfunc_n) ] \, \cavmean \!  \right)\! , \label{eq:pep-updates}
\end{align}
where we have \emph{damped} the updates \citep{jylanki2011robust} using learning rate $\rho$. Examining \cref{eq:pep-updates} leads to the following remark:

\begin{remark}
	The PEP updates for a single factor	take the form of damped Bayes--Newton updates to the approximate likelihood natural parameters, with two important distinctions: {\em (i)}~the Jacobian and Hessian are scaled by a factor $\MR_n$ to account for the fact that the target is a function of the cavity rather than the posterior, and {\em(ii)}~the updates act on the cavity mean rather than the posterior mean.
\end{remark}
We now define
\begin{gather} \label{eq:pep-target}
	\begin{aligned}
		\left.
		\begin{array}{ll}
			\LLbar(\cavmeanfull_n, \cavcovfull_{n,n}) &= \frac{1}{\alpha} \log \E_{q^{\cav }(\vfunc_n)} [ p^\alpha(\vy_n \mid \vfunc_n) ]  \\
			\MR_{k,n} &= {\cavcovfull}_{k,n,n}^{-1} \left( \alpha \nabla_{\cavmeanfull_n}^2\LLbar(\cavmeanfull_{k,n}, \cavcovfull_{k,n,n}) + {\cavcovfull}_{k,n,n}^{-1} \, \right)^{-1} \\
			\jacobian_{k,n} &= \MR_{k,n} \nabla_{\cavmeanfull_n} \LLbar(\cavmeanfull_{k,n}, \cavcovfull_{k,n,n})  \\
			\hessian_{k,n,n} &= \MR_{k,n} \nabla_{\cavmeanfull_n}^2\LLbar(\cavmeanfull_{k,n}, \cavcovfull_{k,n,n})  \\
		\end{array} \,\, \right\} \quad \parbox[]{15em}{surrogate target \\ \& gradients \\ (PEP)}
	\hspace{-1.2cm}
	\end{aligned}
\end{gather}
For notational convenience we can collect the marginal cavity means $\cavmean$ together: $\cavmeanfull =[\postmean^{\cav}_1,\dots,\postmean^{\cav}_N]^\T$, which allows us to apply the local updates of \cref{eq:local-update} by using $\cavmeanfull$ in place of the posterior mean, $\postmean$. \cref{eq:global-update} is then used to update the full posterior.

It is well known that when $\alpha \rightarrow 0$ the PEP energy becomes equivalent to the variational free energy \citep{minka2005divergence, bui2017unifying} and that in this case, if PEP converges, it converges to the same fixed points as VI. However, our presentation reveals even deeper connections between the two methods. In \cref{app:PEP-VI} we show that the surrogate target used in PEP has the following property,
\begin{equation} \label{eq:pep-limit}
	\lim_{\alpha\rightarrow 0} \frac{1}{\alpha} \log \E_{q^{\cav }(\vfunc_n)} [ p^\alpha(\vy_n \mid \vfunc_n) ] = \E_{q(\vfunc_n)} [ \log p(\vy_n \mid \vfunc_n) ] ,
\end{equation}
and furthermore we can see that the PEP scaling factor reverts to the identity in the limit, $\lim_{\alpha\rightarrow 0} \MR_n = \MI$. Combined with the fact that $\lim_{\alpha\rightarrow 0} q^{\cav}(\vfunc_n)=q(\vfunc_n)$, this shows that \cref{eq:pep-target} ($\alpha \rightarrow 0$) is identical to the VI updates given by \cref{eq:vi-target}, which leads to the following result:
\begin{remark}
	When $\alpha \rightarrow 0$, a single step of the power EP algorithm is equivalent to a natural gradient descent step in the variational free energy (\ie, a natural gradient VI step).
\end{remark}

This connection between PEP and natural gradient VI had gone largely unnoticed until \citet{bui2018partitioned} recently derived the same result. However, our presentation arguably makes the connection even clearer.

\subsubsection{The Power EP Energy}\label{sec:pep-energy}

The PEP algorithm also provides a way to compute an approximation to the negative log marginal likelihood. This approximation is often referred to as the PEP energy (PEPE) and to derive it we must recall that the approximate likelihood factors are defined as \emph{unnormalised} Gaussians, $t(\vfunc_n) = z_n \N(\vfunc_n \mid \likmean_n, \likcov_{n,n})$. The PEPE is then given by
\begin{align}
		\text{PEPE}(q(\vfunc)) &= -\log \int p(\vfunc) \prod_{n=1}^N t(\vfunc_n) \, \mathrm{d} \vfunc \nonumber \\
		&= -\log \int p(\vfunc) \prod_{n=1}^N  z_n \N(\vfunc_n \mid \likmean_n, \likcov_{n,n}) \, \mathrm{d} \vfunc \nonumber \\
		&= -\sum_{n=1}^N \log z_n - \log \mathcal{Z} , \label{eq:pep-energy-init}
\end{align}
where $\mathcal{Z}=\int p(\vfunc) \prod_{n=1}^N \N(\vfunc_n \mid \likmean_n, \likcov_{n,n}) \, \mathrm{d} \vfunc$, which is the same term used in the VFE in \cref{eq:vfe}. Computing the constants, $z_n$, is less straightforward. To do so, we match the zero-th moment of the tilted distribution in a similar way to the first two moments, which amounts to setting $z_n$ such that
\begin{equation} \label{eq:pep-z}
	z_n^\alpha \, \E_{q^{\cav }(\vfunc_n)} [ \N^\alpha(\vfunc_n \mid \likmean_n, \likcov_{n,n}) ] =  \E_{q^{\cav }(\vfunc_n)} [ p^\alpha(\vy_n \mid \vfunc_n) ] . \\
\end{equation}
Taking the logarithm and rearranging gives
\begin{equation} \label{eq:pep-z2}
	\log z_n = \frac{1}{\alpha} \left( \log \E_{q^{\cav }(\vfunc_n)} [ p^\alpha(\vy_n \mid \vfunc_n) ] - \log \E_{q^{\cav }(\vfunc_n)} [ \N^\alpha(\vfunc_n \mid \likmean_n, \likcov_{n,n}) ] \right) , \\
\end{equation}
such that the PEP energy becomes
\begin{equation} \label{eq:pep-energy}
	\text{PEPE}(q(\vfunc)) = -\frac{1}{\alpha} \sum_{n=1}^N \log \E_{q^{\cav }(\vfunc_n)} [ p^\alpha(\vy_n \mid \vfunc_n) ] + \frac{1}{\alpha} \sum_{n=1}^N \log \E_{q^{\cav }(\vfunc_n)} [ \N^\alpha(\vfunc_n \mid \likmean_n, \likcov_{n,n}) ]  - \log \mathcal{Z} .
\end{equation}
This presentation of the energy highlights its connection to the VFE in \cref{eq:vfe}: applying \cref{eq:pep-limit} to the first two terms immediately shows that as $\alpha \to 0$, $\text{PEPE} \to \text{VFE}$.

\subsection{Posterior Linearisation}

Posterior linearisation \citep[PL,][]{garcia2016iterated} is another approximate inference method which can be framed as updates to local likelihood factors. Whilst PL is not a commonly used method in the machine learning community, it provides an important link to the approximate Bayesian inference methods developed in the signal processing literature: it is an extension of classical statistical linearisation \citep{gelb1974applied}, which itself generalises the sigma-point smoothing algorithms such as the unscented Kalman smoother \citep{sarkka2013bayesian}. PL seeks an approximate posterior via the likelihood approximation $p(\vy \mid \vfunc) \approx q(\vy \mid \vfunc) = \N(\vy \mid \MA \vfunc + \vb, \MOmega)$, computed via statistical linear regression \citep[SLR,][]{sarkka2013bayesian} of $\E[\vy \mid \vfunc] := \E_{p(\vy \mid \vfunc)}[\vy]$ with respect to the approximate posterior $q(\vfunc)=\N(\vfunc \mid \postmean, \postcov)$. SLR provides the optimal linearisation of the likelihood model in the mean-square-error (MSE) sense,
\begin{gather} \label{eq:pl-mse}
	\begin{aligned}
		\text{MSE}(\MA, \vb) &= \E_{q(\vfunc)} \left[ (\E[\vy \mid \vfunc] - \MA \vfunc -\vb)^\T (\E[\vy \mid \vfunc] - \MA \vfunc -\vb) \right] .
	\end{aligned}
\end{gather}
The PL updates can be derived by setting the derivatives of \cref{eq:pl-mse} with respect to $\MA$, $\vb$ to zero, which gives
\begin{gather}
	\begin{aligned}
		\MA &= \E_{q(\vfunc)}\left[(\vfunc - \postmean)(\E[\vy\mid \vfunc] - \E_{q(\vfunc)}[\E[\vy\mid \vfunc]])^\T \right] \postcov^{-1} , \\
		\vb & = \E_{q(\vfunc)}[\E[\vy\mid \vfunc]] - \MA \postmean .
	\end{aligned}
\end{gather}
The likelihood covariance is then set equal to the mean-square-error \emph{matrix},
	\begin{align}
	\MOmega &= \E_{q(\vfunc),p(\vy \mid \vfunc)} \left[ (\vy - \MA \vfunc -\vb) (\vy- \MA \vfunc -\vb)^\T  \right] \nonumber \\
	&= \E_{q(\vfunc)} \left[ (\E[\vy \mid \vfunc] - \MA \vfunc -\vb) (\E[\vy \mid \vfunc] - \MA \vfunc -\vb)^\T + \Cov[\vy \mid \vfunc] \right] .
	\end{align}
This approach has an intuitive interpretation: PL seeks the best affine fit to the conditional expectation $\E[\vy \mid \vfunc]$ in the region of the approximate posterior, and sets the likelihood covariance equal to the error induced by this approximation. To enable direct comparison with VI and PEP, the approximate likelihood $q(\vy \mid \vfunc)$ can also be written as a Gaussian distribution over $\vfunc$: $t(\vfunc) = \N(\vfunc \mid \likmean, \likcov)$, with update rule (see \cref{app:PL} for the derivation):
\begin{gather} \label{eq:pl-target}
	\begin{aligned}
		\left.
		\begin{array}{ll}
			\LLbar(\postmean_n, &\hspace{-0.65em}\postcov_{n,n}) = \log \N(\vy_n \mid \E_{q(\vfunc_n)} [ \E[\vy_n\mid \vfunc_n] ], \, \MOmega_{n,n})   \\ 
			\jacobian_{k,n} &= {\nabla_{\postmean_n} \E_{q(\vfunc_n)} [ \E[\vy_n\mid \vfunc_n] ]}^\T \MOmega_{k,n,n}^{-1} (\vy_n - \E_{q(\vfunc_n)} [ \E[\vy_n\mid \vfunc_n] ]) \\
			\hessian_{k,n,n} &= -{\nabla_{\postmean_{n}} \E_{q(\vfunc_n)} [ \E[\vy_n\mid \vfunc_n] ]}^\T \MOmega_{k,n,n}^{-1} \nabla_{\postmean_n} \E_{q(\vfunc_n)} [ \E[\vy_n\mid \vfunc_n] ])  \\
		\end{array} \,\, \right\} \quad \parbox[]{15em}{surrogate target \\  \& gradients \\ (PL)}
	\end{aligned}
\end{gather}

The damped local/global Newton updates of \cref{eq:local-update} and \cref{eq:global-update} can now be applied to perform inference. Notice that PL only requires the Jacobian of the objective, whereas both EP and VI also require the Hessian, \ie, they utilise second-order derivative information of the objective. This first-order approximation to the Hessian means PL is a \emph{Gauss--Newton} method. We will discuss Gauss--Newton methods further in \cref{sec:bayes-gauss-newton}.

\subsubsection{Taylor Expansion / the Iterated Extended Kalman Smoother} \label{sec:eks}

Another approximate inference algorithm can be obtained by replacing statistical linear regression in PL by analytical linearisation, \ie, a first-order Taylor expansion (and using a simpler covariance approximation). This can be achieved by defining the simpler surrogate target $\LLbar(\vfunc) =\log p(\vy\mid \vfunc)$ and setting the likelihood covariance to $\MOmega=\Cov[\vy\mid\postmean]:= \Cov[\vy\mid \vfunc]_{\vfunc=\postmean}$, after which the update equations become:
\begin{gather} \label{eq:taylor-target}
	\begin{aligned}
		\left.
		\begin{array}{ll}
			\LLbar(\vfunc_n) &= \log \N(\vy \mid \E[\vy_n\mid \vfunc_n], \, \Cov[\vy_n\mid \vfunc_n]) \\
			\jacobian_{k,n} &= {\nabla_{\vfunc_n} \E[\vy_n\mid \postmean_{k,n}]}^\T \Cov[\vy_n\mid \postmean_{k,n}]^{-1} (\vy_n - \E[\vy_n\mid \postmean_{k,n}]) \\
			\hessian_{k,n,n} &= -{\nabla_{\vfunc_{n}} \E[\vy_n\mid \postmean_{k,n}]}^\T \Cov[\vy_n\mid \postmean_{k,n}]^{-1} \nabla_{\vfunc_n} \E[\vy_n\mid \postmean_{k,n}]) \\
		\end{array} \,\, \right\} \quad \parbox[]{15em}{surrogate target \\ \& gradients \\ (Taylor)}
	\end{aligned}
\end{gather}
where $\E[\vy_n\mid \postmean_{k,n}] = \E[\vy_n\mid \vfunc_n]|_{\vfunc_n=\postmean_{k,n}}$.
Here $\hessian_{k,n,n}$ is again a first-order approximation to the true Hessian, \ie, a Gauss--Newton approximation (see \cref{sec:bayes-gauss-newton} and \cref{sec:eks-gauss-newton}). This approach is equivalent to the updates used in the iterated extended Kalman smoother \citep[EKS,][]{bell1994iterated} when performing inference in nonlinear state space models.

\subsubsection{The Posterior Linearisation Energy}

Unfortunately, it is not clear how to construct a marginal likelihood approximation using the computations involved in PL. \citet{garcia2019gaussian} suggest an approximation for Gaussian process models similar to the PEP energy, but where they discard some terms that do not depend on the model hyperparameters. The best approach may therefore be to use the full PEP energy, however this is sub-optimal since does not re-use the computations performed during inference. Alternatively, the VFE could also be used.

On the other hand, since the Taylor/EKS approach is equivalent to a Gauss--Newton method, the Laplace energy, \cref{eq:laplace-energy}, is a natural candidate to be used as the negative log marginal likelihood approximation in this case.

\subsection{Method Comparison}

\cref{tab:targets} compares the surrogate targets used for updating the approximate likelihood parameters. Laplace (\ie, Newton) and Taylor (\ie, EKS) both compute the derivatives of the target with respect to $\vfunc_n$ and then evaluate them at the mean, $\vf_n=\postmean_n$. The Bayes--Newton methods (PL, PEP, VI) marginalise out $\vfunc_n$ and then compute the gradients with respect to the posterior (or cavity) mean. \cref{tab:approx-types} compares the types of approximations employed by the each inference algorithm, and \cref{fig:flow-chart} is a flow chart showing the connections between the methods.

\begin{table}[t!]
	\scriptsize
	\renewcommand{\arraystretch}{2}
	\caption{Comparison of optimisation target and derivative approximations used by various Bayesian inference schemes. $\MR_n={\cavcovfull}_{n,n}^{-1} ( \alpha \nabla_{\cavmeanfull_n}^2\LLbar(\cavmeanfull_{n}) + {\cavcovfull}_{n,n}^{-1} )^{-1}$ is the PEP scaling factor, $q^{\cav}(\vfunc_n) \propto q(\vfunc_n)/t^\alpha(\vfunc_n)$ is the PEP cavity, $\MOmega_{n,n}$ is the mean square error matrix of the SLR approximation, \cref{eq:pl-mse}, and $\vnu(\vfunc_n)=\E[\vy_n \mid \vfunc_n]$, $\MSigma(\vfunc_n)=\Cov[\vy_n\mid \vfunc_n]$ are the first two moments of $p(\vy \mid \vfunc)$.}
	\setlength{\tabcolsep}{2.6pt}
	\vspace{-0.2cm}
	\begin{tabularx}{\columnwidth}{lccc}
		\toprule
		& \sc Surrogate target, $\LLbar(\cdot)$ & \sc Jacobian, $\jacobian_n$ & \sc Hessian, $\hessian_{n,n}$ \\ 
		\toprule
		\sc Newton & $\log p(\vy_n \mid \vfunc_n)$ & $\nabla_{\vfunc_n} \LLbar_n(\postmean_n)$ & $\nabla^2_{\vfunc_n} \LLbar_n(\postmean_n)$ \\ 
		\sc Taylor & $\log \N(\vy_n \mid \vnu(\vfunc_n), \, \MSigma(\vfunc_n))$ & $ {\nabla_{\vfunc_n} \vnu(\postmean_n)}^\T \MSigma(\postmean_n)^{-1} (\vy_n-\vnu(\postmean_n))$ & $- {\nabla_{\vfunc_n} \vnu(\postmean_n)}^\T \MSigma(\postmean_n)^{-1}{\nabla_{\vfunc_n} \vnu(\postmean_n)}$ \\ 
		\sc PL & $\log \N(\vy_n \mid \E_{q} [ \vnu(\vfunc_n) ], \MOmega_{n,n})$ & ${\nabla_{\postmean_n} \E_{q}[\vnu(\vfunc_n)]}^\T\MOmega_{n,n}^{-1} (\vy_n-\E_{q}[\vnu(\vfunc_n)])$ & $- {\nabla_{\postmean_n} \E_{q}[\vnu(\vfunc_n)]}^\T\MOmega_{n,n}^{-1}{\nabla_{\postmean_n} \E_{q}[\vnu(\vfunc_n)]}$ \\ 
		\sc VI & $\E_{q}[\log p(\vy_n | \vfunc_n)]$ & $\nabla_{\postmean_n} \LLbar_n(\postmean_n, \postcov_{n,n})$  & $\nabla^2_{\postmean_n} \LLbar_n(\postmean_n, \postcov_{n,n})$ \\ 
		\sc PEP & $ \frac{1}{\alpha} \log \E_{q^{\cav}}[p(\vy_n | \vfunc_n)]$ & $\MR_n \nabla_{\cavmeanfull_n} \LLbar(\cavmeanfull_{n}, \cavcovfull_{n,n})$ & $ \MR_n \nabla^2_{\cavmeanfull_n} \LLbar(\cavmeanfull_{n}, \cavcovfull_{n,n})$ \\ 
		\bottomrule
	\end{tabularx}\label{tab:targets}
\end{table}

\begin{table}[t!]
	\footnotesize
	\centering
	\renewcommand{\arraystretch}{2.3}
	\setlength{\tabcolsep}{10pt}
	\caption{A comparison of the different method and approximation types employed by various Bayesian inference schemes. The Newton and Taylor methods are mode-seeking algorithms whose updates depend only on the posterior mean, whereas the Bayes--Newton methods (PL, PEP, VI) optimise the full posterior mean and covariance. It can be shown that Taylor and PL make Gauss--Newton approximations to the Hessian of the optimisation target.}
	\vspace{-0.1cm}
	\begin{tabularx}{\columnwidth}{lcc} 
		\toprule
		& \sc Method Type & \sc Approximation Type \\
		\toprule
		\sc Newton/Laplace &  Newton & Laplace approximation around posterior mode \\ 
		\sc Taylor/EKS & Gauss--Newton & Taylor expansion around posterior mode \\
		\sc PL &  Bayes--Gauss--Newton & \pbox{20cm}{$\mathrm{arg}\min_{q(\vy \mid \vfunc)}{\E_{p(\vfunc \mid \vy)}[\KL{p(\vy \mid \vfunc)}{q(\vy \mid \vfunc)}]}$ \\ Optimal likelihood linearisation in MSE sense} \\
		\sc PEP & \pbox{20cm}{Bayes--Newton} & \pbox{20cm}{$\approx \mathrm{arg}\,\min_{q(\vfunc)}{\DivAlpha{p(\vfunc \mid \vy)}{q(\vfunc)}}$ \\ Stationary point of PEP Energy} \\ 
		\sc VI & \pbox{20cm}{Bayes--Newton} & \pbox{20cm}{$\mathrm{arg}\,\min_{q(\vfunc)}{\KL{q(\vfunc)}{p(\vfunc \mid \vy)}}$ \\ Minimizes variational free energy (VFE)} \\
		\bottomrule
	\end{tabularx}\label{tab:approx-types}
	\vspace{0.3cm}
\end{table}

\tikzstyle{block} = [rectangle, draw, fill=black!12, 
text width=6.5em, text centered, rounded corners, minimum height=4em]
\tikzstyle{line} = [draw=black!50, -latex', line width=2pt]

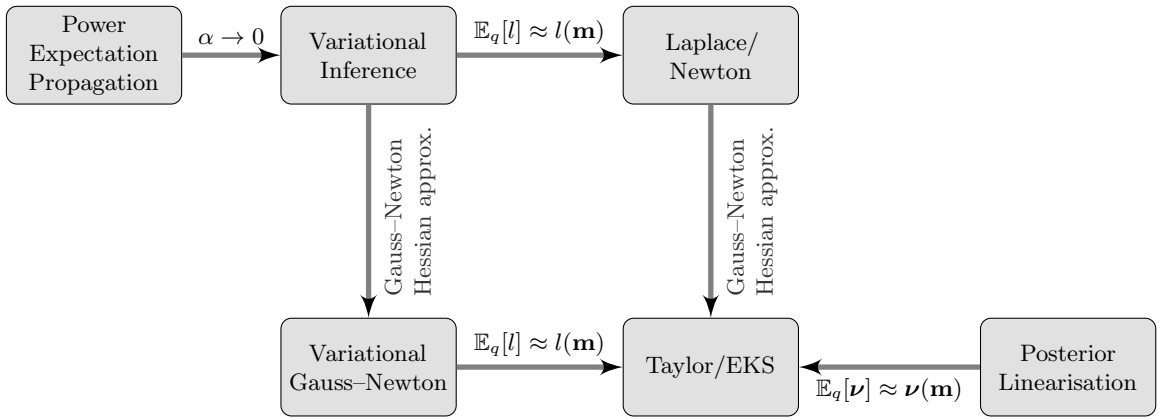
\begin{figure}
	\begin{tikzpicture}[node distance = 4.15cm, auto]
		\footnotesize

		\node [block] (pep) {Power Expectation Propagation};
		\node [block, right of=pep, xshift=-0.5cm] (vi) {Variational Inference};
		\node [block, below of=vi] (vgn) {Variational Gauss--Newton};
		\node [block, right of=vi, xshift=0.4cm] (laplace) {Laplace/ Newton};
		\node [block, below of=laplace] (taylor) {Taylor/EKS};
		\node [block, right of=taylor, xshift=0.6cm] (pl) {Posterior Linearisation};

		\path [line] (pep) -- node {$\alpha \rightarrow 0$}(vi);
		\path [line] (vi) -- node {
			\begin{tikzpicture}
				\node [rotate=90, align=left, opacity=0.8] {Gauss--Newton \\ Hessian approx.};
			\end{tikzpicture}
		}(vgn);
		\path [line] (vi) -- node {$\E_q[l] \approx l(\postmean)$}(laplace);
		\path [line] (vgn) -- node {$\E_q[l] \approx l(\postmean)$}(taylor);
		\path [line] (laplace) -- node {
			\begin{tikzpicture}
				\node [rotate=90, align=left, opacity=0.8] {Gauss--Newton \\ Hessian approx.};
			\end{tikzpicture}
		}(taylor);
		\path [line] (pl) -- node {$\E_q[\vnu] \approx \vnu(\postmean)$}(taylor);
	\end{tikzpicture}
	\caption{A graphical representation of the links between various inference methods. When the EP power tends to zero, the method becomes identical to natural gradient variational inference. Replacing the expectations of $l=\log p(\vy \mid \vfunc)$ in the VI updates with point estimates leads to the Laplace approximation / Newton's method. Applying a (generalised) Gauss--Newton approximation to the Hessian of the target results in the Taylor method (extended Kalman smoother). Similarly, replacing the expectations of $\vnu=\E[\vy\mid\vfunc]$ in PL with point estimates leads to the Taylor method. In \cref{sec:bayes-gauss-newton} we show that a Gauss--Newton approximation can also be applied to VI, which guarantees PSD covariance updates.}\label{fig:flow-chart}
\end{figure}

\subsection{Issues with Newton and Bayes--Newton Methods}

There are two sources of potential instability in the algorithms described above: \emph{(i)}~the algorithms that involve a full Hessian computation may result in the approximate likelihood covariance, $\likcov$, being negative definite, such that the algorithm fails, \emph{(ii)}~the algorithms may oscillate or diverge. In the following sections, we present ways to address issue $i)$ by developing Bayesian variants of approximation schemes from the optimisation literature, namely Gauss--Newton and Quasi-Newton methods.

Regarding issue \emph{(ii)}, whilst a practical approach is to choose a relatively small step size, Newton's method with a constant step size is not guaranteed to converge, even for convex objective functions. It is therefore often used in conjunction with globalisation strategies such as line-search or trust region methods \citep{nocedal2006numerical}. We consider these strategies to be beyond the scope of this work. However, we hope that our approach motivates development of these ideas in the future.

\subsubsection{Heuristic Methods for Ensuring PSD Updates}\label{sec:heuristic}

One common way to improve the conditioning of a matrix is to add a constant value to its diagonal entries. When applying Newton's method, this is a valid way to improve stability, because adding a infinitely large value to the diagonal of the Hessian results in the steepest descent algorithm. For the Bayes--Newton algorithms, adding a large constant to the diagonal is not valid, since the Hessian is also used as the approximate likelihood precision, so modifying its entries can lead to poor results.

Therefore we propose a simple heuristic approach to ensuring that updates remain PSD, which will serve as a baseline method for our Bayes--Gauss--Newton and Bayes-quasi-Newton methods presented in \cref{sec:bayes-gauss-newton} and \cref{sec:bayes-quasi-newton}. Given a possibly indefinite precision matrix, $\likcov^{-1}$, our heuristic approach sets all the off-diagonal entries to zero, and replaces all the negative diagonal entries with a small positive value, $\varepsilon=0.01$.

\section{Bayes--Gauss--Newton} \label{sec:bayes-gauss-newton}

Here we will first derive a Gauss--Newton approximation to Newton's method for approximate inference. After which, we will show how a similar Bayesian analogy results in an efficient and stable VI method that guarantees the likelihood covariance remain PSD, and show how this presentation provides insight into the PL and EKS methods. Unfortunately, it is not possible to derive a similar Gauss--Newton approximation to PEP, because Bonnet's and Price's theorems (see \cref{sec:variational-gauss-newton}) do not apply to the PEP surrogate target.

\subsection{The (Partial) Gauss--Newton Method for Approximate Bayesian Inference}\label{sec:gauss-newton}

Consider again Newton's method applied to the log posterior, $\LL(\vfunc) = \log p(\vfunc \mid \vy)$, with a Gaussian prior, \cref{eq:Laplace}. We now reformulate this algorithm as a \emph{nonlinear least-squares problem}, the form required in order to derive and apply a Gauss--Newton approximation \citep{bjorck1996numerical}. This rewriting of the model into a least-squares form is a crucial step. However, it seems to have been neglected in the approximate inference literature \citep[\eg,][]{khan2021bayesian}, where it is common to simply use the approximation $\nabla_\vfunc^2 \LL(\vfunc) \approx \nabla_\vfunc \LL(\vfunc)^\T \nabla_\vfunc \LL(\vfunc)$, which despite guaranteeing PSD updates is not a correct Gauss--Newton method.

We first consider the case where the likelihood is a Gaussian noise model of the form,
\begin{align} 
	\log p(\vy \mid \vfunc) &= \log \N(\vy \mid \E[\vy \mid \vfunc] , \Cov[\vy \mid \vfunc]) \nonumber \\
	&= \log Z(\vfunc) - \frac{1}{2} (\vy - \E[\vy \mid \vfunc])^\T \Cov[\vy \mid \vfunc]^{-1} (\vy - \E[\vy \mid \vfunc]) , \label{eq:gn_lik}
\end{align}
where $Z(\vfunc)=(2 \pi)^{-N/2} | \Cov[\vy \mid \vfunc] |^{-1/2} $ is the normaliser, and where $\E[\vy \mid \vfunc]$ and $\Cov[\vy \mid \vfunc]$ are the conditional moments, which may be nonlinear functions of $\vfunc$ (making inference intractable). For example, the heteroscedastic noise likelihood (see \cref{sec:hsced}) takes this form, as do many models from the signal processing literature involving dynamical systems corrupted by noise. For models of this type, we can define the vector $\MV(\vfunc)$ to be a collection of \emph{residual components} as follows,
\begin{equation}
	\MV(\vfunc)=\left[ 
	\begin{matrix}
		\Cov[\vy_1\mid \vfunc_1]^{-\frac{1}{2}} (\vy_1 - \E[\vy_1 \mid \vfunc_1]) \\
		\vdots \\
		\Cov[\vy_N\mid \vfunc_N]^{-\frac{1}{2}} (\vy_N - \E[\vy_N \mid \vfunc_N]) \\
		\priorcov^{-\frac{1}{2}} (\vfunc - \priormean)
	\end{matrix}
	\right] ,
\end{equation}
such that the log posterior can be written $\log p(\vfunc \mid \vy) = -\frac{1}{2} \MV(\vfunc)^\T \MV(\vfunc) + \log Z(\vfunc) + c = -\frac{1}{2} \| \MV(\vfunc) \|_2^2 + \log Z(\vfunc) + c$, where $c = -\frac{N}{2} \log(2\pi) -\frac{1}{2} \log |\priorcov| - p(\vy)$ collects the terms that do not depend on $\vfunc$ and so can be ignored. This shows that optimisation of the target can be cast as a \emph{partial nonlinear least-squares problem}. We use the term `partial' because the log normaliser may depend on $\vfunc$, but cannot be written in the required least-squares form.

The Jacobian of the residual vector is obtained by taking the partial derivatives of each component,
\begin{equation}
	\nabla_\vfunc \MV(\vfunc)=\left[ 
	\begin{matrix}
		\MG_1 \\
		\vdots \\
		\MG_N \\
		\priorcov^{-\frac{1}{2}}
	\end{matrix}
\right] , \\
\end{equation}
where,
\begin{align}
	\MG_n &= \nabla_{\vfunc^\T} \left( \Cov[\vy_n\mid \vfunc_n]^{-\frac{1}{2}} (\vy_n - \E[\vy_n \mid \vfunc_n]) \right) \nonumber \\
	&= \left[\bm{0}, \dots, \nabla_{\vfunc_n} \left( \Cov[\vy_n\mid \vfunc_n]^{-\frac{1}{2}} (\vy_n - \E[\vy_n \mid \vfunc_n]) \right), \dots, \bm{0} \right] \in \R^{\ydim\times N\fdim}.
\end{align}
The residual Hessian $\nabla^2_\vfunc \MV(\vfunc)$ is defined similarly. The Jacobian and Hessian of the log posterior can then be computed via the chain rule,
\begin{gather}
	\begin{aligned}
	\nabla_\vfunc \LL(\vfunc) &= -\nabla_\vfunc \MV(\vfunc)^\T \MV(\vfunc) + \nabla_\vfunc \log Z(\vfunc) , \\
	\nabla^2_\vfunc \LL(\vfunc) &= -\nabla_\vfunc \MV(\vfunc)^\T \nabla_\vfunc \MV(\vfunc) - \nabla^2_\vfunc \MV(\vfunc)^\T \MV(\vfunc) + \nabla^2_\vfunc \log Z(\vfunc) .
	\end{aligned}
\end{gather}
A \emph{Gauss--Newton} approximation to the Hessian involves discarding the second-order terms to give
\begin{align} 
		\nabla^2_\vfunc \LL(\vfunc) &\approx -\nabla_\vfunc \MV(\vfunc)^\T \nabla_\vfunc \MV(\vfunc) \nonumber \\
		&= - \sum_{n=1}^N \MG_n^\T \MG_n - \priorcov^{-1} . \label{eq:gauss-newton}
\end{align}
Inference based on \cref{eq:gauss-newton} is guaranteed to result in PSD covariances because $\MG_n^\T \MG_n$ is PSD. Note that $\sum_{n=1}^N \MG_n^\T \MG_n$ is a block-diagonal matrix with block size $\fdim$.

For many likelihood models the second-order term, $\nabla^2_\vfunc \MV(\vfunc)^\T \MV(\vfunc)$, will be zero or close to zero, meaning that the Gauss--Newton approach is either exact or a good approximation. However, Gauss--Newton is arguably most desirable in scenarios where the second-order term is not close to zero, since this often causes non-PSD updates to occur.

\cref{eq:gauss-newton} also discards the log normaliser term, $\nabla^2_\vfunc \log Z(\vfunc)$. For some models the normaliser $Z(\vfunc)$ does not depend on $\vfunc$, and hence this term will be zero. Even if $Z(\vfunc)$ does depend on $\vfunc$, we might still expect the first order term to dominate the Hessian computation. However, inclusion of the normaliser Hessian may improve the inference result. We propose to replace this term with the following approximation: $\nabla^2_\vfunc \log Z(\vfunc) \approx -\nabla_\vfunc \log Z(\vfunc)^\T \nabla_\vfunc \log Z(\vfunc)$. This is not a proper Gauss--Newton approximation, but does guarantee PSD updates. We call this approximation the \emph{partial Gauss--Newton} method:
\begin{equation}\label{eq:pgn}
	\nabla^2_\vfunc \LL(\vfunc) \approx -\nabla_\vfunc \MV(\vfunc)^\T \nabla_\vfunc \MV(\vfunc) -\nabla_\vfunc \log Z(\vfunc)^\T \nabla_\vfunc \log Z(\vfunc) .
\end{equation}

\subsubsection{Generalised Gauss--Newton for General Likelihoods}

We now consider the case where the likelihood model cannot be written in the Gaussian form of \cref{eq:gn_lik}, for example in discrete models such as the Bernoulli or Poisson likelihoods, or continuous models such as the Gamma or Student-$t$ likelihoods. In this case, the model can no longer be cast as a least-squares problem, so we resort to a \emph{generalised Gauss--Newton} approach. The generalised approach, initially developed to address constrained optimisation problems \citep{golub1973differentiation}, and more recently applied to Bayesian neural networks \citep{khan2019approximate}, works by defining a transformation of variables such that application of the chain rule leads to a sum of first- and second-order terms. We propose the transformation $\vu(\vfunc) = \E[\vy \mid \vfunc]$, giving
\begin{gather}
	\begin{aligned}
		\nabla_\vfunc \log p(\vy \mid \vu(\vfunc)) &= \nabla_\vfunc \vu(\vfunc)^\T \nabla_\vu \log p(\vy \mid \vu(\vfunc)) , \\
		\nabla^2_\vfunc \log p(\vy \mid \vu(\vfunc)) &= \nabla^2_\vfunc \vu(\vfunc)^\T \nabla_\vu \log p(\vy \mid \vu(\vfunc)) + \nabla_\vfunc^\T \vu(\vfunc) \nabla^2_\vu \log p(\vy \mid \vu(\vfunc)) \nabla_\vfunc \vu(\vfunc) \\
		&\approx \nabla_\vfunc^\T \vu(\vfunc) \nabla^2_\vu \log p(\vy \mid \vu(\vfunc)) \nabla_\vfunc \vu(\vfunc) ,
	\end{aligned}
\end{gather}
where the final line amounts to the generalised Gauss--Newton approximation. However, $\nabla^2_\vu \log p(\vy \mid \vu(\vfunc))$ is not guaranteed to be negative semi-definite as required. We therefore propose using the Laplace approximation, $\nabla^2_\vu \log p(\vy \mid \vu(\vfunc)) \approx -\Cov[\vy \mid \vfunc]^{-1}$, inspired by the fact that for many models this Hessian term will be well approximated by the negative likelihood precision, which is guaranteed to be negative semi-definite. This leads to the following generalised Gauss--Newton method
\begin{align} 
	\nabla^2_\vfunc \log p(\vy \mid \vfunc) &\approx -\nabla_\vfunc \E[\vy \mid \vfunc]^\T \Cov[\vy \mid \vfunc]^{-1} \nabla_\vfunc \E[\vy \mid \vfunc] \nonumber \\
	&= -\sum_{n=1}^N \MG_n^\T \MG_n , \label{eq:ggn}
\end{align}
where $\MG_n = \Cov[\vy_n \mid \vfunc_n]^{-\frac{1}{2}} \nabla_{\vfunc^\T} \E[\vy_n \mid \vfunc_n]$. This version matches the full Gauss--Newton method when $\nabla_\vfunc \Cov[\vy \mid \vfunc]$ and $\nabla_\vfunc \log Z(\vfunc)$ are both zero (or assumed to be zero). Interestingly, \cref{eq:ggn} also matches the Taylor method / EKS updates in \cref{eq:taylor-target} exactly. Next we will show that our presentation has the benefit of allowing for a variational Gauss--Newton extension, which improves on the EKS approach whilst still ensuring PSD updates.

\subsection{Variational Gauss--Newton}\label{sec:variational-gauss-newton}

As discussed in \cref{sec:vi}, natural gradient VI can be derived by defining the expected log likelihood, $\LLbar(\postmean, \postcov) = \E_{q(\vfunc)}[\log p(\vy \mid \vfunc)]$, to be a surrogate optimisation target. Therefore, assuming the continuous Gaussian model in \cref{eq:gn_lik} and letting
\begin{equation}
	\MV(\vfunc)=\left[ 
	\begin{matrix}
		\Cov[\vy_1\mid \vfunc_1]^{-\frac{1}{2}} (\vy_1 - \E[\vy_1 \mid \vfunc_1]) \\
		\vdots \\
		\Cov[\vy_N\mid \vfunc_N]^{-\frac{1}{2}} (\vy_N - \E[\vy_N \mid \vfunc_N])
	\end{matrix}
	\right] ,
\end{equation}
the expected log likelihood can be written,
\begin{equation}
	\E_{q(\vfunc)}[\log p(\vy \mid \vfunc)] = \E_{q(\vfunc)} \left[ -\frac{1}{2} \MV(\vfunc)^\T \MV(\vfunc) + \log Z(\vfunc) \right] + c ,
\end{equation}
where $c$ again collects the terms that do not depend on $\vfunc$. By Bonnet's and Price's theorems \citep[see][]{lin2019stein} we can write $\nabla_\postmean \E_{q(\vfunc)}[\log p(\vy \mid \vfunc)] = \E_{q(\vfunc)}[ \nabla_\vfunc \log p(\vy \mid \vfunc)]$, and $\nabla^2_\postmean \E_{q(\vfunc)}[\log p(\vy \mid \vfunc)] = \E_{q(\vfunc)}[ \nabla^2_\vfunc \log p(\vy \mid \vfunc)]$, such that
\begin{gather}\label{eq:variational-gn}
	\begin{aligned}
		\nabla_\postmean \E_{q(\vfunc)}[\log p(\vy \mid \vfunc)] &= \E_{q(\vfunc)}[ -\nabla_\vfunc \MV(\vfunc)^\T \MV(\vfunc) + \nabla_\vfunc \log Z(\vfunc) ] , \\
		\nabla^2_\postmean \E_{q(\vfunc)}[\log p(\vy \mid \vfunc)] & = \E_{q(\vfunc)}[ -\nabla_\vfunc \MV(\vfunc)^\T \nabla_\vfunc \MV(\vfunc) - \nabla^2_\vfunc \MV(\vfunc)^\T \MV(\vfunc) + \nabla^2_\vfunc \log Z(\vfunc) ] \\
		&\approx \E_{q(\vfunc)} \left[ - \nabla_\vfunc \MV(\vfunc)^\T \nabla_\vfunc \MV(\vfunc) \right] \\
		&= \E_{q(\vfunc)} \left[ - \sum_{n=1}^N \MG_n^\T \MG_n \right] ,
	\end{aligned}
\end{gather}
where $\MG_n = \nabla_{\vfunc^\T} \left( \Cov[\vy_n\mid \vfunc_n]^{-\frac{1}{2}} (\vy_n - \E[\vy_n \mid \vfunc_n]) \right)$. Here $\MG_n^\T \MG_n$ is guaranteed to be PSD such that $\E_{q(\vfunc_n)} \left[ \MG_n^\T \MG_n \right]$ is also PSD. We refer to \cref{eq:variational-gn} as the \emph{variational Gauss--Newton} method. As in \cref{sec:gauss-newton}, a potentially more accurate approximation is the \emph{partial variational Gauss--Newton} method,
\begin{equation}
	\nabla^2_\postmean \E_{q(\vfunc)}[\log p(\vy \mid \vfunc)] \approx \E_{q(\vfunc)} \left[ - \nabla_\vfunc \MV(\vfunc)^\T \nabla_\vfunc \MV(\vfunc) - \nabla_\vfunc \log Z(\vfunc)^\T \nabla_\vfunc \log Z(\vfunc) \right] .
\end{equation}
Empirically, we find partial variational Gauss--Newton to be very effective.

As above, for discrete likelihoods we again propose the use of $\MG_n = \Cov[\vy_n \mid \vfunc_n]^{-\frac{1}{2}} \nabla_{\vfunc^\T} \E[\vy_n \mid \vfunc_n]$, leading to the following approximation, \sloppy
\begin{equation} \label{eq:vggn}
	\nabla^2_\postmean \E_{q(\vfunc)}[\log p(\vy \mid \vfunc)] \approx - \E_{q(\vfunc)} \left[ \nabla_{\vfunc} \E[\vy \mid \vfunc] \Cov[\vy \mid \vfunc]^{-1} \nabla_{\vfunc} \E[\vy \mid \vfunc] \right] ,
\end{equation}
that we refer to as the \emph{variational generalised Gauss--Newton} method. \cref{eq:vggn} is equivalent to the Taylor/EKS method where we take an expectation with respect to the full posterior rather than simply evaluating the Hessian at the posterior mean. It also bears an interesting resemblance to the PL updates, \cref{eq:pl-target}, making it clear that whilst PL and variational Gauss--Newton are both Bayes--Gauss--Newton methods, PL makes additional limiting assumptions and approximates the likelihood covariance differently.

The variational Gauss--Newton method is guaranteed to produce PSD updates even in the presence of numerical integration error when computing $\E_{q(\vfunc)}[\log p(\vy \mid \vfunc)]$, and in scenarios where the model hyperparameters change across iterations. This gives it a significant advantage relative to the PSD constraints based on Riemannian gradients in \cref{sec:psd-constraints}.

\subsection{Second-Order Posterior Linearisation}\label{sec:improved-pl}

In \cref{sec:pl-gauss-newton}, we show how the PL updates in \cref{eq:pl-target} amount to a Bayes--Gauss--Newton approximation, with the Hessian computation only involving first-order derivatives. Additionally, the covariance, $\MOmega_{n,n}$, is assumed to have zero gradient with respect to $\postmean_{n}$. This means PL results in a poor approximation for likelihood models of the form given in \cref{eq:gn_lik} where $\Cov[\vy \mid \vfunc]$ depends on $\vfunc$. We propose a full Newton-like version of PL which takes into account this dependency and includes the second-order terms. Recall that the PL surrogate target is $\LLbar(\postmean_n, \postcov_{n,n}) = \log \N(\vy_n \mid \E_{q(\vfunc_n)} [ \E[\vy_n\mid \vfunc_n] ], \, \MOmega_{n,n})$. Letting $Z_n$ be the normaliser of this Gaussian gives
\begin{gather} \label{eq:pl-newton-target}
	\begin{aligned}
		\left.
		\begin{array}{ll}
			\MD_{k,n} &\!\!= \MOmega_{k,n,n}^{-\frac{1}{2}} (\vy_n - \E_{q(\vfunc_n)} [ \E[\vy_n\mid \vfunc_n] ]) \\
			\LLbar(\postmean_n, \postcov_{n,n}) &\!\!= \log Z_n - \frac{1}{2} \MD_n^\T \MD_n   \\ 
			\jacobian_{k,n} &\!\!= \nabla_{\postmean_n} \log Z_n - \nabla_{\postmean_n} \!\MD_{k,n}^\T \MD_{k,n} \\
			\hessian_{k,n,n} &\!\!= \nabla^2_{\postmean_n} \log Z_n -\nabla_{\postmean_n} \!\MD_{k,n}^\T \nabla_{\postmean_n}\!\MD_{k,n} - \nabla^2_{\postmean_n} \!\MD_{k,n}^\T \MD_{k,n} \\
		\end{array} \,\, \right\} \quad \parbox[]{15em}{surrogate target \\ \& gradients \\ (2\textsuperscript{nd}-order PL)}
	\end{aligned}
\end{gather}

Following the approach above, a partial Gauss--Newton approximation can then be derived by setting
\begin{equation}
	\hessian_{k,n,n} = -\nabla_{\postmean_n} \!\MD_{k,n}^\T \nabla_{\postmean_n}\!\MD_{k,n} - \nabla_{\postmean_n} \log Z_n^\T \nabla_{\postmean_n} \log Z_n ,
\end{equation}
which improves upon standard PL when $\Cov[\vy \mid \vfunc]$ depends on $\vfunc$, and still guarantees PSD updates. If the likelihood model is not of the form given in \cref{eq:gn_lik}, for example if it is discrete, then a generalised Gauss--Newton approximation in which the normaliser $Z_n$ and the gradients of $\MOmega_{k,n,n}$ are ignored results in exactly the standard PL method.

\section{Bayes-Quasi-Newton} \label{sec:bayes-quasi-newton}

As discussed in \cref{sec:gauss-newton}, Gauss--Newton methods are accurate when the Hessian computation is dominated by the first-order term. If this is not the case, a better approximation may be a quasi-Newton method \citep{broyden1967quasi, nocedal2006numerical}. Quasi-Newton methods replace the full Hessian computation with a series of efficient low-rank updates, and whilst these updates are not guaranteed to result in PSD covariances, they do provide a way of checking whether a given update will be PSD, which can be used to determine whether an update should be applied. A \emph{damped} version of the updates can also be applied, which does guarantee that the resulting covariances are PSD.

\subsection{The Quasi-Newton Method}\label{sec:quasi-newton}

Quasi-Newton methods approximate the Hessian of the optimisation target, $\nabla^2\LL$, via a matrix, $\MB$, in a way that avoids the Hessian computation by utilising changes in first-order derivative information along the optimisation search direction. The application of the quasi-Newton method to the global Newton updates, \cref{eq:Newton}, whilst extremely efficient, is a poor approximation because the full-rank matrix $\postcov^{-1}$ cannot be well approximated by iterative low-rank updates. Therefore, we apply individual quasi-Newton updates to the local approximate likelihood terms, $\LLbar(\vfunc_n) = \log p(\vy_n \mid \vfunc_n)$:
\begin{equation}
	\nabla^2_{\vfunc_n} \LLbar(\vfunc_n) \approx \MB_{k,n} .
\end{equation}

The matrix $\MB_{k,n}$ is chosen to ensure it satisfies the \emph{secant equation}: $\MB_{k+1,n} \vs_{k,n}=\vg_{k,n}$, where $\vs_{k,n}=\postmean_{k+1,n}-\postmean_{k,n}$ and $\vg_{k,n}=\nabla_{\vfunc_n} \LLbar(\postmean_{k+1,n})-\nabla_{\vfunc_n}\LLbar(\postmean_{k,n})$. Various methods exist for doing do, but we focus on the \emph{BFGS} formula \citep{broyden1969new, fletcher1970new, goldfarb1970family, shanno1970conditioning}, because it has been shown empirically to be the most effective, and because it guarantees that $\MB_{k,n}$ remains negative semi-definite whenever the initial value, $\MB_{0,n}$, is negative semi-definite and $\vs_{k,n}^\T \vg_{k,n} < 0$.

The BFGS formula, which amounts to a rank-two update to $\MB_{k,n}$, is given by
\begin{gather}
	\begin{aligned}
		\MB_{k+1,n} = \MB_{k,n} - \frac{\MB_{k,n} \vs_{k,n} \vs_{k,n}^\T\MB_{k,n}}{\vs_{k,n}^\T\MB_{k,n} \vs_{k,n}} + \frac{\vg_{k,n} \vg_{k,n}^\T}{\vs_{k,n}^\T\vg_{k,n}}.
	\end{aligned}
\end{gather}
The curvature condition, $\vs_{k,n}^\T \vg_{k,n} < 0$, is not guaranteed for general nonlinear functions, and we observe it to often be violated for non-log-concave likelihood models. A practical approach to ensure stability is to simply reject updates that do no satisfy the condition. The local quasi-Newton updates are then
\begin{gather} \label{eq:quasi-newton-target}
	\begin{aligned}
		\hspace{-0.2cm}
		\left.
		\begin{array}{ll}
			\LLbar(\vfunc_n) &= \log p(\vy_n \mid \vfunc_n)  \\
			\jacobian_{k,n} &= \nabla_{\vfunc_n} \LLbar(\postmean_{k,n})  \\
			\hessian_{k,n,n} &= \MB_{k,n} \\
			\vs_{k,n}&=\postmean_{k+1,n}-\postmean_{k,n} \, , \quad \vg_{k,n}=\nabla_{\vfunc_n} \LLbar(\postmean_{k+1,n})-\nabla_{\vfunc_n}\LLbar(\postmean_{k,n}) \\
			\MB_{k+1,n} &= \left\{ \begin{array}{ll}
				\MB_{k,n} - \frac{\MB_{k,n} \vs_{k,n} \vs_{k,n}^\T\MB_{k,n}}{\vs_{k,n}^\T\MB_{k,n} \vs_{k,n}} + \frac{\vg_{k,n} \vg_{k,n}^\T}{\vs_{k,n}^\T\vg_{k,n}} , \quad \text{if } \vs_{k,n}^\T \vg_{k,n} < 0 \\
				\MB_{k,n} \, , \hspace{4.95cm} \text{if } \vs_{k,n}^\T \vg_{k,n} \ge 0
			\end{array} \right.
		\end{array} \,\, \right\} \quad \parbox[]{15em}{surrogate target \\ \& gradients \\ (Quasi-Newton)}
	\end{aligned}
\end{gather}
In some scenarios this approach can cause a large number of the updates to be rejected. We present a damped version whose updates are guaranteed to be stable in \cref{sec:damped-quasi-newton}. Next we will derive two Bayesian variants of the quasi-Newton method corresponding to approximations to VI and PEP.

\subsection{Variational Quasi-Newton}\label{sec:variational-quasi-newton}

The quasi-Newton method cannot be straightforwardly applied to VI by simply changing the target density, because the parameters being optimised in Bayes--Newton methods include the posterior covariance, $\postcov$, as well as the posterior mean, $\postmean$, and the changes in covariance must be accounted for in the secant equation. Since the surrogate target, $\LLbar(\postmean, \postcov) = \E_{q(\vfunc)}[\log p(\vy \mid \vfunc)]$, only depends on the marginal variances ($\vy_n$ only depends on $\vfunc_n$) it is sufficient to define a new vector,
\begin{equation}
	\veta_{n} = \left(
	\begin{matrix}
		\postmean_{n} \\
		\mathrm{vec}(\postcov_{n,n})
	\end{matrix} \right) \in \R^{(\fdim+\fdim^2) \times 1} ,
\end{equation}
where $\mathrm{vec}(\postcov_{n,n}) \in \R^{\fdim^2\times 1}$ represents the covariance of the marginal, $q(\vfunc_n)$, stored as a column vector. We must then compute the Jacobian of the target with respect to $\veta_{n}$. Since computing the Jacobian with respect to the covariance comes at the same computational cost as computing the Hessian with respect to the mean, Bayes-quasi-Newton methods are not computationally more efficient than standard Bayes--Newton methods and should be viewed instead as a way to ensure PSD updates.

The \emph{variational quasi-Newton} method is therefore characterised by the following updates,
\begin{gather} \label{eq:variational-quasi-newton-target}
	\begin{aligned}
		\hspace{-0.2cm}
		\left.
		\begin{array}{ll}
			\LLbar(\veta_{n}) &= \E_{q(\vfunc_n)} [\log p(\vy_n \mid \vfunc_n) ] \\
			\jacobian_{k,n} &= \nabla_{\postmean_{k,n}} \LLbar(\veta_{k,n})  \\
			\hessian_{k,n,n} &= \MB_{k,n,1:\fdim,1:\fdim} \\
			\vs_{k,n}&=\veta_{k+1,n}-\veta_{k,n} \, , \quad \vg_{k,n}=\nabla_{\veta_n} \LLbar(\veta_{k+1,n})-\nabla_{\veta_n}\LLbar(\veta_{k,n}) \\
			\MB_{k+1,n} &= \left\{ \begin{array}{ll}
				\MB_{k,n} - \frac{\MB_{k,n} \vs_{k,n} \vs_{k,n}^\T\MB_{k,n}}{\vs_{k,n}^\T\MB_{k,n} \vs_{k,n}} + \frac{\vg_{k,n} \vg_{k,n}^\T}{\vs_{k,n}^\T\vg_{k,n}} , \quad \text{if } \vs_{k,n}^\T \vg_{k,n} < 0 \\
				\MB_{k,n} \, , \hspace{4.95cm} \text{if } \vs_{k,n}^\T \vg_{k,n} \ge 0
			\end{array} \right.
		\end{array} \,\, \right\} \quad \parbox[]{15em}{surrogate target \\ \& gradients \\ (Variational \\ Quasi-Newton)}
	\end{aligned}
\end{gather}
where $\MB_{k,n,1:\fdim,1:\fdim}$ represents the upper left $\fdim\times \fdim$ block of $\MB_{k,n}\in \R^{(\fdim+\fdim^2) \times (\fdim+\fdim^2)}$, which corresponds to the approximate Hessian with respect to the mean, $\postmean_{k,n}$.

\subsection{Power Expectation Propagation Quasi-Newton}\label{sec:quasi-newton-pep}

We take a similar approach to derive a quasi-Newton approximation to power EP. In this case, we define a new vector which stacks the cavity means and marginal cavity variances,
\begin{equation}
	\veta_{n}^\cav = \left(
	\begin{matrix}
		\cavmeanfull_{n} \\
		\mathrm{vec}(\cavcovfull_{n,n})
	\end{matrix} \right) \in \R^{(\fdim+\fdim^2) \times 1} ,
\end{equation}
where $\mathrm{vec}(\cavcovfull_{n,n})$ is a vectorised version of $\cavcovfull_{n,n}$. The updates are then
\begin{gather} \label{eq:quasi-newton-pep-target}
	\begin{aligned}
		\hspace{-0.2cm}
		\left.
		\begin{array}{ll}
			\LLbar(\veta_{n}^\cav \, ) &= \frac{1}{\alpha} \log \E_{q^{\cav }(\vfunc_n)} [ p^\alpha(\vy_n \mid \vfunc_n) ]  \\
			\MR_{k,n} &= {\cavcovfull}_{k,n,n}^{-1} \left( \alpha \MB_{k,n,1:\fdim,1:\fdim} + {\cavcovfull}_{k,n,n}^{-1} \, \right)^{-1} \\
			\jacobian_{k,n} &= \MR_{k,n} \nabla_{\cavmeanfull_n} \LLbar(\veta_{k,n}^\cav \, )  \\
			\hessian_{k,n,n} &= \MR_{k,n} \MB_{k,n,1:\fdim,1:\fdim} \\
			\vs_{k,n}&=\veta^\cav_{k+1,n}-\veta^\cav_{k,n} \, , \quad \vg_{k,n}\!=\!\nabla_{\veta^\cav_n} \LLbar(\veta^\cav_{k+1,n})-\nabla_{\veta^\cav_n}\LLbar(\veta^\cav_{k,n}) \\
			\MB_{k+1,n} &= \left\{ \begin{array}{ll}
				\MB_{k,n} - \frac{\MB_{k,n} \vs_{k,n} \vs_{k,n}^\T\MB_{k,n}}{\vs_{k,n}^\T\MB_{k,n} \vs_{k,n}} + \frac{\vg_{k,n} \vg_{k,n}^\T}{\vs_{k,n}^\T\vg_{k,n}} , \quad \text{if } \vs_{k,n}^\T \vg_{k,n} < 0 \\
				\MB_{k,n} \, , \hspace{4.95cm} \text{if } \vs_{k,n}^\T \vg_{k,n} \ge 0
			\end{array} \right.
		\end{array} \! \right\} \, \parbox[]{15em}{surrogate target \\ \& gradients \\ (PEP Quasi-Newton)}
	\end{aligned}
\end{gather}

\subsection{Posterior Linearisation Quasi-Newton}\label{sec:quasi-newton-pl}

A quasi-Newton approximation can also be applied to PL in a very similar way to the VI case:
\begin{gather} \label{eq:pl-quasi-newton-target}
	\begin{aligned}
		\left.
		\begin{array}{ll}
			\LLbar(\veta_{n}) &= \log \N(\vy_n \mid \E_{q(\vfunc_n)} [ \E[\vy_n\mid \vfunc_n] ], \, \MOmega_{n,n})   \\ 
			\jacobian_{k,n} &= {\nabla_{\postmean_n} \E_{q(\vfunc_n)} [ \E[\vy_n\mid \vfunc_n] ]}^\T \MOmega_{k,n,n}^{-1} (\vy_n - \E_{q(\vfunc_n)} [ \E[\vy_n\mid \vfunc_n] ]) \\
			\hessian_{k,n,n} &= \MB_{k,n,1:\fdim,1:\fdim} \\
			\vs_{k,n}&=\veta_{k+1,n}-\veta_{k,n} \, , \quad \vg_{k,n}=\nabla_{\veta_n} \LLbar(\veta_{k+1,n})-\nabla_{\veta_n}\LLbar(\veta_{k,n}) \\
			\MB_{k+1,n} &= \left\{ \begin{array}{ll}
				\MB_{k,n} - \frac{\MB_{k,n} \vs_{k,n} \vs_{k,n}^\T\MB_{k,n}}{\vs_{k,n}^\T\MB_{k,n} \vs_{k,n}} + \frac{\vg_{k,n} \vg_{k,n}^\T}{\vs_{k,n}^\T\vg_{k,n}} , \quad \text{if } \vs_{k,n}^\T \vg_{k,n} < 0 \\
				\MB_{k,n} \, , \hspace{4.95cm} \text{if } \vs_{k,n}^\T \vg_{k,n} \ge 0
			\end{array} \right.
		\end{array} \,\, \right\} \,\, \parbox[]{15em}{surrogate target \\  \& gradients \\ (PL Quasi-Newton)}
	\end{aligned}
\end{gather}
where $\nabla_{\veta_n}\LLbar(\veta_{k,n})={\nabla_{\veta_n} \E_{q(\vfunc_n)} [ \E[\vy_n\mid \vfunc_n] ]}^\T \MOmega_{k,n,n}^{-1} (\vy_n - \E_{q(\vfunc_n)} [ \E[\vy_n\mid \vfunc_n] ])$, \ie, the gradient of $\MOmega_{n,n}$ is assumed to be zero. This approach can also be adapted to the improved version of PL derived in \cref{sec:improved-pl} by relaxing this assumption (and including the normalisation constant if appropriate).

\subsection{Damped Quasi-Newton Updates}\label{sec:damped-quasi-newton}

For complicated likelihood models, the BFGS updates described above may result in many updates being rejected, which can result in poor performance. The following \emph{damped} BFGS formula can be used as a slightly less accurate approach which guarantees PSD updates. The adaptive damping term, $\vpsi$, is introduced as follows:
\begin{gather}
	\begin{aligned}
	\vpsi_{k,n} = \left\{ \begin{array}{ll}
		1 \, , & \text{if } \vs_{k,n}^\T \vg_{k,n} \le (1-\xi) \vs_{k,n}^\T\MB_{k,n} \vs_{k,n} \\
		\xi \frac{\vs_{k,n}^\T\MB_{k,n} \vs_{k,n}}{\vs_{k,n}^\T\MB_{k,n} \vs_{k,n} - \vs_{k,n}^\T \vg_{k,n}} \, , \quad\quad & \text{if } \vs_{k,n}^\T \vg_{k,n} > (1-\xi) \vs_{k,n}^\T\MB_{k,n} \vs_{k,n}
	\end{array} \right.
	\end{aligned}
\end{gather}
where $\xi$ is the damping factor (often set to $\xi=0.8$). Then letting
\begin{equation}
	\vr_{k,n} = \vpsi \vg_{k,n} + (1-\vpsi) \MB_{k,n} \vs_{k,n},
\end{equation}
the BFGS update is modified to use $\vr_{k,n}$ in place of $\vg_{k,n}$:
\begin{gather}
	\begin{aligned}
		\MB_{k+1,n} = \MB_{k,n} - \frac{\MB_{k,n} \vs_{k,n} \vs_{k,n}^\T\MB_{k,n}}{\vs_{k,n}^\T\MB_{k,n} \vs_{k,n}} + \frac{\vr_{k,n} \vr_{k,n}^\T}{\vs_{k,n}^\T \vr_{k,n}} \, ,
	\end{aligned}
\end{gather}
which is guaranteed to be negative semi-definite because $\vs_{k,n}^\T \vr_{k,n} = -(1-\xi) \vs_{k,n}^\T\MB_{k,n} \vs_{k,n} < 0$. When $\vpsi_{k,n}=1$, the update is equivalent to the standard BFGS formula. When $\vpsi_{k,n}\in(0,1)$, the update results in a negative semi-definite $\MB_{k+1,n}$ that interpolates between the previous estimate and the standard update. The damped approach is also applicable to the VI and PEP quasi-Newton variants.

Perhaps a more common approach to ensuring PSD updates during quasi-Newton is to the use the Wolfe line-search \citep{nocedal2006numerical}. However, since our approach is to apply the BFGS updates to the individual likelihood terms separately, whereas the objective (the model energy) is global, we found that a very small step size is often required to ensure that \emph{all} terms remain PSD. In contrast, the damping approach effectively allows a different step size for each term, which is much more effective.

\section{PSD Constraints via Riemannian Gradients} \label{sec:psd-constraints}

\citet{lin2020handling} used Riemannian gradient methods to deal with the PSD constraint in VI by adding an additional term to the precision update, in a similar vein to the `retraction map' presented in \citet{tran2019variational}. Since the prior precision is guaranteed to be PSD by construction, we derive the corresponding method applied to the approximate likelihood updates by subtracting the prior component from the method of \citet{lin2020handling} to give
\begin{gather} \label{eq:vi-target-psd}
	\begin{aligned}
		\left.
		\begin{array}{ll}
			\LLbar(\postmean_n, \postcov_{n,n}) &= \E_{q(\vfunc_n)}[ \log p(\vy_n \mid \vfunc_n) ]  \\
			\jacobian_{k,n} &= \nabla_{\postmean_n} \LLbar(\postmean_{k,n}, \postcov_{k,n,n})  \\
			\MG_{k,n} &= \likcov_{k,n,n}^{-1} + \nabla_{\postmean_n}^2\LLbar(\postmean_{k,n}, \postcov_{k,n,n})  \\
			\hessian_{k,n,n} &= \nabla_{\postmean_n}^2\LLbar(\postmean_{k,n}, \postcov_{k,n,n}) - \frac{\rho}{2} \MG_{k,n} \likcov_{k,n,n} \MG_{k,n}  \\
		\end{array} \,\, \right\} \quad \parbox[]{15em}{surrogate target \\ \& gradients \\ (VI Riemann)}
	\hspace{-0.5cm}
	\end{aligned}
\end{gather}
This approach provides excellent performance in many cases, although the degree to which the constraint may affect the quality of the approximation is unclear. Unfortunately, despite having theoretical guarantees that \cref{eq:vi-target-psd} results in PSD updates, in practice we find that numerical integration error when computing $\LLbar(\postmean_n, \postcov_{n,n})$ can cause the algorithm to fail. The constraint is also not guaranteed if the model hyperparameters change across optimisation iterations.

Adding a similar retraction map term to the improved PL method, \cref{eq:pl-newton-target}, is also valid, and by approximating the target expectation with a point estimate, $\LLbar(\postmean_n)=\log p(\vy_n \mid \postmean_{n})$, it is straightforward to derive a Newton / Laplace version. Our unifying presentation also allows us to formulate a similar PSD constraint for PEP as follows:
\begin{gather} \label{eq:pep-target-psd}
	\begin{aligned}
		\left.
		\begin{array}{ll}
			\LLbar(\cavmeanfull_n, \cavcovfull_{n,n}) &= \frac{1}{\alpha} \log \E_{q^{\cav }(\vfunc_n)} [ p^\alpha(\vy_n \mid \vfunc_n) ]  \\
			\MR_{k,n} &= {\cavcovfull}^{-1}_{k,n,n} \left( \alpha \nabla_{\cavmeanfull_n}^2\LLbar(\cavmeanfull_{k,n}, \cavcovfull_{k,n,n}) + {\cavcovfull}_{k,n,n}^{-1} \, \right)^{-1} \\
			\MG_{k,n} &= \likcov_{k,n,n}^{-1} + \MR_{k,n} \nabla_{\postmean_n}^2\LLbar(\cavmeanfull_{k,n}, \cavcovfull_{k,n,n})  \\
			\jacobian_{k,n} &= \MR_{k,n} \nabla_{\cavmeanfull_n} \LLbar(\cavmeanfull_{k,n}, \cavcovfull_{k,n,n})  \\
			\hessian_{k,n,n} &= \MR_{k,n} \nabla_{\cavmeanfull_n}^2\LLbar(\cavmeanfull_{k,n}, \cavcovfull_{k,n,n}) - \frac{\rho}{2} \MG_{k,n} \likcov_{k,n,n} \MG_{k,n} \\
		\end{array} \,\, \right\} \quad \parbox[]{15em}{surrogate target \\ \& gradients \\ (PEP Riemann)}
		\hspace{-1.2cm}
	\end{aligned}
\end{gather}
These constraints for the PEP updates only hold if $\big( \alpha \nabla_{\cavmeanfull_n}^2\LLbar(\cavmeanfull_{k,n}, \cavcovfull_{k,n,n}) + {\cavcovfull}^{-1}_{k,n,n} \big)$ is PSD, which is not guaranteed to be the case. However, in practice, we find the method greatly reduces the chance of the update resulting in a negative-definite covariance. We find that most of the practical issues involved with these PSD constraints for both VI and PEP can be alleviated by setting the learning rate $\rho$ to a small value. Reducing the power, $\alpha$, in the PEP case also improves stability.

\section{Examples, Experiments, and Connections to Related Work}\label{sec:examples}

As we have shown, all approximate inference schemes can be cast as updates to local approximate likelihoods, \cref{eq:local-update}, combined with conjugate global parameter updates, \cref{eq:global-update}. For this reason, the choice of approximate inference scheme is completely independent of the choice of model, and of the approach used to compute the global updates. To illustrate this point, we now demonstrate how to apply the inference schemes when the latent variable $\vfunc$ is characterised by a Gaussian process, a sparse Gaussian process, or a state space model. The full algorithms are described in \cref{sec:algos}.

\subsection{Gaussian Processes} \label{sec:gp}

A Gaussian process \citep[GP,][]{rasmussen2003gaussian} is a distribution over functions, which states that realisations of a function, $f(x)$, at any finite collection of inputs, $\MX\in\R^{N\times D_\vx}$, is jointly Gaussian distributed, $\vfunc = f(\MX) \sim \N(f(\MX) \mid \priormean=\mu(\MX), \priorcov=\kappa(\MX, \MX))$. A GP prior, written $f(x)\sim \GP(\mu(x), \kappa(x,x'))$, is characterised by a mean function, $\mu(x)$, and a covariance function, $\kappa(x, x')$, and sophisticated domain knowledge can be incorporated into $\mu$ and $\kappa$, which may have some hyperparameters, $\vtheta$, associated with them.

If a GP prior is combined with a Gaussian likelihood, $p(\vy \mid \vfunc)=\N(\vy \mid \vfunc, \likcov)$, then the application of Bayes' rule allows us to obtain the posterior distribution, $p(\vfunc \mid \vy)\propto p(\vfunc)\, p(\vy \mid \vfunc)$, whose mean and covariance are given by (assuming the prior has zero mean, $\vmu=\bm{0}$),
\begin{gather} \label{eq:gp-posterior}
	\begin{aligned}
		\postmean &= \priorcov (\priorcov + \likcov)^{-1} \vy , \\
		\postcov &=	\priorcov - \priorcov (\priorcov + \likcov)^{-1} \priorcov .
	\end{aligned}	
\end{gather}
When $p(\vy \mid \vfunc)$ is non-Gaussian, the approximate inference schemes presented above can be used by replacing the true likelihood with an \emph{approximate likelihood}, $t(\vfunc)=\N(\vfunc \mid \likmean, \likcov)$. In this case, by setting $\vy=\likmean$, the above equations are equivalent to, but slightly more stable than, the updates given in \cref{eq:global-update}. Additionally, GPs allow for predictions to be made at `test' locations, $\MX^*$, by exploiting the marginalisation and conditional properties of Gaussian densities. In \cref{sec:experiments} we also consider the case where $f$ is vector-valued with $\fdim$ elements: $f(x):\R^{D_\vx}\rightarrow\R^{\fdim}$.

Inference in non-conjugate GP models therefore involves the iterative application of a local update scheme for $t(\vfunc)$, followed by \cref{eq:local-update}, and \cref{eq:gp-posterior} with $\vy=\likmean$. \cref{alg:gp} gives the full algorithm. These iterations can be alternated with updates to the hyperparameters, $\vtheta$, via gradient-based optimisation of the model energy (see \cref{sec:bayes-newton}). If the VI updates of \cref{eq:vi-target} are used, we recover a natural gradient version of the \emph{variational GP} \citep{opper2009variational}. The PEP updates of \cref{eq:pep-target} recover the approach of \citet{minka2001family}. Newton's method, \cref{eq:newton-target}, results in the Laplace approach described in \citet{rasmussen2003gaussian}. PL, \cref{eq:pl-target}, recovers \citet{garcia2019gaussian}.

\subsection{Sparse Gaussian Processes} \label{sec:sparse-gp}

The global update in \cref{eq:gp-posterior} has $\BO(N^3\fdim^3)$ computational scaling, which can be prohibitive. A common approach to reducing this scaling is the \emph{sparse GP} \citep{csato2002sparse}. Sparse GPs introduce a set of \emph{inducing inputs}, $\MZ\in\R^{M\times D_\vx}$, where $M<N$. The \emph{inducing variables}, $\vu=f(\MZ)$, have prior $p(\vu)=\N(\vu \mid \priormean_\vu=\mu(\MZ), \priorcov_{\vu\vu}=\kappa(\MZ,\MZ))$.

The posterior over the inducing variables, $q(\vu)=\N(\vu \mid \postmean_\vu, \postcov_\vu)$, is stored in memory, from which the approximate posterior marginals can be obtained by conditioning: $q(\vfunc_n)=\int p(\vfunc_n \mid \vu) q(\vu)\, \mathrm{d}\vu$, where the conditional is Gaussian: $p(\vfunc_n \mid \vu)=\N(\vfunc_n \mid \MW_{\vfunc_n\!\vu}\vu , \MK_{n,n} - \MK_{\vfunc_n\!\vu} \MW_{\vfunc_n\!\vu}^\T)$, for $\MW_{\vfunc_n\!\vu}=\MK_{\vfunc_n\!\vu}\MK_{\vu\vu}^{-1}$, and $\MK_{\vfunc_n\!\vu}=\kappa(\MX_n,\MZ)$. This leads to computational savings because $\vfunc_i$ and $\vfunc_j$ are now conditionally independent given $\vu$. The approximate likelihoods are also redefined to be functions of $\vu$,
\begin{equation}
	t(\vu)= \prod_{n=1}^N t_n(\vu) .
\end{equation}
In order to update the parameters of $t_n(\vu)$ we must compute the gradients of the surrogate target (see \cref{tab:targets}), but where the posterior marginal, $q(\vfunc_n)$, is computed via the conditional. For example, the sparse VI surrogate target becomes $\E_{q(\vu)} [\E_{p(\vfunc_n \mid \vu)}[\log p(\vy_n \mid \vfunc_n)]]$ and we compute its gradients with respect to the mean of $t_n(\vu)$. Fortunately, \emph{for all} inference schemes this quantity amounts to applying the standard update rules to $t(\vfunc_n)$ followed by a deterministic mapping: recalling that $\liknatone$, $\liknattwo$, are the natural parameters of $t(\vfunc)$, the corresponding natural parameters of $t(\vu)=z_\vu\N(\vu\mid \likmean_\vu, \likcov_\vu)$ are
\begin{gather}
	\begin{aligned}
		\liknatone_\vu &=  \MW_{\vfunc\vu}^\T \liknatone , \\
		\liknattwo_\vu &= \MW_{\vfunc\vu}^\T \liknattwo \MW_{\vfunc\vu} \,.
	\end{aligned}	
\end{gather}
Note that $\liknattwo_\vu\in\R^{M\fdim\times M\fdim}$ is a dense matrix, whereas $\liknattwo\in\R^{N\fdim\times N\fdim}$ is block-diagonal. To update the inducing posterior we now apply a modified version of \cref{eq:gp-posterior},
\begin{gather} \label{eq:sparse-gp-posterior}
	\begin{aligned}
		\postmean_\vu &= \priorcov_{\vu\vu} (\priorcov_{\vu\vu} + \likcov_\vu)^{-1} \likmean_\vu , \\
		\postcov_\vu &=	\priorcov_{\vu\vu} - \priorcov_{\vu\vu} (\priorcov_{\vu\vu} + \likcov_\vu)^{-1} \priorcov_{\vu\vu} .
	\end{aligned}	
\end{gather}
Updates to the local factor, $t(\vfunc_n)$, require the posterior marginal $q(\vfunc_n)=\N(\vfunc_n \mid \postmean_n, \postcov_{n,n})$, which is given by
\begin{gather} \label{eq:sparse-gp-conditional}
	\begin{aligned}
		\postmean_n &= \MW_{\vfunc_n\!\vu} \postmean_\vu  , \\
		\postcov_{n,n} &= \priorcov_{n,n} - \MW_{\vfunc_n\!\vu}\MK_{\vfunc_n\!\vu}^\T + \MW_{\vfunc_n\!\vu} \postcov_\vu \MW_{\vfunc_n\!\vu}^\T .
	\end{aligned}	
\end{gather}
This algorithm has $\BO(N M^2 \fdim^3)$ dominant computational scaling, and can be combined with any of the inference methods presented above by choosing the appropriate update rule for $t(\vfunc_n)$. The full algorithm is given in \cref{alg:sgp}. Use of the VI updates of \cref{eq:vi-target} leads to the method of \citet{adam2021dual}, which they show to be an improved version of natural gradient inference for the \emph{sparse variational GP} \citep{hensman2015scalable, salimbeni2018natural}. The PEP updates of \cref{eq:pep-target} lead to sparse PEP \citep{bui2017unifying}.

If the local factors, $t(\vfunc_n)$, are stored in memory, then the overall memory requirement scales as $\BO(N\fdim+M^2\fdim^2)$. However, for all algorithms apart from PEP (where the local factors must be kept in order to compute the cavities), it's possible to instead store only $t(\vu)$, leading to memory scaling of $\BO(M^2\fdim^2)$. The inducing inputs, $\MZ$, can be treated as hyperparameters and optimised based on the model energy. Crucially, a \emph{stochastic} version of the sparse algorithm can be obtained by updating only a subset of the local factors on each iteration, which is particularly efficient since the marginals of $q(\vfunc)$ can be computed independently given $q(\vu)$. In this case, the parameters for each factor are `tied': the likelihood contribution for each data point is simply given by $t_n(\vu)=t^{\nicefrac{N-1}{N}}(\vu)$. This approach is used with VI in \citet{adam2021dual}, and when used with PEP it leads to \emph{stochastic PEP} \citep{li2015stochastic} which also has memory scaling $\BO(M^2\fdim^2)$.

\subsubsection{The Sparse GP Energy}

The GP energy can also be efficiently approximated via the sparse model. The sparse variational free energy (see \cref{sec:vfe}) is given by
\begin{equation} \label{eq:sparse-vfe}
	\text{VFE}(q(\vu)) = -\sum_{n=1}^N \E_{q(\vu)}[\E_{p(\vfunc_n \mid \vu)} [\log p(\vy_n \mid \vfunc_n)]] + \E_{q(\vu)}[\log \N(\vu \mid \likmean_\vu, \likcov_\vu)] - \log \mathcal{Z}_\vu ,
\end{equation}
with $\log \mathcal{Z}_\vu=\int p(\vu) \N(\vu \mid \likmean_\vu, \likcov_\vu) \, \mathrm{d} \vu$. Replacing the expectations with point estimates in a similar way to \cref{eq:laplace-energy} leads to a sparse version of the Laplace energy.

The sparse PEP energy is

\begin{multline} \label{eq:sparse-pep-energy}
	\text{PEPE}(q(\vu)) = -\frac{1}{\alpha} \sum_{n=1}^N \log \E_{q_n^{\cav }(\vu)} [ \E_{p(\vfunc_n \mid \vu)} [ p^\alpha(\vy_n \mid \vfunc_n) ]] \\ + \frac{1}{\alpha}\log\E_{q^{\cav }(\vu)}[\N^\alpha(\vu \mid \likmean_\vu, \likcov_\vu)]  - \log \mathcal{Z}_\vu , 
\end{multline}
where $q^{\cav }(\vu)=q(\vu)/t^\alpha(\vu)$ is the `global' cavity, and $q_n^{\cav }(\vu)=q(\vu)/t^\alpha_n(\vu)$ is the cavity associated with data point $\vy_n$. This result is arrived at using a similar approach to \cref{sec:pep-energy}: we set $z_\vu$ such that the zero-th moment (\ie, the log normaliser) of the approximate posterior matches the (product of) zero-th moments of the tilted distributions,
\begin{equation} \label{eq:sparse-pep-z}
	z_\vu^\alpha \, \E_{q^{\cav }(\vu)} [ \N^\alpha(\vu \mid \likmean_\vu, \likcov_\vu) ] =  \prod_{n=1}^{N} \E_{q_n^{\cav }(\vu)} [ \E_{p(\vfunc_n \mid \vu)} [ p^\alpha(\vy_n \mid \vfunc_n) ]] ,\\
\end{equation}
and then rearrange in a similar fashion to \cref{eq:pep-z2}.

Again using the property in \cref{eq:pep-limit}, we can see that the PEP energy is equal to the VFE in the limit as $\alpha\rightarrow 0$. Since the sparse algorithm presented above is identical for both methods, we can conclude the following:
\begin{remark}	
	 The connection between PEP and natural gradient VI holds in the sparse GP case: sparse PEP ($\alpha \rightarrow 0$) is equivalent to natural gradient sparse VI.
\end{remark}

\subsection{State Space Models and Markovian Gaussian Processes} \label{sec:markov-gp}

All of the listed inference schemes can be used to perform inference in the discrete-time state space model with linear Gaussian dynamics of the following form,
\begin{gather} \label{eq:state-space-model}
	\begin{aligned}
		\state_{n+1} &= \MA_n \state_n + \vq_n \, , \quad\quad \vq_n \sim \N(\bm{0}, \MQ_n) , \\
		\vy_n \mid \state_n &\sim p(\vy_n \mid \bar{\MH} \state_n) ,
	\end{aligned}	
\end{gather}
where $\state_n\in\R^{S\times 1}$ is the Gaussian distributed state vector, $\MA_n\in\R^{S\times S}$ is the transition matrix, and $\MQ_n\in\R^{S\times S}$ is the process noise. $\bar{\MH}\in\R^{\fdim\times S}$ is the measurement matrix such that $\vfunc_n=\bar{\MH}\state_n$. In essence, the Gaussian prior in \cref{eq:model} has been replaced by the linear Gaussian state space model on the first line of \cref{eq:state-space-model}. Therefore inference again simply involves replacing the true likelihood, $p(\vy_n \mid \bar{\MH} \state_n)$, with the approximate likelihood, $t(\vfunc_n)$, after which the approximate posterior over the state, $q(\state)\approx p(\state \mid \vy)$, is given by application of the linear Kalman filter followed by the Rauch--Tung--Striebel smoother \citep{sarkka2013bayesian}. That is, the global posterior update, \cref{eq:global-update}, is replaced with linear filtering and smoothing. The full algorithm is given in \cref{alg:mgp}.

\subsubsection{Markovian Gaussian Processes}

Consider again the GP model in \cref{sec:gp}. If the inputs are vector valued and ordered, $\MX\in\R^{N\times 1}=[x_1 , \dots, x_N]^\T$, for example if they represent sequential time steps, then for many common covariance functions the GP can be rewritten as a linear time-invariant stochastic differential equation \citep{Hartikainen:2013},
\begin{equation}
	\mathrm{d}\state(x) = \MF \, \state(x) \,\mathrm{d}x + \ML\, \mathrm{d}\vbeta(x) ,
\end{equation}
where $\MF$ is the feedback matrix and $\mathrm{d}\vbeta(x)$ has spectral density $\MQ_c$. $\MF$ and $\MQ_c$ are determined by the GP covariance function such that the model exhibits similar covariance properties to the standard GP \citep[see][for details]{sarkka2019applied}. There is a linear relationship between the state, $\state$, and the function, $f$, characterised by the measurement matrix, $f(x_n)=\bar{\MH}\state(x_n)$. The discrete-time solution to this SDE is given by \cref{eq:state-space-model} where
\begin{equation}
	\MA_{n} = \MPhi( \MF \Delta_n) \quad \text{and} \quad
	\MQ_{n} =  \int_{0}^{\Delta_n} \MPhi(\Delta_n-\tau)\,\ML\, \MQ_c\, \ML^{\top}\, \MPhi(\Delta_n-\tau)^{\top} \mathrm{d} \tau ,
\end{equation}
for step size $\Delta_n=x_{n+1}-x_n$, where $\MPhi(\cdot)$ is the matrix exponential. The initial state is $\state_0 \sim \N(\bm{0}, \MP_0)$, where $\MP_0$ is the solution to the Lyapunov equation, $\MF \MP_0 + \MP_0\MF^\T + \ML \MQ_c \ML^\T=\bm{0}$. For many covariance functions, $\MA_n$, $\MQ_n$, and $\MP_0$ can be computed in closed form. Once constructed, inference can again proceed by application of linear filtering and smoothing, which will return the exact same posterior as that given by the approach in \cref{sec:gp}. This algorithm scales linearly in the number of data points, $\BO(S^3N)$.

The use of the VI updates in conjunction with filtering and smoothing gives the approach of \citet{chang2020fast}, whilst the use of PEP was explored in \citet{wilkinson2020state}. The Newton/Laplace approach was first applied to Markovian GPs in \citet{nickisch2018state}. PL was initially derived as an approach for inference in state space models, and generalises the iterated nonlinear Kalman smoothers such as the extended and unscented smoothers. The Markovian approach can be extended to spatio-temporal data with more than one input dimension via the use of infinite-dimensional filtering methods \citep{Sarkka+Solin+Hartikainen:2013, hamelijnck2021spatio, tebbutt2021combining}. The sparse and Markovian approaches can also be combined to further reduce the computational scaling \citep{wilkinson2021sparse}.

\subsubsection{The Markovian GP Energy}

The GP model energy can also be computed efficiently when using the Markovian approach. Consider the energy functions for VI, Laplace, and PEP, given by \cref{eq:vfe}, \cref{eq:laplace-energy} and \cref{eq:pep-energy} respectively. In each case, the first two terms can be computed as usual, since they only require the marginals, $q(\vfunc_n)$, which are given by filtering and smoothing. The final term, $\log \mathcal{Z}=\log \int p(\vfunc) \prod_{n=1}^N \N(\vfunc_n \mid \likmean_n, \likcov_{n,n}) \, \mathrm{d} \vfunc$, can be computed sequentially during the forward filter as,
\begin{equation}
	\log \mathcal{Z} = \sum_{n=1}^N \log \int p(\state(x_n) \mid \likmean_{1:n-1}) \, \N(\likmean_n \mid \bar{\MH}\state(x_n) , \likcov_{n,n}) \,\mathrm{d}\state(x_n) ,
\end{equation}
where $p(\state(x_n) \mid \likmean_{1:n-1})$ is the predictive filtering distribution. Therefore the energy for all inference schemes can also be computed in $\BO(S^3N)$.

\begin{figure*}[t!]
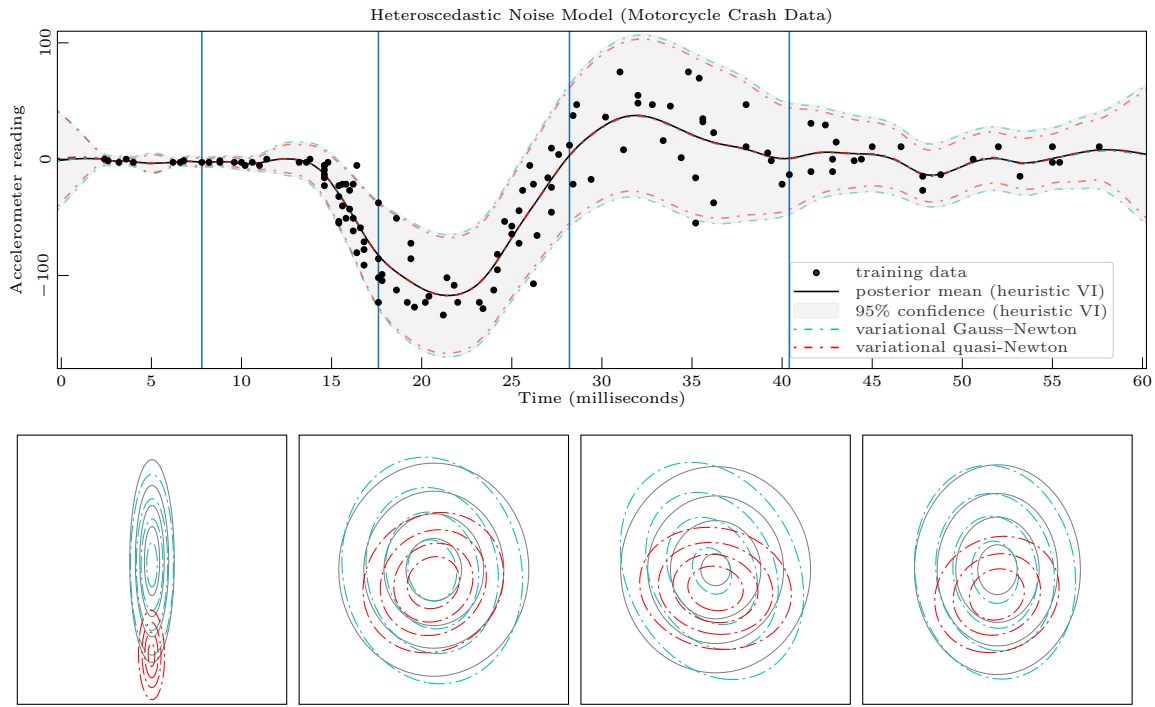

	\centering\tiny
	\pgfplotsset{yticklabel style={rotate=90}, ylabel style={yshift=0pt},scale only axis,axis on top,title style={yshift=-6pt}, xlabel style={yshift=3pt}}
	\pgfplotsset{legend style={inner xsep=1pt, inner ysep=1pt, row sep=0pt},legend style={at={(0.98,0.95)},anchor=north east},legend style={rounded corners=1pt}}
	\setlength{\figurewidth}{0.95\textwidth}
	\setlength{\figureheight}{.31\figurewidth}
	\begin{subfigure}[t]{\textwidth}
		\input{./fig/hsced-demo.tex}
	\end{subfigure}\\
	\pgfplotsset{ticks=none}
	\setlength{\figurewidth}{0.235\textwidth}
	\setlength{\figureheight}{\figurewidth}
	\begin{subfigure}[t]{0.24\textwidth}
		\input{./fig/hsced-contour0.tex}
	\end{subfigure}
	\begin{subfigure}[t]{0.24\textwidth}
		\input{./fig/hsced-contour1.tex}
	\end{subfigure}
	\begin{subfigure}[t]{0.24\textwidth}
		\input{./fig/hsced-contour2.tex}
	\end{subfigure}
	\begin{subfigure}[t]{0.24\textwidth}
		\input{./fig/hsced-contour3.tex}
	\end{subfigure}\\
	\caption{Example heteroscedastic noise model results. The top figure compares the posterior obtained for the motorcycle crash data set when using heuristic VI, variational Gauss--Newton and variational quasi-Newton. All methods obtain similar mean values, but different covariances. The contour plots show the marginal posterior covariance between $f_1$ (x-axis) and $f_2$ (y-axis, the noise standard deviation GP) at the time points marked by vertical lines in the top figure. The heuristic method (grey) assumes posterior independence between the two latents, variational Gauss--Newton (cyan) captures small amounts of cross-covariance, whilst variational quasi-Newton (red) captures more significant cross-covariance.} \label{fig:hsced-plot}
\end{figure*}

\subsection{Experiments}\label{sec:experiments}

To evaluate our proposed methods, we consider three case studies. These include likelihood models that are a nonlinear function of multiple latent Gaussian processes, and which consistently result in non-PSD covariances when inference is applied na\"ively. These models can be seen as instances of \emph{chained GPs} \citep{saul2016chained}, with the distinction that \citet{saul2016chained} assume independence between the latent processes to make the model tractable and stable, which amounts to a similar approach to the heuristic method discussed in \cref{sec:heuristic}. We do not assume independence between the posterior processes, and we demonstrate that this can improve the inference result. Such an approach can become computationally prohibitive when the number of latent processes is large, although the sparse algorithm given in \cref{sec:sparse-gp} is applicable in all cases, and the Markovian approach of \cref{sec:markov-gp} is applicable whenever the inputs are one-dimensional. We compute the negative log predictive density (NLPD) of the test data as the main performance metric. We set the EP power to $\alpha=0.5$. In almost all cases, we find that the methods based on Riemannian gradients result in non-PSD covariances when the method nears its optima. Therefore in the reported results we take the last stable step before the algorithm fails as the final result.

We have also added a first-order variational inference method as a baseline to each experiment. This is achieved by training the approximate likelihood mean and covariance via gradient-base optimisation of the VFE using the Adam optimiser with a learning rate of $0.1$ (we empirically found this to give the best performance and convergence). This is equivalent to the variational-GP method of \citet{opper2009variational}. As expected, the first-order method converges more slowly than the other methods.

\begin{figure*}[t!]
	\centering\tiny
	\pgfplotsset{yticklabel style={rotate=90}, ylabel style={yshift=0pt},scale only axis,axis on top,title style={yshift=-6pt}, xlabel style={yshift=3pt}}
	\pgfplotsset{legend style={inner xsep=1pt, inner ysep=1pt, row sep=0pt},legend style={at={(0.98,0.95)},anchor=north east},legend style={rounded corners=1pt}}
	\setlength{\figurewidth}{.47\textwidth}
	\setlength{\figureheight}{.34\figurewidth}
	\begin{subfigure}[t]{.49\textwidth}
		\raggedright
		% This file was created with tikzplotlib v0.9.14.
\begin{tikzpicture}

\definecolor{color0}{rgb}{0.12156862745098,0.466666666666667,0.705882352941177}
\definecolor{color1}{rgb}{1,0.498039215686275,0.0549019607843137}
\definecolor{color2}{rgb}{0.172549019607843,0.627450980392157,0.172549019607843}
\definecolor{color3}{rgb}{0.83921568627451,0.152941176470588,0.156862745098039}

\begin{axis}[
height=\figureheight,
legend cell align={left},
legend style={fill opacity=0.8, draw opacity=1, text opacity=1, draw=white!80!black},
tick pos=left,
title={Training Loss (Newton)},
width=\figurewidth,
x grid style={white!69.0196078431373!black},
xlabel={iteration number},
xmin=-24.5, xmax=514.5,
xtick style={color=black},
y grid style={white!69.0196078431373!black},
ymin=62, ymax=120,
ytick style={color=black}
]
\addplot [very thick, color0, dashed]
table {%
0 94.6017210357658
10 63.9506019142359
20 66.1247748171824
30 66.774485439138
40 66.9211034089153
50 66.9501647134724
60 66.9559036278152
70 66.9570648308675
80 66.9573053261316
90 66.9573560514724
100 66.9573669012473
110 66.9573692475833
120 66.9573697595551
130 66.9573698721159
140 66.9573698970262
150 66.9573699025728
160 66.9573699038139
170 66.9573699040932
180 66.9573699041567
190 66.9573699041698
200 66.9573699041729
210 66.9573699041754
220 66.9573699041758
230 66.9573699041756
240 66.9573699041757
250 66.9573699041758
260 66.9573699041757
270 66.9573699041758
280 66.9573699041758
290 66.9573699041756
300 66.9573699041757
310 66.9573699041757
320 66.9573699041757
330 66.9573699041758
340 66.9573699041757
350 66.9573699041758
360 66.9573699041758
370 66.9573699041758
380 66.9573699041758
390 66.9573699041759
400 66.9573699041759
410 66.9573699041757
420 66.9573699041758
430 66.9573699041758
440 66.9573699041757
450 66.9573699041757
460 66.9573699041758
470 66.9573699041757
480 66.9573699041758
490 66.9573699041758
};
\addlegendentry{heuristic Newton}
\addplot [very thick, color1, dashed]
table {%
0 185.504834885082
10 80.9109903590356
20 73.7305215607612
30 76.9503563328751
40 80.1048653240769
50 82.5907343925605
60 83.8551705890017
70 84.0433643808112
80 84.2444156623629
90 84.4180381984239
100 84.5712994715693
110 84.7471199661863
120 84.9038583611272
130 84.9038583611272
140 84.9038583611272
150 84.9038583611272
160 84.9038583611272
170 84.9038583611272
180 84.9038583611272
190 84.9038583611272
200 84.9038583611272
210 84.9038583611272
220 84.9038583611272
230 84.9038583611272
240 84.9038583611272
250 84.9038583611272
260 84.9038583611272
270 84.9038583611272
280 84.9038583611272
290 84.9038583611272
300 84.9038583611272
310 84.9038583611272
320 84.9038583611272
330 84.9038583611272
340 84.9038583611272
350 84.9038583611272
360 84.9038583611272
370 84.9038583611272
380 84.9038583611272
390 84.9038583611272
400 84.9038583611272
410 84.9038583611272
420 84.9038583611272
430 84.9038583611272
440 84.9038583611272
450 84.9038583611272
460 84.9038583611272
470 84.9038583611272
480 84.9038583611272
490 84.9038583611272
};
\addlegendentry{Riemannian grads Newton}
\addplot [very thick, color2]
table {%
0 110.938629658403
10 63.4113406255307
20 64.934906784698
30 66.3395246662522
40 66.9199463558037
50 67.1859203483545
60 67.3131100868088
70 67.374188993577
80 67.403397782074
90 67.417309679196
100 67.4239186611278
110 67.4270537336126
120 67.4285397471317
130 67.4292438305689
140 67.4295773628833
150 67.4297353450913
160 67.4298101718716
170 67.4298456120138
180 67.4298623973007
190 67.4298703471648
200 67.4298741123755
210 67.4298758956518
220 67.4298767402453
230 67.4298771402581
240 67.4298773297146
250 67.4298774194426
260 67.4298774619391
270 67.4298774820677
280 67.4298774916008
290 67.4298774961141
300 67.4298774982526
310 67.4298774992652
320 67.4298774997455
330 67.4298774999728
340 67.4298775000806
350 67.4298775001306
360 67.4298775001551
370 67.4298775001662
380 67.4298775001723
390 67.4298775001743
400 67.4298775001754
410 67.4298775001757
420 67.4298775001759
430 67.4298775001761
440 67.4298775001761
450 67.4298775001762
460 67.4298775001761
470 67.4298775001761
480 67.4298775001761
490 67.4298775001761
};
\addlegendentry{Gauss-Newton}
\addplot [very thick, color3]
table {%
0 187.863370299047
10 162.259518250551
20 131.647389296858
30 119.631526598162
40 105.254958064624
50 79.7308165504288
60 68.1272083021333
70 74.6753270969991
80 75.5823816570908
90 75.7357115054368
100 76.3714643055354
110 77.7967608648784
120 76.4711829417277
130 74.7847247607988
140 73.8704918364601
150 73.6626737671975
160 73.5000083740268
170 73.4869293737514
180 73.4869293737455
190 73.4869293737523
200 73.4869293737487
210 73.4869293737485
220 73.4869293737516
230 73.4869293737507
240 73.4869293737524
250 73.4869293737468
260 73.4869293737494
270 73.4869293737523
280 73.4869293737491
290 73.4869293737537
300 73.4869293737507
310 73.4869293737505
320 73.4869293737491
330 73.4869293737472
340 73.4869293737504
350 73.4869293737519
360 73.4869293737456
370 73.4869293737498
380 73.4869293737514
390 73.4869293737506
400 73.4869293737475
410 73.4869293737493
420 73.4869293737481
430 73.486929373743
440 73.4869293737504
450 73.4869293737488
460 73.4869293737536
470 73.4869293737471
480 73.4869293737448
490 73.4869293737483
};
\addlegendentry{quasi-Newton}
\end{axis}

\end{tikzpicture}
	\end{subfigure}
	\hspace*{\fill}
	\begin{subfigure}[t]{.49\textwidth}
		\raggedleft
		% This file was created with tikzplotlib v0.9.14.
\begin{tikzpicture}

\definecolor{color0}{rgb}{0.12156862745098,0.466666666666667,0.705882352941177}
\definecolor{color1}{rgb}{1,0.498039215686275,0.0549019607843137}
\definecolor{color2}{rgb}{0.172549019607843,0.627450980392157,0.172549019607843}
\definecolor{color3}{rgb}{0.83921568627451,0.152941176470588,0.156862745098039}

\begin{axis}[
height=\figureheight,
tick pos=left,
title={Test NLPD (Newton)},
width=\figurewidth,
x grid style={white!69.0196078431373!black},
xlabel={iteration number},
xmin=-24.5, xmax=514.5,
xtick style={color=black},
y grid style={white!69.0196078431373!black},
ymin=0.31, ymax=1,
ytick style={color=black}
]
\addplot [very thick, color0, dashed]
table {%
0 0.942549931649287
10 0.373818867868403
20 0.37623829429658
30 0.381027523856301
40 0.382102469609124
50 0.382323016742215
60 0.382370497090926
70 0.382380892754627
80 0.382383180776538
90 0.382383685959312
100 0.382383797860728
110 0.382383822736923
120 0.382383828288956
130 0.382383829533365
140 0.382383829813514
150 0.382383829876874
160 0.382383829891262
170 0.382383829894549
180 0.382383829895305
190 0.382383829895469
200 0.382383829895507
210 0.382383829895531
220 0.382383829895534
230 0.382383829895534
240 0.382383829895534
250 0.382383829895534
260 0.382383829895534
270 0.382383829895534
280 0.382383829895534
290 0.382383829895534
300 0.382383829895534
310 0.382383829895534
320 0.382383829895534
330 0.382383829895534
340 0.382383829895534
350 0.382383829895534
360 0.382383829895534
370 0.382383829895534
380 0.382383829895534
390 0.382383829895534
400 0.382383829895534
410 0.382383829895534
420 0.382383829895534
430 0.382383829895534
440 0.382383829895534
450 0.382383829895534
460 0.382383829895534
470 0.382383829895534
480 0.382383829895534
490 0.382383829895534
};
\addplot [very thick, color1, dashed]
table {%
0 1.50526676106506
10 0.584723368760057
20 0.390355500268957
30 0.386047929699894
40 0.386945952829159
50 0.386682259424048
60 0.386995184871831
70 0.387231814528306
80 0.387413002215486
90 0.38755130052757
100 0.387656021664604
110 0.387734061663259
120 0.387792588025028
130 0.387792588025028
140 0.387792588025028
150 0.387792588025028
160 0.387792588025028
170 0.387792588025028
180 0.387792588025028
190 0.387792588025028
200 0.387792588025028
210 0.387792588025028
220 0.387792588025028
230 0.387792588025028
240 0.387792588025028
250 0.387792588025028
260 0.387792588025028
270 0.387792588025028
280 0.387792588025028
290 0.387792588025028
300 0.387792588025028
310 0.387792588025028
320 0.387792588025028
330 0.387792588025028
340 0.387792588025028
350 0.387792588025028
360 0.387792588025028
370 0.387792588025028
380 0.387792588025028
390 0.387792588025028
400 0.387792588025028
410 0.387792588025028
420 0.387792588025028
430 0.387792588025028
440 0.387792588025028
450 0.387792588025028
460 0.387792588025028
470 0.387792588025028
480 0.387792588025028
490 0.387792588025028
};
\addplot [very thick, color2]
table {%
0 0.984609382560311
10 0.41356346864779
20 0.358229750988971
30 0.370599387997344
40 0.385479699545568
50 0.39484638334119
60 0.399840208394259
70 0.402333744887649
80 0.403543847937683
90 0.40412354560543
100 0.40439958065815
110 0.404530648806604
120 0.404592799950754
130 0.404622252707693
140 0.404636205859662
150 0.404642815180087
160 0.404645945667264
170 0.404647428366666
180 0.404648130610094
190 0.404648463207708
200 0.404648620732554
210 0.40464869533934
220 0.404648730674517
230 0.404648747409901
240 0.404648755336106
250 0.404648759090091
260 0.404648760868047
270 0.404648761710126
280 0.404648762108952
290 0.40464876229783
300 0.404648762387293
310 0.404648762429662
320 0.404648762449732
330 0.404648762459237
340 0.404648762463741
350 0.404648762465869
360 0.404648762466879
370 0.404648762467353
380 0.404648762467584
390 0.404648762467689
400 0.404648762467736
410 0.40464876246776
420 0.404648762467771
430 0.404648762467775
440 0.404648762467776
450 0.404648762467777
460 0.404648762467777
470 0.404648762467777
480 0.404648762467777
490 0.404648762467777
};
\addplot [very thick, color3]
table {%
0 1.5308438635698
10 1.22080575410101
20 0.949068019890152
30 0.861109884843033
40 0.800938532430496
50 0.629088471668863
60 0.362799388152785
70 0.374531681979159
80 0.380259612916808
90 0.378030031599832
100 0.382630111741056
110 0.390192468125916
120 0.390373057279134
130 0.385015908581966
140 0.382159318759953
150 0.380574571264602
160 0.379964799158598
170 0.379893478657806
180 0.379893479169203
190 0.379893479577063
200 0.379893479867557
210 0.379893480085913
220 0.379893480259717
230 0.379893480405564
240 0.379893480533501
250 0.379893480649637
260 0.379893480757739
270 0.379893480860133
280 0.379893480958269
290 0.379893481053091
300 0.379893481145217
310 0.379893481235014
320 0.379893481322782
330 0.379893481408687
340 0.379893481492902
350 0.379893481575464
360 0.379893481656482
370 0.379893481736009
380 0.379893481814081
390 0.379893481890759
400 0.37989348196605
410 0.379893482040005
420 0.379893482112647
430 0.379893482184005
440 0.37989348225408
450 0.379893482322931
460 0.379893482390559
470 0.379893482457007
480 0.379893482522269
490 0.379893482586377
};
\end{axis}

\end{tikzpicture} 
	\end{subfigure}\\
	\vspace{-1em}
	\begin{subfigure}[t]{.49\textwidth}
		\raggedright
		% This file was created with tikzplotlib v0.9.14.
\begin{tikzpicture}

\definecolor{color0}{rgb}{0.12156862745098,0.466666666666667,0.705882352941177}
\definecolor{color1}{rgb}{1,0.498039215686275,0.0549019607843137}
\definecolor{color2}{rgb}{0.172549019607843,0.627450980392157,0.172549019607843}
\definecolor{color3}{rgb}{0.83921568627451,0.152941176470588,0.156862745098039}

\begin{axis}[
height=\figureheight,
legend cell align={left},
legend style={fill opacity=0.8, draw opacity=1, text opacity=1, draw=white!80!black},
tick pos=left,
title={Training Loss (VI)},
width=\figurewidth,
x grid style={white!69.0196078431373!black},
xlabel={iteration number},
xmin=-24.5, xmax=514.5,
xtick style={color=black},
y grid style={white!69.0196078431373!black},
ymin=62, ymax=120,
ytick style={color=black}
]
\addplot [very thick, color0, dashed]
table {%
0 125.95054334865
10 84.8376370397111
20 76.6203655922732
30 75.1582904915146
40 74.8533737956721
50 74.7869809690654
60 74.7716704097895
70 74.7678875176347
80 74.766886362495
90 74.7666049287764
100 74.7665220544379
110 74.7664968291832
120 74.7664889822831
130 74.7664865095249
140 74.7664857256203
150 74.7664854772011
160 74.7664853990927
170 74.7664853750046
180 74.7664853678752
190 74.7664853659492
200 74.7664853655451
210 74.7664853655399
220 74.7664853656105
230 74.7664853656746
240 74.7664853657189
250 74.7664853657467
260 74.7664853657634
270 74.7664853657732
280 74.7664853657788
290 74.7664853657821
300 74.766485365784
310 74.7664853657851
320 74.7664853657856
330 74.766485365786
340 74.7664853657863
350 74.7664853657863
360 74.7664853657863
370 74.7664853657864
380 74.7664853657864
390 74.7664853657865
400 74.7664853657864
410 74.7664853657865
420 74.7664853657865
430 74.7664853657864
440 74.7664853657864
450 74.7664853657865
460 74.7664853657865
470 74.7664853657864
480 74.7664853657864
490 74.7664853657864
};
\addlegendentry{heuristic VI}
\addplot [very thick, color1, dashed]
table {%
0 240.153333134871
10 146.867948884316
20 98.2888655570521
30 82.2280066359114
40 78.5100467892068
50 78.0325141669307
60 78.3436232189339
70 78.7431292109904
80 78.875603520327
90 79.1170362544749
100 79.2465119082602
110 79.1288544999222
120 79.0146692923671
130 79.0134255912552
140 79.080198992986
150 79.1870710134736
160 79.3207721499611
170 79.3552332173341
180 79.5040139337997
190 79.537127592231
200 79.5702774392056
210 79.6034980366127
220 79.6354653466826
230 79.6691644708048
240 79.6994816858253
250 79.7333107131522
260 79.7627617795134
270 79.7958420243754
280 79.8256680149994
290 79.8572810249877
300 79.8876352930583
310 79.9174018639939
320 79.9495070065318
330 79.9797544508022
340 80.0096920666594
350 80.041500866226
360 80.0721036842869
370 80.097110426504
380 80.113457859083
390 80.1614454880047
400 80.1753891843835
410 80.1922205802047
420 80.2103601036994
430 80.2214739502758
440 80.2312929734157
450 80.2623126242842
460 80.2566554586506
470 80.240425934126
480 80.2901183870905
490 80.2901183870905
};
\addlegendentry{Riemannian grads VI}
\addplot [very thick, color2]
table {%
0 150.769505859671
10 87.7002782679278
20 77.7117863142287
30 75.4953453254777
40 74.8886285625978
50 74.7285036832507
60 74.6872864164059
70 74.6765714402957
80 74.6737278094915
90 74.6729721439333
100 74.6727841795939
110 74.6727496936349
120 74.6727532311794
130 74.6727631300982
140 74.6727716260681
150 74.6727775757509
160 74.6727814316083
170 74.6727838418221
180 74.6727853212219
190 74.6727862208524
200 74.672786765389
210 74.6727870942989
220 74.672787292822
230 74.6727874126455
240 74.6727874849926
250 74.6727875286959
260 74.6727875551097
270 74.6727875710821
280 74.6727875807448
290 74.6727875865927
300 74.6727875901331
310 74.672787592277
320 74.6727875935756
330 74.6727875943624
340 74.6727875948391
350 74.6727875951279
360 74.672787595303
370 74.6727875954091
380 74.6727875954735
390 74.6727875955124
400 74.6727875955361
410 74.6727875955504
420 74.6727875955591
430 74.6727875955644
440 74.6727875955675
450 74.6727875955695
460 74.6727875955707
470 74.6727875955714
480 74.6727875955717
490 74.6727875955721
};
\addlegendentry{variational Gauss-Newton}
\addplot [very thick, color3]
table {%
0 176.581867085947
10 164.943346523145
20 137.326877406739
30 125.267674421643
40 114.002810693443
50 97.6713245797151
60 84.2874785125375
70 80.0257258867112
80 78.2949229805829
90 77.4080280091218
100 77.2170825060646
110 77.192874058512
120 77.2773057617108
130 77.4209139791086
140 77.4455202883532
150 77.5770126367841
160 77.8128458270438
170 77.7747277980336
180 78.0148431418632
190 78.1417976444471
200 78.2380113794123
210 78.288788884512
220 78.3766167402994
230 78.3876901321215
240 78.3777769968127
250 78.493484650105
260 78.5160748323576
270 78.5015383753365
280 78.5973417299178
290 78.5720759441238
300 78.6086512640867
310 78.64176416495
320 78.6021967656801
330 78.6484481766024
340 78.6284402643612
350 78.6937210767927
360 78.7451618381216
370 78.7021821025187
380 78.7329390105198
390 78.7116236654307
400 78.618500386376
410 78.5917221122531
420 78.6256466809512
430 78.5597936068277
440 78.5846422617994
450 78.6162521039117
460 78.6361293642692
470 78.6313658522862
480 78.5668969707498
490 78.6548796433025
};
\addlegendentry{variational quasi-Newton}
\addplot [very thick, black, dash pattern=on 1pt off 3pt on 3pt off 3pt]
table {%
0 559.514616158418
10 292.384613134885
20 195.099392232684
30 176.649091947635
40 176.927000996841
50 175.839865954668
60 173.220642484927
70 171.181103869958
80 168.688058156863
90 164.287499952806
100 154.377364219027
110 144.262446960777
120 140.455915878809
130 135.147670475069
140 128.861242438556
150 121.534748102092
160 116.320766274359
170 112.663565354967
180 108.834034712508
190 105.299395062961
200 100.767086113136
210 96.7586022822021
220 93.7277128391025
230 89.1153332323034
240 87.1538666284322
250 86.1796095015545
260 85.4485375684822
270 84.2935804920565
280 82.8478098566676
290 81.9644086014419
300 81.432649286017
310 80.9377959943161
320 80.3911945765828
330 80.8014972493421
340 79.8835579471522
350 79.5424073195077
360 79.3288759812175
370 79.1793866418308
380 79.0844042711372
390 79.4094596644951
400 79.883279662344
410 79.6561359781074
420 79.787739448659
430 79.0620090834695
440 79.1503798639673
450 78.8200499416498
460 78.8177803839294
470 78.1930937065707
480 78.3756016010454
490 78.0360626953708
};
\addlegendentry{first-order VI}
\end{axis}

\end{tikzpicture}
	\end{subfigure}
	\hspace*{\fill}
	\begin{subfigure}[t]{.49\textwidth}
		\raggedleft
		% This file was created with tikzplotlib v0.9.14.
\begin{tikzpicture}

\definecolor{color0}{rgb}{0.12156862745098,0.466666666666667,0.705882352941177}
\definecolor{color1}{rgb}{1,0.498039215686275,0.0549019607843137}
\definecolor{color2}{rgb}{0.172549019607843,0.627450980392157,0.172549019607843}
\definecolor{color3}{rgb}{0.83921568627451,0.152941176470588,0.156862745098039}

\begin{axis}[
height=\figureheight,
tick pos=left,
title={Test NLPD (VI)},
width=\figurewidth,
x grid style={white!69.0196078431373!black},
xlabel={iteration number},
xmin=-24.5, xmax=514.5,
xtick style={color=black},
y grid style={white!69.0196078431373!black},
ymin=0.31, ymax=1,
ytick style={color=black}
]
\addplot [very thick, color0, dashed]
table {%
0 1.04507490996386
10 0.637980806539876
20 0.479036368262427
30 0.427656661718154
40 0.407362727226052
50 0.39856198964135
60 0.394471922800135
70 0.392473089522325
80 0.391460553952773
90 0.390934063364445
100 0.390654784588001
110 0.3905043464386
120 0.390422333243829
130 0.390377201021797
140 0.390352181001949
150 0.390338230243699
160 0.390330416176159
170 0.39032602377143
180 0.390323547822588
190 0.390322149088279
200 0.390321357536403
210 0.390320908982684
220 0.39032065452523
230 0.390320510053409
240 0.390320427972717
250 0.390320381314523
260 0.390320354780939
270 0.39032033968684
280 0.390320331098051
290 0.390320326209878
300 0.390320323427395
310 0.390320321843323
320 0.390320320941415
330 0.390320320427862
340 0.390320320135423
350 0.390320319968886
360 0.390320319874044
370 0.39032031982003
380 0.390320319789268
390 0.390320319771747
400 0.390320319761768
410 0.390320319756084
420 0.390320319752847
430 0.390320319751003
440 0.390320319749953
450 0.390320319749355
460 0.390320319749014
470 0.39032031974882
480 0.39032031974871
490 0.390320319748647
};
\addplot [very thick, color1, dashed]
table {%
0 1.58812487301727
10 1.08235942289137
20 0.70986008669961
30 0.52819833302972
40 0.446419538098774
50 0.416308287431628
60 0.403030391295143
70 0.396547864085366
80 0.393226911529614
90 0.391500919306819
100 0.390538555670851
110 0.389930296851062
120 0.38967516410431
130 0.389547958063539
140 0.389342893548613
150 0.389095087482858
160 0.388885575656998
170 0.388712826279869
180 0.388578121321989
190 0.388499444627263
200 0.388433655852154
210 0.388380254207115
220 0.388334916408078
230 0.388296454707596
240 0.388264408755689
250 0.388235447364923
260 0.388212130189797
270 0.388189951091168
280 0.388172100432872
290 0.388155222680848
300 0.388141024375671
310 0.388128341086825
320 0.388116943739728
330 0.388107282250059
340 0.388098299859448
350 0.388090733077802
360 0.388083808805178
370 0.388077800367681
380 0.388072538352548
390 0.388067771461484
400 0.388063452207187
410 0.388059531721873
420 0.388056195366561
430 0.388053211890576
440 0.388050516583342
450 0.388048183997612
460 0.388045993461168
470 0.388043780510298
480 0.388041903230448
490 0.388041903230448
};
\addplot [very thick, color2]
table {%
0 1.37405793255524
10 0.677886616759615
20 0.50288257138915
30 0.441790743357275
40 0.41440275742103
50 0.401430278870174
60 0.395022966573461
70 0.391728473442714
80 0.389976981855215
90 0.389020895417731
100 0.388487980505391
110 0.388185944141707
120 0.388012455798439
130 0.387911728346818
140 0.387852739518773
150 0.387817954548114
160 0.387797328590365
170 0.387785044123507
180 0.387777701801621
190 0.387773300935804
200 0.387770657156376
210 0.387769066050528
220 0.387768107083724
230 0.387767528436256
240 0.387767178949381
250 0.38776696771029
260 0.387766839954543
270 0.387766762651225
280 0.387766715857649
290 0.387766687523396
300 0.387766670362178
310 0.387766659965995
320 0.387766653666989
330 0.387766649849934
340 0.387766647536634
350 0.387766646134551
360 0.387766645284693
370 0.387766644769531
380 0.387766644457239
390 0.387766644267919
400 0.387766644153145
410 0.387766644083563
420 0.387766644041376
430 0.3877666440158
440 0.387766644000293
450 0.387766643990892
460 0.387766643985191
470 0.387766643981736
480 0.38776664397964
490 0.387766643978369
};
\addplot [very thick, color3]
table {%
0 1.41444876523496
10 1.22872151728005
20 0.982534178501148
30 0.869911430373378
40 0.82399078820057
50 0.720776934206165
60 0.578446532233142
70 0.502514907844254
80 0.462764468714663
90 0.436195277678388
100 0.417844217259967
110 0.40342502833908
120 0.393386146883044
130 0.386320273787149
140 0.377948443727436
150 0.372845520124856
160 0.368116387580113
170 0.366400323490057
180 0.364378476887112
190 0.363150974209166
200 0.363069122346061
210 0.3631472281851
220 0.363320366060383
230 0.364201820105861
240 0.364181504235435
250 0.364238267026156
260 0.364194376699261
270 0.364140975582712
280 0.364236393039499
290 0.364311906985767
300 0.364920203840023
310 0.364318487503444
320 0.363924755436652
330 0.363879723325317
340 0.363725204189813
350 0.364690521833316
360 0.363666181699843
370 0.363814657332036
380 0.364902470901352
390 0.365379836882427
400 0.365491792958708
410 0.36467458879665
420 0.364940231906845
430 0.365426113700123
440 0.365094983459374
450 0.364569103478405
460 0.364745670610532
470 0.36462678345043
480 0.364435127201513
490 0.365028322748045
};
\addplot [very thick, black, dash pattern=on 1pt off 3pt on 3pt off 3pt]
table {%
0 1.46389826406696
10 1.44577036627437
20 1.4977999828531
30 1.58425349011803
40 1.63813719412644
50 1.62925257560177
60 1.58949168367121
70 1.55259712491429
80 1.51629442627031
90 1.45136558886416
100 1.33382224968692
110 1.17077207511269
120 1.15544775707169
130 1.08511788413133
140 0.997679220525326
150 0.944664701009361
160 0.873356542178694
170 0.835139367041672
180 0.803646672173904
190 0.775911670389367
200 0.734764306237052
210 0.688346543026843
220 0.643713549479384
230 0.625482083461103
240 0.610272061142421
250 0.598480723165636
260 0.576283170101709
270 0.569208029817714
280 0.550698058564312
290 0.544017301597915
300 0.528390594548729
310 0.52311752614896
320 0.508546556872312
330 0.526367487862296
340 0.489459263958747
350 0.487897703728194
360 0.482608292627086
370 0.47620317271086
380 0.480633545839384
390 0.478621995173765
400 0.499994456731871
410 0.490303845884524
420 0.488786298142029
430 0.480927314629223
440 0.471401314253472
450 0.474122858079528
460 0.473046449984723
470 0.457596722454573
480 0.453516661389931
490 0.455017122104679
};
\end{axis}

\end{tikzpicture}
	\end{subfigure}\\
	\vspace{-1em}
	\begin{subfigure}[t]{.49\textwidth}
		\raggedright
		% This file was created with tikzplotlib v0.9.14.
\begin{tikzpicture}

\definecolor{color0}{rgb}{0.12156862745098,0.466666666666667,0.705882352941177}
\definecolor{color1}{rgb}{1,0.498039215686275,0.0549019607843137}
\definecolor{color2}{rgb}{0.172549019607843,0.627450980392157,0.172549019607843}
\definecolor{color3}{rgb}{0.83921568627451,0.152941176470588,0.156862745098039}

\begin{axis}[
height=\figureheight,
legend cell align={left},
legend style={fill opacity=0.8, draw opacity=1, text opacity=1, draw=white!80!black},
tick pos=left,
title={Training Loss (PEP)},
unbounded coords=jump,
width=\figurewidth,
x grid style={white!69.0196078431373!black},
xlabel={iteration number},
xmin=-24.5, xmax=514.5,
xtick style={color=black},
y grid style={white!69.0196078431373!black},
ymin=62, ymax=120,
ytick style={color=black}
]
\addplot [very thick, color0, dashed]
table {%
0 138.18930109732
10 77.9657333775572
20 74.7461156866694
30 74.3900470696912
40 74.3336270547311
50 74.3214025036516
60 74.3182050215383
70 74.317268285115
80 74.3169714569193
90 74.3168715749755
100 74.3168363329248
110 74.3168234281428
120 74.3168185664596
130 74.3168166954617
140 74.3168159640518
150 74.3168156748666
160 74.316815559594
170 74.3168155133778
180 74.3168154947719
190 74.3168154872599
200 74.3168154842204
210 74.3168154829888
220 74.3168154824893
230 74.3168154822867
240 74.3168154822043
250 74.3168154821708
260 74.3168154821572
270 74.3168154821517
280 74.3168154821495
290 74.3168154821487
300 74.3168154821483
310 74.3168154821481
320 74.316815482148
330 74.316815482148
340 74.316815482148
350 74.316815482148
360 74.316815482148
370 74.3168154821481
380 74.316815482148
390 74.3168154821479
400 74.316815482148
410 74.316815482148
420 74.316815482148
430 74.3168154821479
440 74.316815482148
450 74.316815482148
460 74.316815482148
470 74.316815482148
480 74.316815482148
490 74.316815482148
};
\addlegendentry{heuristic PEP}
\addplot [very thick, color1, dashed]
table {%
0 184.594644260846
10 111.654311726311
20 80.8260280528049
30 78.3773063175711
40 79.9570568311326
50 82.116059158423
60 84.9690034765099
70 86.3667155697458
80 87.8538685908043
90 87.8538685908043
100 87.8538685908043
110 87.8538685908043
120 87.8538685908043
130 87.8538685908043
140 87.8538685908043
150 87.8538685908043
160 87.8538685908043
170 87.8538685908043
180 87.8538685908043
190 87.8538685908043
200 87.8538685908043
210 87.8538685908043
220 87.8538685908043
230 87.8538685908043
240 87.8538685908043
250 87.8538685908043
260 87.8538685908043
270 87.8538685908043
280 87.8538685908043
290 87.8538685908043
300 87.8538685908043
310 87.8538685908043
320 87.8538685908043
330 87.8538685908043
340 87.8538685908043
350 87.8538685908043
360 87.8538685908043
370 87.8538685908043
380 87.8538685908043
390 87.8538685908043
400 87.8538685908043
410 87.8538685908043
420 87.8538685908043
430 87.8538685908043
440 87.8538685908043
450 87.8538685908043
460 87.8538685908043
470 87.8538685908043
480 87.8538685908043
490 87.8538685908043
};
\addlegendentry{Riemannian grads PEP}
\addplot [very thick, color2, forget plot]
table {%
0 nan
10 nan
20 nan
30 nan
40 nan
50 nan
60 nan
70 nan
80 nan
90 nan
100 nan
110 nan
120 nan
130 nan
140 nan
150 nan
160 nan
170 nan
180 nan
190 nan
200 nan
210 nan
220 nan
230 nan
240 nan
250 nan
260 nan
270 nan
280 nan
290 nan
300 nan
310 nan
320 nan
330 nan
340 nan
350 nan
360 nan
370 nan
380 nan
390 nan
400 nan
410 nan
420 nan
430 nan
440 nan
450 nan
460 nan
470 nan
480 nan
490 nan
};
\addplot [very thick, color3]
table {%
0 167.310517733466
10 186.126137218983
20 156.605831314398
30 135.446133750421
40 126.64785938292
50 116.635669773087
60 103.364616428405
70 88.5244125614508
80 81.6185569507672
90 81.0020750904034
100 80.5768991797783
110 79.7695614434059
120 79.1871986760429
130 78.7961327467048
140 78.3416413763433
150 78.1729732689636
160 78.1280628384773
170 78.0505671869606
180 78.0483859607795
190 78.2222689130971
200 78.3417389349107
210 78.3847124931619
220 78.426149835783
230 78.4552020971082
240 78.5389711876882
250 78.5569280015631
260 78.7360015715078
270 78.723975401333
280 78.8676151330392
290 79.0555879779162
300 79.1118455527945
310 78.9836933947264
320 79.0497233097835
330 79.0424026297043
340 79.0977955215603
350 78.9770969035599
360 79.0877871447082
370 79.0892371241242
380 79.1828327753988
390 79.0371276511458
400 78.9236602415021
410 79.0344764074446
420 78.9224753167518
430 79.0700915405141
440 79.0651087093626
450 78.9438327863675
460 78.9108588223296
470 78.9192726342591
480 78.9493252027355
490 79.1967627090485
};
\addlegendentry{PEP quasi-Newton}
\end{axis}

\end{tikzpicture}
	\end{subfigure}
	\hspace*{\fill}
	\begin{subfigure}[t]{.49\textwidth}
		\raggedleft
		% This file was created with tikzplotlib v0.9.14.
\begin{tikzpicture}

\definecolor{color0}{rgb}{0.12156862745098,0.466666666666667,0.705882352941177}
\definecolor{color1}{rgb}{1,0.498039215686275,0.0549019607843137}
\definecolor{color2}{rgb}{0.172549019607843,0.627450980392157,0.172549019607843}
\definecolor{color3}{rgb}{0.83921568627451,0.152941176470588,0.156862745098039}

\begin{axis}[
height=\figureheight,
tick pos=left,
title={Test NLPD (PEP)},
unbounded coords=jump,
width=\figurewidth,
x grid style={white!69.0196078431373!black},
xlabel={iteration number},
xmin=-24.5, xmax=514.5,
xtick style={color=black},
y grid style={white!69.0196078431373!black},
ymin=0.31, ymax=1,
ytick style={color=black}
]
\addplot [very thick, color0, dashed]
table {%
0 1.27026630111306
10 0.527669944374846
20 0.434698813737232
30 0.41222287947429
40 0.405718191774957
50 0.403608052545637
60 0.402873613111721
70 0.402605804753435
80 0.402504975546812
90 0.402466157920414
100 0.402450977959191
110 0.402444975642752
120 0.402442583517575
130 0.402441624803829
140 0.402441239020836
150 0.402441083332931
160 0.402441020371677
170 0.402440994871278
180 0.40244098453189
190 0.402440980336361
200 0.402440978632916
210 0.402440977941003
220 0.402440977659873
230 0.402440977545622
240 0.402440977499184
250 0.402440977480306
260 0.402440977472631
270 0.40244097746951
280 0.402440977468242
290 0.402440977467726
300 0.402440977467516
310 0.402440977467431
320 0.402440977467396
330 0.402440977467382
340 0.402440977467377
350 0.402440977467374
360 0.402440977467373
370 0.402440977467373
380 0.402440977467373
390 0.402440977467373
400 0.402440977467373
410 0.402440977467373
420 0.402440977467373
430 0.402440977467373
440 0.402440977467373
450 0.402440977467373
460 0.402440977467373
470 0.402440977467373
480 0.402440977467373
490 0.402440977467373
};
\addplot [very thick, color1, dashed]
table {%
0 1.53222702805757
10 0.885027829984683
20 0.475789788431622
30 0.414574768563283
40 0.404878036646235
50 0.401696405260786
60 0.399818823731182
70 0.399327656560185
80 0.399128730180288
90 0.399128730180288
100 0.399128730180288
110 0.399128730180288
120 0.399128730180288
130 0.399128730180288
140 0.399128730180288
150 0.399128730180288
160 0.399128730180288
170 0.399128730180288
180 0.399128730180288
190 0.399128730180288
200 0.399128730180288
210 0.399128730180288
220 0.399128730180288
230 0.399128730180288
240 0.399128730180288
250 0.399128730180288
260 0.399128730180288
270 0.399128730180288
280 0.399128730180288
290 0.399128730180288
300 0.399128730180288
310 0.399128730180288
320 0.399128730180288
330 0.399128730180288
340 0.399128730180288
350 0.399128730180288
360 0.399128730180288
370 0.399128730180288
380 0.399128730180288
390 0.399128730180288
400 0.399128730180288
410 0.399128730180288
420 0.399128730180288
430 0.399128730180288
440 0.399128730180288
450 0.399128730180288
460 0.399128730180288
470 0.399128730180288
480 0.399128730180288
490 0.399128730180288
};
\addplot [very thick, color2]
table {%
0 nan
10 nan
20 nan
30 nan
40 nan
50 nan
60 nan
70 nan
80 nan
90 nan
100 nan
110 nan
120 nan
130 nan
140 nan
150 nan
160 nan
170 nan
180 nan
190 nan
200 nan
210 nan
220 nan
230 nan
240 nan
250 nan
260 nan
270 nan
280 nan
290 nan
300 nan
310 nan
320 nan
330 nan
340 nan
350 nan
360 nan
370 nan
380 nan
390 nan
400 nan
410 nan
420 nan
430 nan
440 nan
450 nan
460 nan
470 nan
480 nan
490 nan
};
\addplot [very thick, color3]
table {%
0 1.51573621526773
10 1.38523919948486
20 1.13272593154447
30 0.931170173018682
40 0.880279479980962
50 0.848575779555826
60 0.776933196813402
70 0.64277158210454
80 0.550872616210666
90 0.52618213340473
100 0.508825751394018
110 0.493674717033811
120 0.479724623541331
130 0.47038186641802
140 0.455678589486433
150 0.446964979438121
160 0.438468333265495
170 0.432985569696065
180 0.423876187037591
190 0.417820172404449
200 0.414578774625144
210 0.411185377349959
220 0.408011726849929
230 0.406483230862136
240 0.406760245830023
250 0.409430750963774
260 0.409523097649959
270 0.405646430574324
280 0.405737771623115
290 0.405533878054703
300 0.407266732370758
310 0.406254413840269
320 0.409098542833174
330 0.40764378282798
340 0.407411570640846
350 0.41100159430206
360 0.410128132563734
370 0.4095814807712
380 0.412609537109765
390 0.410972756812061
400 0.412989841481886
410 0.408982462649801
420 0.409655933245084
430 0.409636807065454
440 0.406804103871411
450 0.406946294994875
460 0.405930251965241
470 0.405470999886905
480 0.404392353967396
490 0.407108833311557
};
\end{axis}

\end{tikzpicture}
	\end{subfigure}\\
	\vspace{-1em}
	\begin{subfigure}[t]{.49\textwidth}
		\raggedright
		% This file was created with tikzplotlib v0.9.14.
\begin{tikzpicture}

\definecolor{color0}{rgb}{0.12156862745098,0.466666666666667,0.705882352941177}
\definecolor{color1}{rgb}{1,0.498039215686275,0.0549019607843137}
\definecolor{color2}{rgb}{0.172549019607843,0.627450980392157,0.172549019607843}
\definecolor{color3}{rgb}{0.83921568627451,0.152941176470588,0.156862745098039}
\definecolor{color4}{rgb}{0.580392156862745,0.403921568627451,0.741176470588235}

\begin{axis}[
height=\figureheight,
legend cell align={left},
legend style={fill opacity=0.8, draw opacity=1, text opacity=1, draw=white!80!black},
tick pos=left,
title={Training Loss (PL2)},
unbounded coords=jump,
width=\figurewidth,
x grid style={white!69.0196078431373!black},
xlabel={iteration number},
xmin=-24.5, xmax=514.5,
xtick style={color=black},
y grid style={white!69.0196078431373!black},
ymin=62, ymax=120,
ytick style={color=black}
]
\addplot [very thick, color0, dashed]
table {%
0 111.524336689331
10 104.929661932953
20 104.954106475365
30 104.957055432058
40 104.95720228855
50 104.957208231526
60 104.957208450092
70 104.957208457698
80 104.957208457953
90 104.957208457961
100 104.957208457962
110 104.957208457963
120 104.957208457962
130 104.957208457962
140 104.957208457963
150 104.957208457962
160 104.957208457962
170 104.957208457961
180 104.957208457961
190 104.957208457963
200 104.957208457963
210 104.957208457961
220 104.957208457962
230 104.957208457963
240 104.957208457962
250 104.957208457962
260 104.957208457962
270 104.957208457963
280 104.957208457964
290 104.957208457963
300 104.957208457961
310 104.957208457962
320 104.957208457963
330 104.957208457962
340 104.957208457962
350 104.95720845796
360 104.957208457961
370 104.957208457961
380 104.957208457961
390 104.957208457961
400 104.957208457961
410 104.957208457962
420 104.957208457963
430 104.957208457963
440 104.957208457964
450 104.957208457964
460 104.957208457962
470 104.957208457961
480 104.957208457961
490 104.957208457959
};
\addlegendentry{PL}
\addplot [very thick, color1, dashed, forget plot]
table {%
0 nan
10 nan
20 nan
30 nan
40 nan
50 nan
60 nan
70 nan
80 nan
90 nan
100 nan
110 nan
120 nan
130 nan
140 nan
150 nan
160 nan
170 nan
180 nan
190 nan
200 nan
210 nan
220 nan
230 nan
240 nan
250 nan
260 nan
270 nan
280 nan
290 nan
300 nan
310 nan
320 nan
330 nan
340 nan
350 nan
360 nan
370 nan
380 nan
390 nan
400 nan
410 nan
420 nan
430 nan
440 nan
450 nan
460 nan
470 nan
480 nan
490 nan
};
\addplot [very thick, color2]
table {%
0 105.876377948056
10 75.103676661711
20 76.9693275904691
30 78.016879824497
40 78.5397021949731
50 78.8099781275398
60 78.948754062891
70 79.0184627068594
80 79.0528771240095
90 79.0696910744676
100 79.0778597430445
110 79.0818168013361
120 79.083730886361
130 79.0846560882877
140 79.0851031396212
150 79.0853191145154
160 79.085423445494
170 79.0854738426771
180 79.085498186306
190 79.0855099454932
200 79.085515625492
210 79.0855183686405
220 79.085519694059
230 79.0855203338747
240 79.0855206434725
250 79.0855207926939
260 79.0855208651633
270 79.0855208998806
280 79.0855209166519
290 79.0855209246309
300 79.0855209286598
310 79.0855209305561
320 79.0855209311441
330 79.0855209320857
340 79.0855209319645
350 79.085520932159
360 79.0855209318877
370 79.0855209319129
380 79.0855209322118
390 79.085520932034
400 79.0855209322426
410 79.0855209323463
420 79.0855209323945
430 79.0855209324194
440 79.0855209324279
450 79.0855209324296
460 79.0855209324351
470 79.0855209324343
480 79.0855209324402
490 79.0855209324336
};
\addlegendentry{PL2 Gauss-Newton}
\addplot [very thick, color3]
table {%
0 187.7865913842
10 165.58872795763
20 134.323645953774
30 122.751574126902
40 111.62944447628
50 94.6517592185864
60 77.4984676247914
70 78.2431074083785
80 79.934943164749
90 79.9131147682143
100 80.0959934451164
110 80.4978235005133
120 80.9967733470718
130 81.3150976882372
140 81.7136156337807
150 81.5482860663743
160 81.4068021353778
170 80.9907361885618
180 81.5902407500795
190 82.2053945611245
200 81.9926247482164
210 81.0910128211314
220 81.6268108313126
230 81.7570095646815
240 82.856456561095
250 81.8596719157897
260 81.5290996982991
270 81.2805428237989
280 81.5966258380257
290 82.6278846308461
300 82.2534243885854
310 82.1244276669032
320 81.9722704236419
330 81.5318130126743
340 81.5299361700707
350 82.033657225186
360 82.5883059759486
370 83.1653252284372
380 83.0202558828895
390 82.3947185431312
400 81.6440284278898
410 82.2709058029834
420 83.2775422342395
430 83.3328060196771
440 83.3413845555772
450 82.6291425361459
460 82.2314874969878
470 82.2908787529369
480 82.8310312376302
490 83.4300847581794
};
\addlegendentry{PL2 quasi-Newton}
\addplot [very thick, color4, dashed]
table {%
0 102.985961115877
10 76.3769702399156
20 77.5498732180926
30 77.9219554388594
40 77.9982439867494
50 78.013344667675
60 78.0163975153279
70 78.0170289359208
80 78.0171617861659
90 78.0171900748624
100 78.0171961490967
110 78.0171974612103
120 78.0171977459127
130 78.017197807822
140 78.0171978213682
150 78.0171978243795
160 78.0171978249642
170 78.0171978251628
180 78.0171978251435
190 78.0171978250784
200 78.017197825122
210 78.0171978251314
220 78.0171978251327
230 78.0171978251338
240 78.0171978251337
250 78.0171978251339
260 78.0171978251341
270 78.017197825134
280 78.017197825134
290 78.0171978251342
300 78.017197825134
310 78.017197825134
320 78.017197825134
330 78.017197825134
340 78.0171978251336
350 78.0171978251341
360 78.0171978251339
370 78.0171978251341
380 78.0171978251338
390 78.0171978251342
400 78.0171978251339
410 78.0171978251339
420 78.0171978251339
430 78.017197825134
440 78.0171978251342
450 78.0171978251341
460 78.0171978251342
470 78.0171978251338
480 78.017197825134
490 78.017197825134
};
\addlegendentry{heuristic PL2}
\end{axis}

\end{tikzpicture}
	\end{subfigure}
	\hspace*{\fill}
	\begin{subfigure}[t]{.49\textwidth}
		\raggedleft
		% This file was created with tikzplotlib v0.9.14.
\begin{tikzpicture}

\definecolor{color0}{rgb}{0.12156862745098,0.466666666666667,0.705882352941177}
\definecolor{color1}{rgb}{1,0.498039215686275,0.0549019607843137}
\definecolor{color2}{rgb}{0.172549019607843,0.627450980392157,0.172549019607843}
\definecolor{color3}{rgb}{0.83921568627451,0.152941176470588,0.156862745098039}
\definecolor{color4}{rgb}{0.580392156862745,0.403921568627451,0.741176470588235}

\begin{axis}[
height=\figureheight,
tick pos=left,
title={Test NLPD (PL2)},
unbounded coords=jump,
width=\figurewidth,
x grid style={white!69.0196078431373!black},
xlabel={iteration number},
xmin=-24.5, xmax=514.5,
xtick style={color=black},
y grid style={white!69.0196078431373!black},
ymin=0.31, ymax=1,
ytick style={color=black}
]
\addplot [very thick, color0, dashed]
table {%
0 1.07191624932715
10 0.954872204782482
20 0.954958386385451
30 0.955008527535692
40 0.955011290406905
50 0.955011406478405
60 0.95501141083128
70 0.95501141098458
80 0.955011410989767
90 0.955011410989938
100 0.955011410989944
110 0.955011410989944
120 0.955011410989944
130 0.955011410989944
140 0.955011410989944
150 0.955011410989944
160 0.955011410989944
170 0.955011410989944
180 0.955011410989944
190 0.955011410989944
200 0.955011410989944
210 0.955011410989944
220 0.955011410989944
230 0.955011410989944
240 0.955011410989944
250 0.955011410989944
260 0.955011410989944
270 0.955011410989944
280 0.955011410989944
290 0.955011410989944
300 0.955011410989944
310 0.955011410989944
320 0.955011410989944
330 0.955011410989944
340 0.955011410989944
350 0.955011410989944
360 0.955011410989944
370 0.955011410989944
380 0.955011410989944
390 0.955011410989944
400 0.955011410989944
410 0.955011410989944
420 0.955011410989944
430 0.955011410989944
440 0.955011410989944
450 0.955011410989944
460 0.955011410989944
470 0.955011410989944
480 0.955011410989944
490 0.955011410989944
};
\addplot [very thick, color1, dashed]
table {%
0 nan
10 nan
20 nan
30 nan
40 nan
50 nan
60 nan
70 nan
80 nan
90 nan
100 nan
110 nan
120 nan
130 nan
140 nan
150 nan
160 nan
170 nan
180 nan
190 nan
200 nan
210 nan
220 nan
230 nan
240 nan
250 nan
260 nan
270 nan
280 nan
290 nan
300 nan
310 nan
320 nan
330 nan
340 nan
350 nan
360 nan
370 nan
380 nan
390 nan
400 nan
410 nan
420 nan
430 nan
440 nan
450 nan
460 nan
470 nan
480 nan
490 nan
};
\addplot [very thick, color2]
table {%
0 0.975266809706321
10 0.399324646763277
20 0.363305970512693
30 0.377587105215023
40 0.391862769115198
50 0.400729225064725
60 0.405512418804
70 0.407943559199523
80 0.40914645589245
90 0.409734216269072
100 0.410019700305343
110 0.41015796750701
120 0.410224841337091
130 0.410257163690659
140 0.410272781111597
150 0.410280325910807
160 0.410283970538427
170 0.410285731065552
180 0.410286581468159
190 0.410286992241936
200 0.410287190659013
210 0.41028728650072
220 0.410287332795236
230 0.410287355156919
240 0.410287365958289
250 0.410287371175687
260 0.410287373695847
270 0.410287374913162
280 0.410287375501164
290 0.410287375785188
300 0.41028737592238
310 0.410287375988648
320 0.410287376020662
330 0.410287376036117
340 0.41028737604359
350 0.410287376047196
360 0.410287376048944
370 0.410287376049785
380 0.410287376050187
390 0.410287376050387
400 0.410287376050478
410 0.410287376050522
420 0.410287376050542
430 0.410287376050552
440 0.410287376050556
450 0.410287376050559
460 0.41028737605056
470 0.41028737605056
480 0.410287376050561
490 0.410287376050561
};
\addplot [very thick, color3]
table {%
0 1.55613583936095
10 1.27043123908483
20 0.977295633670686
30 0.866303640816546
40 0.817535649671965
50 0.700877431430585
60 0.482762054466504
70 0.373453111884532
80 0.382715146836315
90 0.387493756432469
100 0.384904244636106
110 0.383572235532929
120 0.381215037239365
130 0.387205584991715
140 0.38782619936113
150 0.385174818476938
160 0.378621290750297
170 0.389726910609271
180 0.38680050511166
190 0.389022450272736
200 0.38298195101556
210 0.391363311945353
220 0.396159005736161
230 0.390667114487307
240 0.387169162268674
250 0.397840089259621
260 0.392847791537297
270 0.38440379617735
280 0.380760311815326
290 0.377182177857307
300 0.404359536863514
310 0.391954568022224
320 0.382457727945509
330 0.391594523042808
340 0.393146218716523
350 0.386076810390012
360 0.389448208854602
370 0.393729499022342
380 0.393210010170352
390 0.387623058853231
400 0.389054082599509
410 0.384656346146665
420 0.392977708184947
430 0.385461227005169
440 0.379566016607007
450 0.388662471053338
460 0.394507973404442
470 0.397349992845599
480 0.400490652366021
490 0.392826962093358
};
\addplot [very thick, color4, dashed]
table {%
0 0.955830821073266
10 0.376989935878443
20 0.381798537194223
30 0.385150389532196
40 0.385981456200981
50 0.386167758363936
60 0.386208706556739
70 0.386217675099714
80 0.386219636729273
90 0.386220065648605
100 0.386220159461561
110 0.386220179994152
120 0.38622018449219
130 0.386220185478605
140 0.386220185695166
150 0.38622018574276
160 0.386220185753234
170 0.386220185755539
180 0.386220185756049
190 0.386220185756163
200 0.386220185756187
210 0.386220185756192
220 0.386220185756193
230 0.386220185756193
240 0.386220185756193
250 0.386220185756193
260 0.386220185756193
270 0.386220185756193
280 0.386220185756193
290 0.386220185756193
300 0.386220185756193
310 0.386220185756193
320 0.386220185756193
330 0.386220185756193
340 0.386220185756193
350 0.386220185756193
360 0.386220185756193
370 0.386220185756193
380 0.386220185756193
390 0.386220185756193
400 0.386220185756193
410 0.386220185756193
420 0.386220185756193
430 0.386220185756193
440 0.386220185756193
450 0.386220185756193
460 0.386220185756193
470 0.386220185756193
480 0.386220185756193
490 0.386220185756193
};
\end{axis}

\end{tikzpicture}
	\end{subfigure}\\
	\vspace{-0.35cm}
	\caption{Heteroscedastic noise results. Mean of 4-fold cross validation shown. Variational quasi-Newton is capable of obtaining the best predictive performance, however all quasi-Newton methods converge slowly due to the need for damping. Gauss--Newton variants obtain similar performance to the heuristic method, but do not require computation of second-order derivatives. PL2 is the second-order PL method, and it significantly outperforms PL because it takes into account the gradients of the likelihood covariance.} \label{fig:hsced-results}
	\vspace{-0.2cm}
\end{figure*}
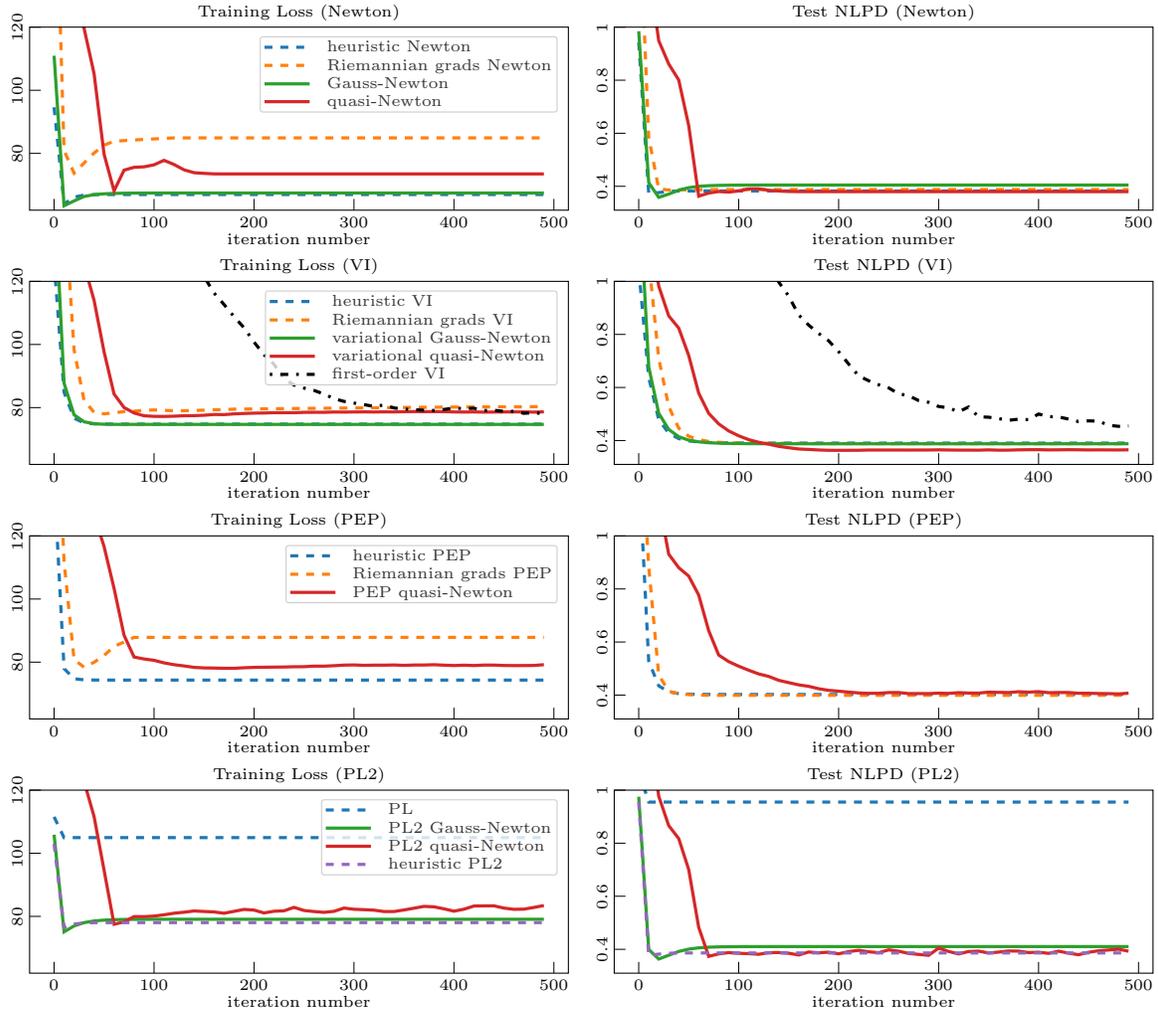

\subsubsection{Gaussian Process Regression with Heteroscedastic Noise}\label{sec:hsced}

First we consider the model of \citet{goldberg1997regression}, which augments a standard regression model with an additional GP prior on the observation noise standard deviation,
\begin{gather}
	\begin{aligned}
		&f_1(\cdot) \sim \GP(0, \kappa_1(\cdot, \cdot)) \, , \quad f_2(\cdot) \sim \GP(0, \kappa_2(\cdot, \cdot)) \, ,\\
		&\vy_n \mid f_1(\MX_n), f_2(\MX_n) \sim \N(\vy_n \mid f_1(\MX_n),\, \phi(f_2(\MX_n))^2) \, ,
	\end{aligned}
\end{gather}
where $\phi(\cdot)=\log(1+\exp(\cdot))$ is the softplus function which ensures the standard deviation is positive. There are two latent processes, $\fdim=2$, and the observations are scalars, $\ydim=1$.
This model has been a focus of much research due to its relevance to real-world applications \citep{tolvanen2014expectation, lazaro2011variational}. The model is applied to data simulating $N=133$ accelerometer readings from a motorcycle crash \citep{silverman1985some}. $\kappa_1$, $\kappa_2$ are Mat\'ern-$\nicefrac{3}{2}$ kernels, and we use a learning rate of $\rho=0.3$ and a quasi-Newton damping rate of $\xi=0.5$. The data inputs and outputs are scaled to have zero mean and unit variance, and the kernel hyperparameters (lengthscales and variances) are all fixed at the value 1. We use Gauss--Hermite integration with 20$^2 = $\,400 points to solve the intractable integrals required for the VI-, EP- and PL-based methods. An example inference result is shown in \cref{fig:hsced-plot}, and the test performance using 4-fold cross validation is shown in \cref{fig:hsced-results}.

Standard PL fails badly on this task since it does not take into account the gradients of the likelihood covariance (see \cref{sec:improved-pl} for discussion). From \cref{fig:hsced-plot} we can see that variational Gauss--Newton does not capture a significant amount of cross-covariance between the latent components and provides very similar performance to the heuristic method, but has the benefit of not requiring computation of the full Hessian. The variational quasi-Newton method provides the best test performance, but all quasi-Newton methods converge slowly due to the use of damped BFGS updates (see \cref{sec:damped-quasi-newton}).

\begin{figure*}[t!]
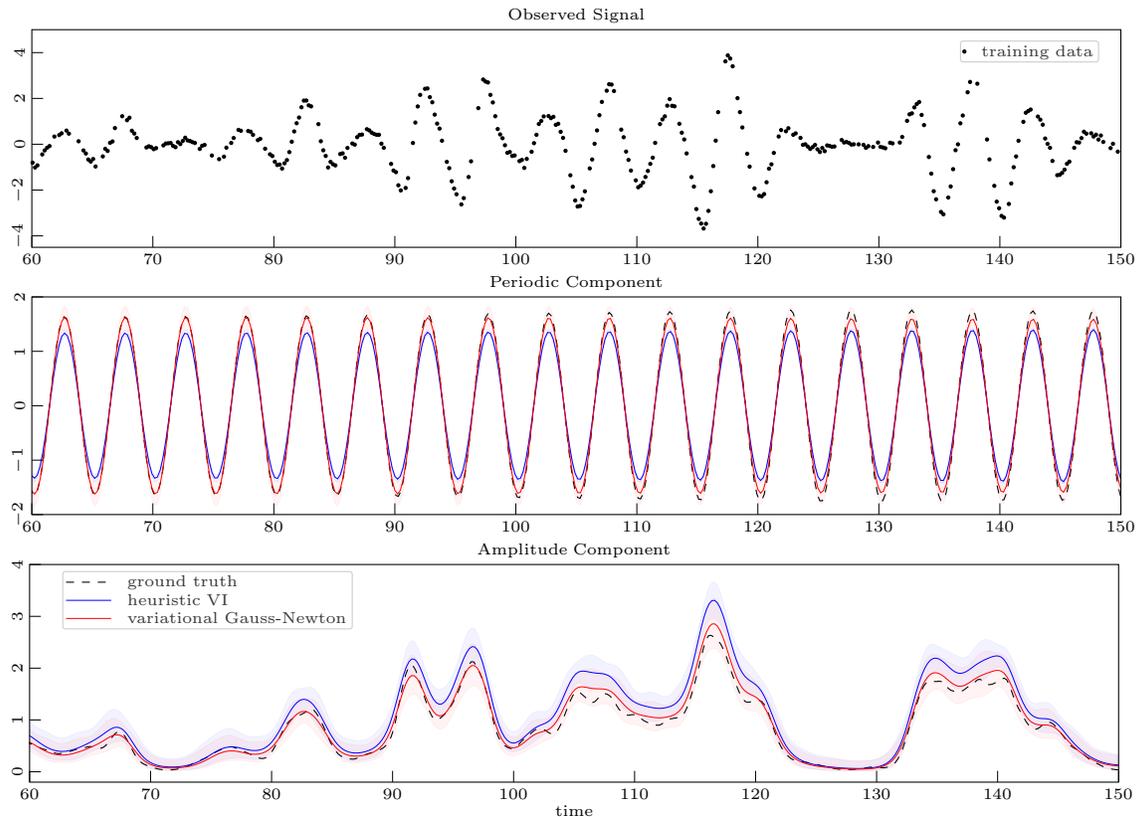

	\centering\tiny
	\pgfplotsset{yticklabel style={rotate=90}, ylabel style={yshift=0pt},scale only axis,axis on top,title style={yshift=-6pt}, xlabel style={yshift=3pt}}
	\pgfplotsset{legend style={inner xsep=1pt, inner ysep=1pt, row sep=0pt},legend style={at={(0.98,0.95)},anchor=north east},legend style={rounded corners=1pt}}
	\setlength{\figurewidth}{0.95\textwidth}
	\setlength{\figureheight}{.2\figurewidth}
	\begin{subfigure}[t]{\textwidth}
		\input{./fig/product-sig.tex}
	\end{subfigure}\\
	\vspace{-0.2em}
	\begin{subfigure}[t]{\textwidth}
		\input{./fig/product-subband.tex}
	\end{subfigure}\\
	\vspace{-0.2em}
	\begin{subfigure}[t]{\textwidth}
		\input{./fig/product-modulator.tex}
	\end{subfigure}\\
	\vspace{-0.5em}
	\caption{A short segment of the signal used for the amplitude demodulation experiment. The observed signal is produced via the product of a periodic component and a positive amplitude envelope. By taking into account cross-covariance between components, variational Gauss--Newton is better able to recover the ground truth than the heuristic approach.} \label{fig:product-plot}
\end{figure*}

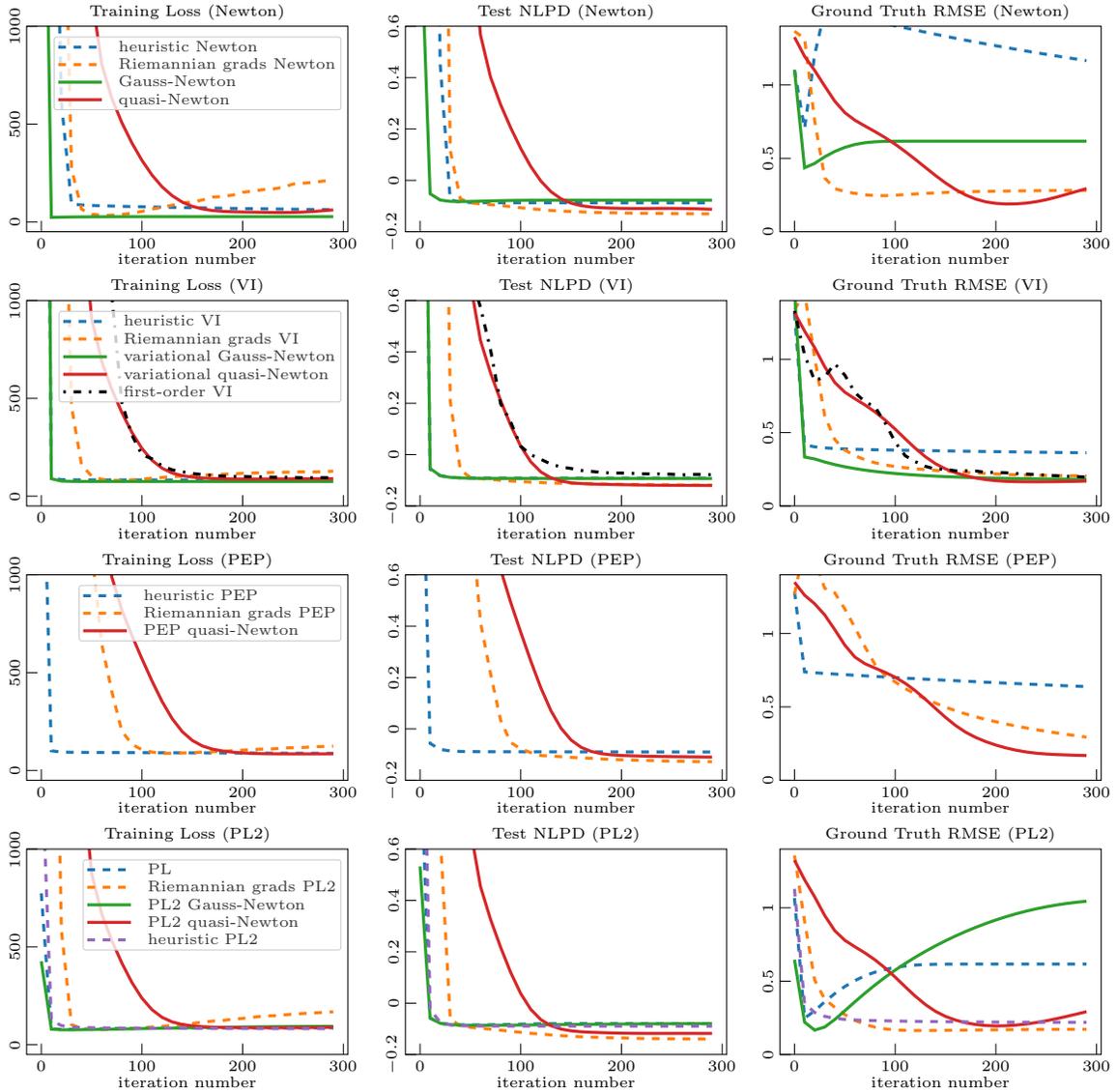
\begin{figure*}[t!]
	\centering\tiny
	\pgfplotsset{yticklabel style={rotate=90}, ylabel style={yshift=0pt},scale only axis,axis on top,title style={yshift=-6pt}, xlabel style={yshift=3pt}}
	\pgfplotsset{legend style={inner xsep=1pt, inner ysep=1pt, row sep=0pt},legend style={at={(0.98,0.95)},anchor=north east},legend style={rounded corners=1pt}}
	\setlength{\figurewidth}{.29\textwidth}
	\setlength{\figureheight}{.64\figurewidth}
	\begin{subfigure}[t]{.32\textwidth}
		\raggedright
		% This file was created with tikzplotlib v0.9.14.
\begin{tikzpicture}

\definecolor{color0}{rgb}{0.12156862745098,0.466666666666667,0.705882352941177}
\definecolor{color1}{rgb}{1,0.498039215686275,0.0549019607843137}
\definecolor{color2}{rgb}{0.172549019607843,0.627450980392157,0.172549019607843}
\definecolor{color3}{rgb}{0.83921568627451,0.152941176470588,0.156862745098039}

\begin{axis}[
height=\figureheight,
legend cell align={left},
legend style={fill opacity=0.8, draw opacity=1, text opacity=1, draw=white!80!black},
tick pos=left,
title={Training Loss (Newton)},
width=\figurewidth,
x grid style={white!69.0196078431373!black},
xlabel={iteration number},
xmin=-14.5, xmax=304.5,
xtick style={color=black},
y grid style={white!69.0196078431373!black},
ymin=-50, ymax=1000,
ytick style={color=black}
]
\addplot [very thick, color0, dashed]
table {%
0 3347.35365145776
10 2798.65883459206
20 610.735752696563
30 92.1808708901732
40 86.6063006048996
50 84.300244626142
60 82.4947198682178
70 80.9860773933353
80 79.6681228357773
90 78.4738942878518
100 77.3634077729868
110 76.3136558148209
120 75.3112427876399
130 74.3484091644259
140 73.4203854540481
150 72.5239841778842
160 71.6568917581192
170 70.817294745704
180 70.003681744996
190 69.2147367835227
200 68.4492803285668
210 67.7062353505994
220 66.9846067623534
230 66.2834681676194
240 65.6019527324669
250 64.939246477778
260 64.2945830659256
270 63.6673421847614
280 63.0567324828017
290 62.4620509148492
};
\addlegendentry{heuristic Newton}
\addplot [very thick, color1, dashed]
table {%
0 8376.54616445138
10 6035.09042314098
20 2463.52582170802
30 273.509657426135
40 65.0723815169741
50 39.5701748919991
60 33.5329041877366
70 34.2605398663856
80 38.4236003729792
90 44.2877944819768
100 53.6993628259381
110 65.6610572450425
120 74.0916314747896
130 85.4875198055786
140 97.5605125640707
150 101.680689884124
160 112.244673852015
170 126.976125307703
180 130.105214708776
190 137.156239481999
200 151.412931307462
210 156.992977201204
220 164.728447094178
230 168.059713756759
240 176.800358214276
250 193.374597546725
260 197.889953251305
270 201.899997190053
280 208.923388041971
290 211.197063732956
};
\addlegendentry{Riemannian grads Newton}
\addplot [very thick, color2]
table {%
0 3347.35365145776
10 23.0357886863341
20 24.3877727621601
30 25.1990188343486
40 25.6011598246523
50 25.8923205993399
60 26.1562349899005
70 26.3772724007446
80 26.5358073191575
90 26.6328387392546
100 26.6839759287522
110 26.7076826277323
120 26.7176900387915
130 26.721674936021
140 26.7232105759414
150 26.7237921423692
160 26.7240103285492
170 26.7240918581389
180 26.7241221534347
190 26.7241331531079
200 26.7241381796709
210 26.7241403553487
220 26.7241407883956
230 26.7241394271325
240 26.7240482620714
250 26.7244247288547
260 26.7237294784958
270 26.7240993899778
280 26.72498551858
290 26.7223080327273
};
\addlegendentry{Gauss-Newton}
\addplot [very thick, color3]
table {%
0 7995.36004076806
10 6388.46786888743
20 5085.46285826456
30 3444.96967500014
40 1980.32151697899
50 1163.26537333666
60 807.324570610245
70 632.562688159498
80 507.619041336775
90 403.787479319319
100 314.567090742888
110 239.429389825929
120 179.375505888856
130 133.970805683187
140 101.284422973406
150 79.6102462241399
160 66.1701599579634
170 58.2154960163123
180 53.9752750387889
190 51.7804690351571
200 50.6438192602508
210 49.8996393646473
220 49.1103366026069
230 48.1807692467803
240 47.8286279804196
250 48.5243126958222
260 50.3546986699567
270 53.2685628736944
280 57.0700555825774
290 61.5035738196098
};
\addlegendentry{quasi-Newton}
\end{axis}

\end{tikzpicture}
	\end{subfigure}
	\hspace*{\fill}
	\begin{subfigure}[t]{.32\textwidth}
		% This file was created with tikzplotlib v0.9.14.
\begin{tikzpicture}

\definecolor{color0}{rgb}{0.12156862745098,0.466666666666667,0.705882352941177}
\definecolor{color1}{rgb}{1,0.498039215686275,0.0549019607843137}
\definecolor{color2}{rgb}{0.172549019607843,0.627450980392157,0.172549019607843}
\definecolor{color3}{rgb}{0.83921568627451,0.152941176470588,0.156862745098039}

\begin{axis}[
height=\figureheight,
tick pos=left,
title={Test NLPD (Newton)},
width=\figurewidth,
x grid style={white!69.0196078431373!black},
xlabel={iteration number},
xmin=-14.5, xmax=304.5,
xtick style={color=black},
y grid style={white!69.0196078431373!black},
ymin=-0.2, ymax=0.6,
ytick style={color=black}
]
\addplot [very thick, color0, dashed]
table {%
0 1.0597247072457
10 3.4093924455801
20 0.451146173654095
30 -0.0762405046320048
40 -0.0844359428263825
50 -0.0860992823436642
60 -0.0866425189204535
70 -0.0868466380365491
80 -0.086942836164412
90 -0.0870052841241719
100 -0.087057218834288
110 -0.0871056885110839
120 -0.0871524679424279
130 -0.0871980177701532
140 -0.0872424065664935
150 -0.0872855844059317
160 -0.0873274880625226
170 -0.0873680702547987
180 -0.0874073053419137
190 -0.0874451876384899
200 -0.0874817276567125
210 -0.0875169481410192
220 -0.0875508805555558
230 -0.0875835622139132
240 -0.0876150340459572
250 -0.0876453389204101
260 -0.0876745204206474
270 -0.0877026833259939
280 -0.0877298044475734
290 -0.0877558927479997
};
\addplot [very thick, color1, dashed]
table {%
0 10.9086705529464
10 7.74429838803463
20 2.92497461387961
30 0.117399609498175
40 -0.0722435895617121
50 -0.0879371135096002
60 -0.0913182239309942
70 -0.0944653689112923
80 -0.0983491771759699
90 -0.102233559080618
100 -0.106869464585183
110 -0.110830392580347
120 -0.113795749153155
130 -0.117257952642229
140 -0.120019973832688
150 -0.121073159950469
160 -0.122711136260536
170 -0.124789274128186
180 -0.125489574380053
190 -0.126230942150718
200 -0.127138761492498
210 -0.127685706642309
220 -0.128266756638518
230 -0.128562115927671
240 -0.129023484224245
250 -0.129669294078047
260 -0.129825128056881
270 -0.1301327518397
280 -0.130339776167099
290 -0.130383459923332
};
\addplot [very thick, color2]
table {%
0 1.0597247072457
10 -0.0523339792058533
20 -0.0768052113845692
30 -0.0818474419875839
40 -0.0825573969115476
50 -0.0820157357390433
60 -0.0810936417495893
70 -0.080046762581114
80 -0.0790532443668472
90 -0.0782900940909502
100 -0.0778276180815225
110 -0.077602664346874
120 -0.0775094768972321
130 -0.0774743251661098
140 -0.0774616664793539
150 -0.0774572015882057
160 -0.0774556405337534
170 -0.0774550967474252
180 -0.0774549076543293
190 -0.0774548415119412
200 -0.0774548180333517
210 -0.0774548103029634
220 -0.077454810810534
230 -0.0774547969457949
240 -0.077454790130845
250 -0.0774547042877455
260 -0.0774552261516463
270 -0.0774551757858216
280 -0.0774545885045234
290 -0.0774552479081647
};
\addplot [very thick, color3]
table {%
0 10.3591210414083
10 8.06751076840868
20 6.2641660945783
30 4.02654202606209
40 2.05627483954226
50 0.9869807974587
60 0.566621705412985
70 0.401697497140555
80 0.295065650057901
90 0.205009991374835
100 0.124042796928799
110 0.0537252513745621
120 -0.00287802662190423
130 -0.0443023334174886
140 -0.0718704988244291
150 -0.0888498518022312
160 -0.0986597946129148
170 -0.103969960291852
180 -0.106782218265838
190 -0.108259187952227
200 -0.109089019689315
210 -0.109481419089752
220 -0.109648735502229
230 -0.109367800141976
240 -0.109116444935821
250 -0.109189129035798
260 -0.109605147844832
270 -0.110366814211943
280 -0.111415489114235
290 -0.112663292321699
};
\end{axis}

\end{tikzpicture} 
	\end{subfigure}
	\hspace*{\fill}
	\begin{subfigure}[t]{.32\textwidth}
		\raggedleft
		% This file was created with tikzplotlib v0.9.14.
\begin{tikzpicture}

\definecolor{color0}{rgb}{0.12156862745098,0.466666666666667,0.705882352941177}
\definecolor{color1}{rgb}{1,0.498039215686275,0.0549019607843137}
\definecolor{color2}{rgb}{0.172549019607843,0.627450980392157,0.172549019607843}
\definecolor{color3}{rgb}{0.83921568627451,0.152941176470588,0.156862745098039}

\begin{axis}[
height=\figureheight,
tick pos=left,
title={Ground Truth RMSE (Newton)},
width=\figurewidth,
x grid style={white!69.0196078431373!black},
xlabel={iteration number},
xmin=-14.5, xmax=304.5,
xtick style={color=black},
y grid style={white!69.0196078431373!black},
ymin=0, ymax=1.4,
ytick style={color=black}
]
\addplot [very thick, color0, dashed]
table {%
0 1.10267337034807
10 0.702871971999582
20 1.20505612936285
30 1.50716887514466
40 1.50293630620379
50 1.48534658445393
60 1.4679528943787
70 1.45121460961069
80 1.43497895256138
90 1.41915562970586
100 1.40370606787018
110 1.38861224080316
120 1.37386231887176
130 1.35944570113959
140 1.34535172850648
150 1.33156955105163
160 1.31808829445334
170 1.30489724229421
180 1.29198596739811
190 1.27934441067908
200 1.26696292084829
210 1.25483226827467
220 1.24294364294554
230 1.23128864321913
240 1.21985925963309
250 1.20864785639052
260 1.19764715208703
270 1.18685020483018
280 1.17625039079287
290 1.1658413665151
};
\addplot [very thick, color1, dashed]
table {%
0 1.36609137780195
10 1.31670820611578
20 0.756191997668308
30 0.366578836502959
40 0.293574389112715
50 0.273625453755985
60 0.261268134175375
70 0.251635560219126
80 0.245727570087551
90 0.245311327471695
100 0.248919637847953
110 0.253320804736067
120 0.257335622686199
130 0.261181597041642
140 0.264536584842386
150 0.267277847229995
160 0.269658312116445
170 0.271650133725693
180 0.273326451395021
190 0.274771910492631
200 0.275953122424858
210 0.277012851693057
220 0.277961429855666
230 0.278737213205222
240 0.279425007376334
250 0.280006282205376
260 0.280508321040099
270 0.280843351892384
280 0.281028273396062
290 0.28112284485759
};
\addplot [very thick, color2]
table {%
0 1.10267337034807
10 0.435586236753639
20 0.465276629631139
30 0.510339246985833
40 0.546968504825905
50 0.574755748231836
60 0.594284290098372
70 0.60641230778748
80 0.612804657207401
90 0.615574398074257
100 0.616554054831759
110 0.616848449403001
120 0.616929980826877
130 0.616952631501265
140 0.616959301195814
150 0.616961415212818
160 0.6169621277707
170 0.616962377938522
180 0.616962468303936
190 0.61696250023833
200 0.616962513057017
210 0.616962515894877
220 0.616962513176717
230 0.616962524786948
240 0.616962525470534
250 0.616962186966602
260 0.616963022702121
270 0.616964134616914
280 0.616962655254906
290 0.616961957092087
};
\addplot [very thick, color3]
table {%
0 1.32585120429672
10 1.19525283405214
20 1.09933409968265
30 0.992555144845264
40 0.889359365237588
50 0.811443365427487
60 0.759770984320442
70 0.720145946265006
80 0.682053640557765
90 0.641244829727018
100 0.596059025156954
110 0.546292819789116
120 0.492907368846061
130 0.437754402548594
140 0.383280868309904
150 0.332209733002802
160 0.28708953485841
170 0.249987114626616
180 0.222140305717634
190 0.203604189097174
200 0.193154833609025
210 0.188989871490201
220 0.189766367616515
230 0.194943854461219
240 0.204453671747862
250 0.21809134225559
260 0.235117101970608
270 0.254282549776961
280 0.274202716338178
290 0.293791846649663
};
\end{axis}

\end{tikzpicture} 
	\end{subfigure}\\
	\vspace{-1em}
	\begin{subfigure}[t]{.32\textwidth}
		\raggedright
		% This file was created with tikzplotlib v0.9.14.
\begin{tikzpicture}

\definecolor{color0}{rgb}{0.12156862745098,0.466666666666667,0.705882352941177}
\definecolor{color1}{rgb}{1,0.498039215686275,0.0549019607843137}
\definecolor{color2}{rgb}{0.172549019607843,0.627450980392157,0.172549019607843}
\definecolor{color3}{rgb}{0.83921568627451,0.152941176470588,0.156862745098039}

\begin{axis}[
height=\figureheight,
legend cell align={left},
legend style={fill opacity=0.8, draw opacity=1, text opacity=1, draw=white!80!black},
tick pos=left,
title={Training Loss (VI)},
width=\figurewidth,
x grid style={white!69.0196078431373!black},
xlabel={iteration number},
xmin=-14.5, xmax=304.5,
xtick style={color=black},
y grid style={white!69.0196078431373!black},
ymin=-50, ymax=1000,
ytick style={color=black}
]
\addplot [very thick, color0, dashed]
table {%
0 6165.55983695117
10 93.3928803546534
20 84.6887146152302
30 83.5610815555014
40 83.2807295212848
50 83.1658644590468
60 83.0956428806849
70 83.0407744822852
80 82.992639401751
90 82.9483545840323
100 82.906823086947
110 82.8675613213549
120 82.8303106882189
130 82.7948990330393
140 82.7611925269702
150 82.7290771438669
160 82.6984510253701
170 82.6692210418962
180 82.6413010583085
190 82.6146109370231
200 82.5890758845055
210 82.5646259716923
220 82.541195750902
230 82.5187239315245
240 82.4971530949302
250 82.4764294378139
260 82.4565025376255
270 82.4373251361556
280 82.4188529386794
290 82.4010444268252
};
\addlegendentry{heuristic VI}
\addplot [very thick, color1, dashed]
table {%
0 8416.22770352752
10 5432.65007892717
20 2435.00400105911
30 460.010609741802
40 160.149335554815
50 102.615002996291
60 86.5722406285392
70 81.3837071009891
80 81.1484572807902
90 83.153776722301
100 87.326622160713
110 91.8035643245198
120 95.8247284074017
130 100.122980400298
140 102.534840638558
150 104.788253031041
160 107.736667151601
170 109.363402817499
180 112.531939749837
190 114.04765061134
200 117.14047196259
210 118.43129249595
220 119.976697076903
230 121.574881986946
240 122.211661391617
250 122.88677820977
260 124.61674317956
270 123.841087988149
280 125.665961437944
290 127.735789249689
};
\addlegendentry{Riemannian grads VI}
\addplot [very thick, color2]
table {%
0 6613.39016311349
10 89.206156585692
20 77.4504301619699
30 75.8192024089246
40 75.4020551178109
50 75.2355966400175
60 75.1425847205127
70 75.0796594389785
80 75.0331942170047
90 74.9976164649726
100 74.9699764297212
110 74.948377265207
120 74.9314573574065
130 74.9181883351683
140 74.9077768620343
150 74.8996055852716
160 74.8931922657239
170 74.8881594531749
180 74.8842113223865
190 74.8811157756424
200 74.8786905486862
210 74.8767923979805
220 74.8753086712341
230 74.8741507256193
240 74.8732487812018
250 74.8725478908861
260 74.8720047805223
270 74.871585368285
280 74.8712628151577
290 74.8710159912734
};
\addlegendentry{variational Gauss-Newton}
\addplot [very thick, color3]
table {%
0 7888.26076368036
10 6375.69548169165
20 4817.64361084951
30 2815.05455141668
40 1465.77320674349
50 909.619210508585
60 689.818177227493
70 546.56499451249
80 428.161109607203
90 326.474580001689
100 243.90249152413
110 183.323366777892
120 143.014371402199
130 118.457426274276
140 104.814930443039
150 97.4093936175708
160 93.1044186553526
170 90.5602884827668
180 89.0404746123332
190 88.1950904017969
200 87.7733692967948
210 87.6786966726134
220 87.9248586488099
230 88.1934649040795
240 88.4215024561879
250 88.6170957243979
260 88.7903473403339
270 89.0698426487984
280 89.1784555089567
290 89.3204780801995
};
\addlegendentry{variational quasi-Newton}
\addplot [very thick, black, dash pattern=on 1pt off 3pt on 3pt off 3pt]
table {%
0 9341.96356224049
10 5878.08886751283
20 4525.18373256778
30 3368.51104657699
40 2618.04668034867
50 2146.76025991158
60 1605.85583695315
70 1028.67424325461
80 479.162180873608
90 312.757701734911
100 211.623107268352
110 191.395158380061
120 155.346165097771
130 134.789138658024
140 124.928274262814
150 116.290638940917
160 110.355575858061
170 106.684071326203
180 103.220124838602
190 101.234647255487
200 101.209707234793
210 99.5421552170365
220 99.2927933043323
230 97.1250330674293
240 97.0592630529025
250 96.1401116458842
260 94.5404142733954
270 94.7861931602491
280 95.3370484711835
290 94.2660912925366
};
\addlegendentry{first-order VI}
\end{axis}

\end{tikzpicture}
	\end{subfigure}
	\hspace*{\fill}
	\begin{subfigure}[t]{.32\textwidth}
		% This file was created with tikzplotlib v0.9.14.
\begin{tikzpicture}

\definecolor{color0}{rgb}{0.12156862745098,0.466666666666667,0.705882352941177}
\definecolor{color1}{rgb}{1,0.498039215686275,0.0549019607843137}
\definecolor{color2}{rgb}{0.172549019607843,0.627450980392157,0.172549019607843}
\definecolor{color3}{rgb}{0.83921568627451,0.152941176470588,0.156862745098039}

\begin{axis}[
height=\figureheight,
tick pos=left,
title={Test NLPD (VI)},
width=\figurewidth,
x grid style={white!69.0196078431373!black},
xlabel={iteration number},
xmin=-14.5, xmax=304.5,
xtick style={color=black},
y grid style={white!69.0196078431373!black},
ymin=-0.2, ymax=0.6,
ytick style={color=black}
]
\addplot [very thick, color0, dashed]
table {%
0 2.94731672507759
10 -0.057176591349261
20 -0.0819790925435907
30 -0.0882787209746363
40 -0.0903626151498867
50 -0.0911183166788183
60 -0.0914061198639069
70 -0.0915210008809785
80 -0.0915704045158691
90 -0.0915948720156104
100 -0.0916095889444296
110 -0.0916204597776822
120 -0.0916297965612844
130 -0.0916384574138901
140 -0.0916467466593976
150 -0.0916547611458484
160 -0.0916625244336132
170 -0.0916700377466894
180 -0.0916772982579784
190 -0.091684304810233
200 -0.0916910590460935
210 -0.0916975650513725
220 -0.0917038286658029
230 -0.0917098568473825
240 -0.0917156571862757
250 -0.0917212375641408
260 -0.091726605927697
270 -0.0917317701442035
280 -0.0917367379127333
290 -0.0917415167123556
};
\addplot [very thick, color1, dashed]
table {%
0 10.9520437782745
10 6.85596561940655
20 2.78862389262785
30 0.208411971294231
40 -0.0598825671167384
50 -0.0887890429125738
60 -0.0936393462398088
70 -0.0963277414697413
80 -0.0998597414556761
90 -0.102654257308253
100 -0.105215839183498
110 -0.107160653186145
120 -0.108863263498736
130 -0.110302680209869
140 -0.11121146348717
150 -0.112052846754473
160 -0.11278811875827
170 -0.11343930346124
180 -0.11440867462908
190 -0.114941230018931
200 -0.1157093323907
210 -0.116324096374525
220 -0.116712669380873
230 -0.117303721478014
240 -0.117551186442313
250 -0.1177415764044
260 -0.11813472199114
270 -0.118266272645134
280 -0.118461209667006
290 -0.11875199988018
};
\addplot [very thick, color2]
table {%
0 3.09682171870145
10 -0.0539611030498949
20 -0.0818326092483394
30 -0.0885405497146896
40 -0.0908383398100323
50 -0.0917590832852156
60 -0.092169401749566
70 -0.092370742348945
80 -0.0924796148016662
90 -0.0925445095968753
100 -0.0925868925409307
110 -0.0926168247570048
120 -0.0926392987392286
130 -0.0926569446077466
140 -0.092671239342652
150 -0.0926830698299212
160 -0.0926930054829746
170 -0.092701435447187
180 -0.092708640462915
190 -0.092714832146386
200 -0.0927201754820147
210 -0.0927248023632153
220 -0.092728820180464
230 -0.0927323175550634
240 -0.0927353683529507
250 -0.0927380346120719
260 -0.0927403687502295
270 -0.0927424152747262
280 -0.0927442121331172
290 -0.0927457917968229
};
\addplot [very thick, color3]
table {%
0 10.2138778677376
10 8.04886307070652
20 5.89558392114983
30 3.17415246642055
40 1.37010583547456
50 0.666598466728396
60 0.444813031956046
70 0.31788370026826
80 0.210855640668919
90 0.114020845741265
100 0.0324419517811786
110 -0.028115862037546
120 -0.0674710199027464
130 -0.090535569518305
140 -0.103191929124243
150 -0.109652875794783
160 -0.112844423712334
170 -0.114543399198671
180 -0.115579303829516
190 -0.116412460781778
200 -0.117062108038687
210 -0.117698657034076
220 -0.11838762446431
230 -0.118867160992554
240 -0.119253485853624
250 -0.119547527860468
260 -0.119783453802952
270 -0.120005048918497
280 -0.120176947454866
290 -0.120422253153794
};
\addplot [very thick, black, dash pattern=on 1pt off 3pt on 3pt off 3pt]
table {%
0 2.0679091311975
10 1.18367189514632
20 0.862032399141001
30 0.750768279926852
40 0.748616982123503
50 0.706490298377876
60 0.573856395274478
70 0.408649588895487
80 0.19843398976727
90 0.127736465206161
100 0.0325671586431566
110 -0.000466578390920217
120 -0.0182269318946703
130 -0.0370286305027888
140 -0.0480912156167328
150 -0.0540010184584737
160 -0.0616069336027072
170 -0.0651967471140371
180 -0.0690090336216573
190 -0.0715408842183027
200 -0.0718196979141777
210 -0.0726735438482148
220 -0.074676137458429
230 -0.0750011914530467
240 -0.0750690483856116
250 -0.0763982409865499
260 -0.0786283487637149
270 -0.0770021052483443
280 -0.0769808448692298
290 -0.0782593262037825
};
\end{axis}

\end{tikzpicture}
	\end{subfigure}
	\hspace*{\fill}
	\begin{subfigure}[t]{.32\textwidth}
		\raggedleft
		% This file was created with tikzplotlib v0.9.14.
\begin{tikzpicture}

\definecolor{color0}{rgb}{0.12156862745098,0.466666666666667,0.705882352941177}
\definecolor{color1}{rgb}{1,0.498039215686275,0.0549019607843137}
\definecolor{color2}{rgb}{0.172549019607843,0.627450980392157,0.172549019607843}
\definecolor{color3}{rgb}{0.83921568627451,0.152941176470588,0.156862745098039}

\begin{axis}[
height=\figureheight,
tick pos=left,
title={Ground Truth RMSE (VI)},
width=\figurewidth,
x grid style={white!69.0196078431373!black},
xlabel={iteration number},
xmin=-14.5, xmax=304.5,
xtick style={color=black},
y grid style={white!69.0196078431373!black},
ymin=0, ymax=1.4,
ytick style={color=black}
]
\addplot [very thick, color0, dashed]
table {%
0 1.3167929914745
10 0.416690747238122
20 0.402925844150436
30 0.395803996656942
40 0.39147153750771
50 0.388550731363379
60 0.386393052228513
70 0.384663186561677
80 0.383181759474836
90 0.381850603113602
100 0.380614853976705
110 0.379442519900615
120 0.378315253621362
130 0.377222188868515
140 0.376156684815638
150 0.375114539730089
160 0.374092981805897
170 0.373090090838358
180 0.372104463214859
190 0.371135015858202
200 0.37018087024289
210 0.369241282946088
220 0.368315603472763
230 0.367403248209284
240 0.366503684006849
250 0.365616417576555
260 0.36474098843471
270 0.363876964047533
280 0.363023936360791
290 0.362181519218282
};
\addplot [very thick, color1, dashed]
table {%
0 1.3358278534243
10 1.45406371553743
20 1.01347910082116
30 0.615372866132014
40 0.452309332510237
50 0.381096042903994
60 0.342662449215464
70 0.314909769584026
80 0.294264492777906
90 0.279511265411166
100 0.268737536911875
110 0.260562494098934
120 0.253621318302689
130 0.247735884265989
140 0.242425787383045
150 0.23760379475207
160 0.233272990935161
170 0.229198813977973
180 0.225271466077753
190 0.221606871097481
200 0.218242368582866
210 0.2150591452716
220 0.212911293216197
230 0.210888680035027
240 0.209738234712802
250 0.208686381252654
260 0.20765857920573
270 0.206661887075554
280 0.205748270987122
290 0.204837967110986
};
\addplot [very thick, color2]
table {%
0 1.47240895852731
10 0.333549603412978
20 0.320501147910715
30 0.299789440530111
40 0.281894415613661
50 0.267080734940163
60 0.25472016032917
70 0.244304717116954
80 0.235458311490519
90 0.227900460080466
100 0.221416160295221
110 0.215835534121599
120 0.211020952373662
130 0.20685886171169
140 0.203254401610758
150 0.200127678316419
160 0.197411049796416
170 0.195047053519255
180 0.192986764606383
190 0.191188459093772
200 0.189616505842403
210 0.188240437983762
220 0.187034170235471
230 0.185975337372988
240 0.185044734588306
250 0.184225844036851
260 0.183504434429127
270 0.182868222508133
280 0.182306586886295
290 0.181810326107721
};
\addplot [very thick, color3]
table {%
0 1.3184989155841
10 1.19242553023413
20 1.0785877830424
30 0.949061200362433
40 0.841640392428222
50 0.775158418019399
60 0.730523941692379
70 0.688265345714619
80 0.64096469794839
90 0.585908694596239
100 0.523606623832134
110 0.457646862046393
120 0.393172412477548
130 0.335496756306486
140 0.288339448669551
150 0.252741682543787
160 0.226184763196064
170 0.20596898940457
180 0.190780522380633
190 0.17986742915443
200 0.172477707698416
210 0.167865592995738
220 0.16554269531948
230 0.164810104587053
240 0.164718842579113
250 0.165004543371559
260 0.165657356677614
270 0.166701582404125
280 0.168082981567461
290 0.169801340392379
};
\addplot [very thick, black, dash pattern=on 1pt off 3pt on 3pt off 3pt]
table {%
0 1.32970500643868
10 1.03038121384351
20 0.853522208224468
30 0.880154624165209
40 0.969588604862186
50 0.891552532197264
60 0.772774759264497
70 0.710098021399991
80 0.674870142504206
90 0.566679141698831
100 0.43491157933571
110 0.34233529791242
120 0.301899483850607
130 0.270137945841304
140 0.251583865545308
150 0.249954016341779
160 0.23965600507832
170 0.240088963961494
180 0.238446575793971
190 0.22947074206561
200 0.228626309542256
210 0.22651906491124
220 0.221640074458408
230 0.217718104543192
240 0.214026701206974
250 0.210254442825758
260 0.206154982414423
270 0.202489060489528
280 0.200835547459594
290 0.196449113190233
};
\end{axis}

\end{tikzpicture}
	\end{subfigure}\\
	\vspace{-1em}
	\begin{subfigure}[t]{.32\textwidth}
		\raggedright
		% This file was created with tikzplotlib v0.9.14.
\begin{tikzpicture}

\definecolor{color0}{rgb}{0.12156862745098,0.466666666666667,0.705882352941177}
\definecolor{color1}{rgb}{1,0.498039215686275,0.0549019607843137}
\definecolor{color2}{rgb}{0.172549019607843,0.627450980392157,0.172549019607843}
\definecolor{color3}{rgb}{0.83921568627451,0.152941176470588,0.156862745098039}

\begin{axis}[
height=\figureheight,
legend cell align={left},
legend style={fill opacity=0.8, draw opacity=1, text opacity=1, draw=white!80!black},
tick pos=left,
title={Training Loss (PEP)},
unbounded coords=jump,
width=\figurewidth,
x grid style={white!69.0196078431373!black},
xlabel={iteration number},
xmin=-14.5, xmax=304.5,
xtick style={color=black},
y grid style={white!69.0196078431373!black},
ymin=-50, ymax=1000,
ytick style={color=black}
]
\addplot [very thick, color0, dashed]
table {%
0 2059.92653618413
10 100.477377083198
20 94.1014820616792
30 93.046281214388
40 92.580447486061
50 92.2317710701819
60 91.9251250186482
70 91.6427774686113
80 91.3781671076329
90 91.127821488181
100 90.8895158622907
110 90.6616920732888
120 90.4432053761936
130 90.2331860437088
140 90.0309538071778
150 89.8359616790474
160 89.6477579405684
170 89.4659599923821
180 89.2902361778235
190 89.1202930925168
200 88.9558665737456
210 88.796715164272
220 88.6426158067213
230 88.4933604589039
240 88.3487536866545
250 88.2086108929567
260 88.0727569898674
270 87.9410253845749
280 87.8132571898972
290 87.6893005962143
};
\addlegendentry{heuristic PEP}
\addplot [very thick, color1, dashed]
table {%
0 8379.81546566209
10 7446.95426597325
20 6086.39674670304
30 3323.89394381425
40 1817.2130213511
50 1125.0128787214
60 647.810927420401
70 405.20848251705
80 201.482878306497
90 134.110982996619
100 105.600411487005
110 93.8506503964644
120 87.7447758619227
130 88.1536109771366
140 89.3163808989373
150 91.0472321768858
160 93.5994730040189
170 95.2561696911667
180 97.7440980005147
190 100.95181025714
200 103.909559236908
210 106.98162658305
220 109.46401198153
230 110.318356212826
240 114.78807236595
250 116.411297967425
260 117.634242446392
270 120.985797560168
280 122.512653781211
290 124.116234238661
};
\addlegendentry{Riemannian grads PEP}
\addplot [very thick, color2, forget plot]
table {%
0 nan
10 nan
20 nan
30 nan
40 nan
50 nan
60 nan
70 nan
80 nan
90 nan
100 nan
110 nan
120 nan
130 nan
140 nan
150 nan
160 nan
170 nan
180 nan
190 nan
200 nan
210 nan
220 nan
230 nan
240 nan
250 nan
260 nan
270 nan
280 nan
290 nan
};
\addplot [very thick, color3]
table {%
0 7332.46200717665
10 7289.0501181558
20 6587.76615772302
30 5510.39023414866
40 3892.1955332386
50 2354.04330370631
60 1384.57113960844
70 990.176337894176
80 820.042738992517
90 690.537753573468
100 570.898137974116
110 457.042043272879
120 352.494854300787
130 264.622475712663
140 198.31726053239
150 153.267853816898
160 125.244665857041
170 108.982803277061
180 99.5375867737865
190 93.5154592192829
200 89.6232605462125
210 87.1606927473195
220 85.7254282637256
230 84.995543454575
240 84.6919018182291
250 84.6247734977147
260 84.5889896772795
270 84.6110920153679
280 84.6482243342722
290 84.8077242613645
};
\addlegendentry{PEP quasi-Newton}
\end{axis}

\end{tikzpicture}
	\end{subfigure}
	\hspace*{\fill}
	\begin{subfigure}[t]{.32\textwidth}
		% This file was created with tikzplotlib v0.9.14.
\begin{tikzpicture}

\definecolor{color0}{rgb}{0.12156862745098,0.466666666666667,0.705882352941177}
\definecolor{color1}{rgb}{1,0.498039215686275,0.0549019607843137}
\definecolor{color2}{rgb}{0.172549019607843,0.627450980392157,0.172549019607843}
\definecolor{color3}{rgb}{0.83921568627451,0.152941176470588,0.156862745098039}

\begin{axis}[
height=\figureheight,
tick pos=left,
title={Test NLPD (PEP)},
unbounded coords=jump,
width=\figurewidth,
x grid style={white!69.0196078431373!black},
xlabel={iteration number},
xmin=-14.5, xmax=304.5,
xtick style={color=black},
y grid style={white!69.0196078431373!black},
ymin=-0.2, ymax=0.6,
ytick style={color=black}
]
\addplot [very thick, color0, dashed]
table {%
0 1.50030957000432
10 -0.0566739102326967
20 -0.079940928126939
30 -0.085792786231123
40 -0.0877422044440125
50 -0.0884880440133557
60 -0.0888135861959902
70 -0.0889819868942823
80 -0.0890882071329733
90 -0.0891681060528422
100 -0.0892357666822814
110 -0.0892969620217578
120 -0.089354187637323
130 -0.0894085993559461
140 -0.0894607818578219
150 -0.0895110635782945
160 -0.0895596505768869
170 -0.089606685955765
180 -0.0896522776145264
190 -0.0896964388722234
200 -0.0897393467846856
210 -0.0897810545717247
220 -0.0898216045922668
230 -0.0898610410463441
240 -0.0898994063068081
250 -0.0899367402112309
260 -0.089973080073935
270 -0.0900084608817429
280 -0.0900429155270702
290 -0.0900764750324459
};
\addplot [very thick, color1, dashed]
table {%
0 11.0168720040291
10 9.68436104491853
20 7.78672256376146
30 3.68185848042189
40 1.55890633938275
50 0.872173470105949
60 0.416356874841675
70 0.214846082561371
80 0.0340255879450982
90 -0.0528522233534418
100 -0.0777881576642916
110 -0.0921833109598767
120 -0.103815611403501
130 -0.106024318621946
140 -0.108414494589589
150 -0.110891598256847
160 -0.113000592755617
170 -0.114525387451908
180 -0.11677418325222
190 -0.118936588001343
200 -0.120368113771084
210 -0.12152418914601
220 -0.12259242512975
230 -0.123262991219405
240 -0.124757206700531
250 -0.125437709887674
260 -0.12605835140863
270 -0.127218654712301
280 -0.127607946382653
290 -0.127613014910306
};
\addplot [very thick, color2]
table {%
0 nan
10 nan
20 nan
30 nan
40 nan
50 nan
60 nan
70 nan
80 nan
90 nan
100 nan
110 nan
120 nan
130 nan
140 nan
150 nan
160 nan
170 nan
180 nan
190 nan
200 nan
210 nan
220 nan
230 nan
240 nan
250 nan
260 nan
270 nan
280 nan
290 nan
};
\addplot [very thick, color3]
table {%
0 7.81539672891838
10 9.34536893481516
20 8.33191497408497
30 6.87038244939162
40 4.70609563015438
50 2.64732439261482
60 1.34221815552145
70 0.81342529156184
80 0.621374816085562
90 0.492977218790585
100 0.376429177464372
110 0.264107829800617
120 0.159631258208542
130 0.0705694492541003
140 0.0026432305501305
150 -0.0435107171071038
160 -0.0718073443324976
170 -0.0878753750663054
180 -0.0965309896402751
190 -0.101065398865542
200 -0.103505972838783
210 -0.104924667641892
220 -0.105987018158053
230 -0.106845240508678
240 -0.107550777712734
250 -0.108097537867253
260 -0.1085077750535
270 -0.10894803001142
280 -0.109398425732671
290 -0.109741747110885
};
\end{axis}

\end{tikzpicture}
	\end{subfigure}
	\hspace*{\fill}
	\begin{subfigure}[t]{.32\textwidth}
		\raggedleft
		% This file was created with tikzplotlib v0.9.14.
\begin{tikzpicture}

\definecolor{color0}{rgb}{0.12156862745098,0.466666666666667,0.705882352941177}
\definecolor{color1}{rgb}{1,0.498039215686275,0.0549019607843137}
\definecolor{color2}{rgb}{0.172549019607843,0.627450980392157,0.172549019607843}
\definecolor{color3}{rgb}{0.83921568627451,0.152941176470588,0.156862745098039}

\begin{axis}[
height=\figureheight,
tick pos=left,
title={Ground Truth RMSE (PEP)},
unbounded coords=jump,
width=\figurewidth,
x grid style={white!69.0196078431373!black},
xlabel={iteration number},
xmin=-14.5, xmax=304.5,
xtick style={color=black},
y grid style={white!69.0196078431373!black},
ymin=0, ymax=1.4,
ytick style={color=black}
]
\addplot [very thick, color0, dashed]
table {%
0 1.28614335514033
10 0.738053909248581
20 0.732505921632251
30 0.727942262140349
40 0.723635718787195
50 0.719434071876802
60 0.715315173894675
70 0.711278098463981
80 0.707322425655683
90 0.703445838287568
100 0.699644865056362
110 0.695915676174923
120 0.692254525663028
130 0.688657936375217
140 0.685122749608366
150 0.681646114882367
160 0.678225458219189
170 0.674858446103212
180 0.671542951875099
190 0.6682770183768
200 0.665058851311478
210 0.66188680121751
220 0.65875932000861
230 0.655674960944132
240 0.652632368123266
250 0.649630266341308
260 0.646667452443885
270 0.643742788007498
280 0.640855193105181
290 0.638003640960671
};
\addplot [very thick, color1, dashed]
table {%
0 1.26745149367197
10 1.5988577191765
20 1.50841932717891
30 1.30809550161472
40 1.27730543765112
50 1.16702708858684
60 1.04108697832557
70 0.908889029747916
80 0.788786838626715
90 0.715677016469635
100 0.669742788617357
110 0.624619450333759
120 0.588524997458091
130 0.555069265006281
140 0.525634388258941
150 0.498223948258371
160 0.473778003759226
170 0.451889282978905
180 0.432159725769799
190 0.414171456818983
200 0.398654085913867
210 0.383506032935874
220 0.369005295812477
230 0.356721704202356
240 0.344860932285045
250 0.333110013283646
260 0.322773395834854
270 0.312641107006662
280 0.302886956596117
290 0.293930981183796
};
\addplot [very thick, color2]
table {%
0 nan
10 nan
20 nan
30 nan
40 nan
50 nan
60 nan
70 nan
80 nan
90 nan
100 nan
110 nan
120 nan
130 nan
140 nan
150 nan
160 nan
170 nan
180 nan
190 nan
200 nan
210 nan
220 nan
230 nan
240 nan
250 nan
260 nan
270 nan
280 nan
290 nan
};
\addplot [very thick, color3]
table {%
0 1.34959928120118
10 1.26124642583971
20 1.20277601026295
30 1.12701175578228
40 1.02620682801928
50 0.920755642268773
60 0.839431903848627
70 0.794114919937843
80 0.764417526256316
90 0.735505447684299
100 0.70191592438593
110 0.660562467607147
120 0.610139933752453
130 0.551874804082979
140 0.489553609026222
150 0.42827913473858
160 0.372988259609496
170 0.327221419400332
180 0.291473533014068
190 0.26297408672142
200 0.239598996546411
210 0.220323164996851
220 0.2047832523038
230 0.192839420485765
240 0.184252291425299
250 0.178653058201921
260 0.175037053563643
270 0.172426444788041
280 0.170235088262673
290 0.168371216009313
};
\end{axis}

\end{tikzpicture}
	\end{subfigure}\\
	\vspace{-1em}
	\begin{subfigure}[t]{.32\textwidth}
		\raggedright
		% This file was created with tikzplotlib v0.9.14.
\begin{tikzpicture}

\definecolor{color0}{rgb}{0.12156862745098,0.466666666666667,0.705882352941177}
\definecolor{color1}{rgb}{1,0.498039215686275,0.0549019607843137}
\definecolor{color2}{rgb}{0.172549019607843,0.627450980392157,0.172549019607843}
\definecolor{color3}{rgb}{0.83921568627451,0.152941176470588,0.156862745098039}
\definecolor{color4}{rgb}{0.580392156862745,0.403921568627451,0.741176470588235}

\begin{axis}[
height=\figureheight,
legend cell align={left},
legend style={fill opacity=0.8, draw opacity=1, text opacity=1, draw=white!80!black},
tick pos=left,
title={Training Loss (PL2)},
width=\figurewidth,
x grid style={white!69.0196078431373!black},
xlabel={iteration number},
xmin=-14.5, xmax=304.5,
xtick style={color=black},
y grid style={white!69.0196078431373!black},
ymin=-50, ymax=1000,
ytick style={color=black}
]
\addplot [very thick, color0, dashed]
table {%
0 773.634605959304
10 81.6655648671685
20 78.4528287872797
30 78.959198268557
40 79.7745763021277
50 80.628841677475
60 81.4957217165701
70 82.3820650773782
80 83.287516443146
90 84.1740440274414
100 84.9638722105869
110 85.589403773919
120 86.033478402883
130 86.3195835390147
140 86.4872158772591
150 86.5765539176182
160 86.620223258549
170 86.6401397853238
180 86.6487900834484
190 86.6524329784256
200 86.6539397018594
210 86.6545566874299
220 86.6548076952881
230 86.6549108633944
240 86.6549528592641
250 86.654954332012
260 86.654966978756
270 86.6550592905078
280 86.6551236460646
290 86.6551333406223
};
\addlegendentry{PL}
\addplot [very thick, color1, dashed]
table {%
0 7932.36106773019
10 3322.52137266569
20 586.439137077033
30 103.191545321007
40 82.4522167787288
50 78.5604365554724
60 77.7033630859987
70 77.9659537657
80 80.5948537753994
90 82.9056424534236
100 85.9194807386359
110 90.792907045568
120 93.5984768878798
130 99.1326493475945
140 104.06775381851
150 108.785115951662
160 115.245908333461
170 120.851203235483
180 124.2659567974
190 130.163672476782
200 134.692677832836
210 140.180155480203
220 146.180965446696
230 148.061996726674
240 151.943134924477
250 156.637952481768
260 159.748131107606
270 162.676103393056
280 165.394629612041
290 168.495996422649
};
\addlegendentry{Riemannian grads PL2}
\addplot [very thick, color2]
table {%
0 426.84703558095
10 79.6495130223223
20 76.4758836408377
30 76.5351507602725
40 77.1119158001844
50 77.8661383195248
60 78.7113989967736
70 79.6101180436222
80 80.5398821901098
90 81.4858199371072
100 82.4376785186207
110 83.3873337870939
120 84.3264261715996
130 85.2457086763164
140 86.1366970849429
150 86.9931496166074
160 87.8105751894641
170 88.5853361825227
180 89.3144868128293
190 89.9957490645545
200 90.6274191672844
210 91.2083198756941
220 91.7377736920787
230 92.2155887588196
240 92.6420585775584
250 93.0179737531578
260 93.3446439052557
270 93.6239257921864
280 93.8582510958483
290 94.0506439793137
};
\addlegendentry{PL2 Gauss-Newton}
\addplot [very thick, color3]
table {%
0 8108.12371825716
10 6381.4784004346
20 4659.0883632496
30 2668.51362516942
40 1403.6150701311
50 876.154015278918
60 669.684779114743
70 533.050414817321
80 418.065638201707
90 318.572964423681
100 237.871269719791
110 178.972453028792
120 140.067937146248
130 116.563715974448
140 103.5309373086
150 96.4308002839772
160 92.3118114194777
170 89.8752380423697
180 88.388555441765
190 87.5500163926937
200 87.2038506941297
210 87.0582016880507
220 86.952723336058
230 86.8488750072898
240 86.7706514225499
250 86.829170256792
260 87.0976497469867
270 87.6627175810687
280 88.4145268079076
290 89.2450355034284
};
\addlegendentry{PL2 quasi-Newton}
\addplot [very thick, color4, dashed]
table {%
0 1497.63709058479
10 125.342003034786
20 97.6922434728585
30 90.9864984083935
40 88.345091320143
50 87.0183299512957
60 86.2195606424237
70 85.6632196945814
80 85.2327787645439
90 84.8766444025903
100 84.5699847072206
110 84.299728726213
120 84.0583957499366
130 83.8412389416935
140 83.6450040881005
150 83.4672694135814
160 83.3061073279488
170 83.1599058300204
180 83.0272712465789
190 82.9069851621499
200 82.7979091410388
210 82.6990643485203
220 82.6095480851612
230 82.528534602288
240 82.4552703304612
250 82.3890685904753
260 82.3293016796893
270 82.2753962686526
280 82.2268287747362
290 82.1831210867914
};
\addlegendentry{heuristic PL2}
\end{axis}

\end{tikzpicture}
	\end{subfigure}
	\hspace*{\fill}
	\begin{subfigure}[t]{.32\textwidth}
		% This file was created with tikzplotlib v0.9.14.
\begin{tikzpicture}

\definecolor{color0}{rgb}{0.12156862745098,0.466666666666667,0.705882352941177}
\definecolor{color1}{rgb}{1,0.498039215686275,0.0549019607843137}
\definecolor{color2}{rgb}{0.172549019607843,0.627450980392157,0.172549019607843}
\definecolor{color3}{rgb}{0.83921568627451,0.152941176470588,0.156862745098039}
\definecolor{color4}{rgb}{0.580392156862745,0.403921568627451,0.741176470588235}

\begin{axis}[
height=\figureheight,
tick pos=left,
title={Test NLPD (PL2)},
width=\figurewidth,
x grid style={white!69.0196078431373!black},
xlabel={iteration number},
xmin=-14.5, xmax=304.5,
xtick style={color=black},
y grid style={white!69.0196078431373!black},
ymin=-0.2, ymax=0.6,
ytick style={color=black}
]
\addplot [very thick, color0, dashed]
table {%
0 1.02467212892926
10 -0.0578070521617263
20 -0.078536417507457
30 -0.083359058699071
40 -0.0847698235432348
50 -0.0850771588964386
60 -0.08489449889592
70 -0.0843914745883732
80 -0.0836284280024624
90 -0.0826955627967229
100 -0.0817536046531039
110 -0.0809589749871556
120 -0.0803804167998756
130 -0.0800054807262038
140 -0.0797866883545654
150 -0.07967172327285
160 -0.0796172151433615
170 -0.0795937038191709
180 -0.0795843845325135
190 -0.0795809715540808
200 -0.0795798232028297
210 -0.0795794778429885
220 -0.0795793931968838
230 -0.079579381645562
240 -0.0795793870298715
250 -0.0795794023164825
260 -0.079579407272849
270 -0.079579380389739
280 -0.0795794261681146
290 -0.0795792623126432
};
\addplot [very thick, color1, dashed]
table {%
0 10.473897268143
10 4.36430360254302
20 0.654580561545172
30 -0.0620284332526936
40 -0.0883224010445937
50 -0.0941535401210158
60 -0.0977384857405934
70 -0.101174800723226
80 -0.105363027457831
90 -0.109578539441322
100 -0.113645803802369
110 -0.117456235909061
120 -0.120248050258025
130 -0.123610242588223
140 -0.126348199018698
150 -0.128398062149339
160 -0.130987883961156
170 -0.132331060064581
180 -0.133583845199715
190 -0.134649611825011
200 -0.135514278895178
210 -0.136613473566102
220 -0.137459768452587
230 -0.137664344678834
240 -0.138190428636279
250 -0.138740336822263
260 -0.138998587043004
270 -0.139457748902344
280 -0.139591812985878
290 -0.139877419118822
};
\addplot [very thick, color2]
table {%
0 0.532389957265313
10 -0.058180391109279
20 -0.0782533687534692
30 -0.0836069490438813
40 -0.0853344819442082
50 -0.0858755443217294
60 -0.0859753517488702
70 -0.0858847029175502
80 -0.0856907738780835
90 -0.0854246198891462
100 -0.0850993539487909
110 -0.0847254663633774
120 -0.0843170681071322
130 -0.083891354322711
140 -0.0834636717366395
150 -0.0830444334555267
160 -0.0826403799597367
170 -0.0822563842037756
180 -0.0818958808766051
190 -0.0815611008336751
200 -0.0812533956849179
210 -0.0809734560607955
220 -0.0807214675188396
230 -0.0804972205416932
240 -0.0803001807599536
250 -0.0801295283539164
260 -0.0799841738500533
270 -0.0798627579512545
280 -0.0797636437523162
290 -0.0796849110854505
};
\addplot [very thick, color3]
table {%
0 10.590722773943
10 8.1343009151207
20 5.81153659927465
30 3.10542829876925
40 1.37490042657408
50 0.67750179820527
60 0.454726576295194
70 0.326248169902535
80 0.217841114561252
90 0.119812527124537
100 0.0371408300355598
110 -0.0242960248636016
120 -0.0643364416986618
130 -0.0878160571572169
140 -0.100855908797666
150 -0.107809516485179
160 -0.111383945213137
170 -0.113427522490425
180 -0.114802176947312
190 -0.115886634379462
200 -0.116797178099899
210 -0.117384128474718
220 -0.117689847770986
230 -0.11779912334903
240 -0.117774552721055
250 -0.117700046303736
260 -0.117648747353155
270 -0.117695271002915
280 -0.117780145287386
290 -0.117834155073244
};
\addplot [very thick, color4, dashed]
table {%
0 1.91602037779864
10 -0.0251532551309915
20 -0.0738357427177618
30 -0.0847864527493416
40 -0.0879082805433602
50 -0.0888300925793391
60 -0.0891000520298266
70 -0.0891841615274341
80 -0.0892192788820321
90 -0.0892429631686949
100 -0.0892641689602529
110 -0.0892845960559909
120 -0.0893040576400611
130 -0.0893223851986156
140 -0.0893395139744561
150 -0.0893554489942712
160 -0.0893702352695239
170 -0.0893839387470213
180 -0.0893966312823386
190 -0.0894082421379654
200 -0.0894190207047625
210 -0.0894290306428765
220 -0.089438306750719
230 -0.0894468508861242
240 -0.0894547203088263
250 -0.0894620249155413
260 -0.0894687733083672
270 -0.0894749931278062
280 -0.0894807161145858
290 -0.0894859736627878
};
\end{axis}

\end{tikzpicture}
	\end{subfigure}
	\hspace*{\fill}
	\begin{subfigure}[t]{.32\textwidth}
		\raggedleft
		% This file was created with tikzplotlib v0.9.14.
\begin{tikzpicture}

\definecolor{color0}{rgb}{0.12156862745098,0.466666666666667,0.705882352941177}
\definecolor{color1}{rgb}{1,0.498039215686275,0.0549019607843137}
\definecolor{color2}{rgb}{0.172549019607843,0.627450980392157,0.172549019607843}
\definecolor{color3}{rgb}{0.83921568627451,0.152941176470588,0.156862745098039}
\definecolor{color4}{rgb}{0.580392156862745,0.403921568627451,0.741176470588235}

\begin{axis}[
height=\figureheight,
tick pos=left,
title={Ground Truth RMSE (PL2)},
width=\figurewidth,
x grid style={white!69.0196078431373!black},
xlabel={iteration number},
xmin=-14.5, xmax=304.5,
xtick style={color=black},
y grid style={white!69.0196078431373!black},
ymin=0, ymax=1.4,
ytick style={color=black}
]
\addplot [very thick, color0, dashed]
table {%
0 1.06613209615189
10 0.247866018057652
20 0.297318935312808
30 0.359330370679997
40 0.415138016604312
50 0.463447598758275
60 0.503872053213353
70 0.536383721485223
80 0.561450624319975
90 0.580143931189403
100 0.5937498054655
110 0.603272290619372
120 0.609459745156407
130 0.613077786826663
140 0.614949212591938
150 0.615801695177421
160 0.616146807608553
170 0.616274273313758
180 0.616319130814145
190 0.616334957675486
200 0.616340780793424
210 0.61634304313378
220 0.616343961676836
230 0.616344344168932
240 0.616344499304577
250 0.616344586465252
260 0.616344596187541
270 0.616344545563518
280 0.616344684835024
290 0.616344130810576
};
\addplot [very thick, color1, dashed]
table {%
0 1.35681174192566
10 0.90926738190848
20 0.508110787264799
30 0.377372090280106
40 0.321651487489296
50 0.27223890933521
60 0.232949911041881
70 0.20511467395956
80 0.186280405073777
90 0.174571269864661
100 0.167948800595239
110 0.164651848721678
120 0.163445814861346
130 0.163401252970426
140 0.163960413355148
150 0.164772345438976
160 0.165660097519414
170 0.166500701588903
180 0.167291178595845
190 0.168007363170831
200 0.168647611796784
210 0.169219237996271
220 0.169708681373809
230 0.170136653323812
240 0.170510260949122
250 0.170857671245025
260 0.171169484860166
270 0.171447701522637
280 0.171622973163728
290 0.171793987221585
};
\addplot [very thick, color2]
table {%
0 0.646106612202245
10 0.220521858299569
20 0.166235103822012
30 0.187168847120803
40 0.238434913400603
50 0.298382794801829
60 0.359260021174579
70 0.418183469928458
80 0.473997790387172
90 0.526299127384246
100 0.575084834308704
110 0.620582503878031
120 0.663109277449593
130 0.70294472861678
140 0.740277886092643
150 0.775236246261463
160 0.807916959050883
170 0.838388784322013
180 0.86669653486845
190 0.89287129957803
200 0.916937133940476
210 0.938916251374381
220 0.958833423868281
230 0.976719545775279
240 0.992614697798821
250 1.00657087418836
260 1.01865447404353
270 1.02894855084132
280 1.03755469113351
290 1.0445942485248
};
\addplot [very thick, color3]
table {%
0 1.32411352271529
10 1.19101813303093
20 1.07448326298406
30 0.947534661607294
40 0.843629050570228
50 0.778449994922557
60 0.734249677043767
70 0.69215148962538
80 0.644901826574066
90 0.589835716729807
100 0.527514224148183
110 0.46163624971022
120 0.397500651781813
130 0.340543943659558
140 0.294387085525862
150 0.25979625201256
160 0.234577169569135
170 0.216506801287913
180 0.204403765033312
190 0.197475101650194
200 0.194951109895452
210 0.196142252878896
220 0.200526362109973
230 0.207839256818603
240 0.217771039586025
250 0.229938416749191
260 0.243904671152615
270 0.259209116187066
280 0.275381213824151
290 0.291999201700551
};
\addplot [very thick, color4, dashed]
table {%
0 1.12861247131437
10 0.347710724420668
20 0.286715119630894
30 0.260125341004225
40 0.246673369148425
50 0.23951958644582
60 0.235439295746837
70 0.232840591979556
80 0.230968110883421
90 0.229474611699204
100 0.228202483837934
110 0.227079770329191
120 0.226073164618358
130 0.225166599553515
140 0.224350995072074
150 0.223620205011161
160 0.222969187085185
170 0.222393245877624
180 0.221887765886123
190 0.221448356545423
200 0.221070463359387
210 0.220749359099542
220 0.220480777229866
230 0.220260670502639
240 0.220085196666622
250 0.219950734874436
260 0.219853897889956
270 0.219791536434816
280 0.219760736893331
290 0.219758814608779
};
\end{axis}

\end{tikzpicture}
	\end{subfigure}\\
	\vspace{-0.2cm}
	\caption{Bayesian amplitude demodulation results. Mean of 4-fold cross validation shown. Whilst the test NLPD is similar for most methods, those capable of capturing the cross-covariance between latent components obtain better RMSE relative to the ground truth. PL quasi-Newton results not plotted since it is identical to PL2 quasi-Newton for this model.} \label{fig:product-results}
\end{figure*}

\subsubsection{Bayesian Amplitude Demodulation: the Product Likelihood}

Next we apply our methods to the model of \citet{turner2011demodulation}, which assumes an observed signal is produced by the product of a periodic component and a positive amplitude envelope. The task is to uncover these latent components from the signal alone. We use the following generative model,
\begin{gather}
	\begin{aligned}
		&f_1(\cdot) \sim \GP(0, \kappa_1(\cdot, \cdot)) \, , \quad f_2(\cdot) \sim \GP(0, \kappa_2(\cdot, \cdot)) \, ,\\
		&\vy_n \mid f_1(\MX_n), f_2(\MX_n) \sim \N(\vy_n \mid f_1(\MX_n)\phi(f_2(\MX_n)) , \, \sigma^2) \, ,
	\end{aligned}
\end{gather}
where $\phi(\cdot)$ is again the softplus function to ensure the amplitude is positive. Again we have $\fdim=2$, and $\ydim=1$. $\kappa_1$ is an oscillator kernel made up of the product of the cosine kernel and the Mat\'ern-$\nicefrac{3}{2}$: $\kappa_1(x, x') = \cos(\omega(x-x')) \kappa_{\text{Mat-}\nicefrac{3}{2}}(x,x')$ with $\omega=\frac{2}{5}\pi$, unit variance and a lengthscale of 500. We consider $N=1000$ evenly spaced inputs in the range [0, 200], therefore this amounts to an almost perfectly sinusoidal prior. $\kappa_2$ is a Mat\'ern-$\nicefrac{5}{2}$ kernel with a lengthscale of 3 and a variance of 2. The likelihood noise variance is $\sigma^2=0.1$.

We use a learning rate of $\rho=0.1$ and a quasi-Newton damping rate of $\xi=0.5$. We again use Gauss--Hermite integration to solve the intractable integrals. We draw a sample from the model to be used as the training and test data, and the hyperparameters are fixed to their true values. The aim of Bayesian amplitude demodulation is to uncover the ground truth latent functions whilst characterising the model uncertainty. \cref{fig:product-plot} shows an example inference result, and we plot the 4-fold cross validation results in \cref{fig:product-results} where we measure both the test NLPD and the RMSE of the posterior mean relative to the ground truth components. Whilst the test NLPD results show similar behaviour to the heteroscedastic noise task, we can see that the methods which take into account the cross-covariance between the latent components are able to better recover the ground truth oscillator and amplitude envelope.

\subsubsection{The Gaussian Process Regression Network}\label{sec:gprn}

The Gaussian process regression network \citep[GPRN,][]{wilson2012gaussian} is a multi-output GP model capable of capturing complex time-varying dependencies between observation streams. We use the following generative model,
\begin{gather}\label{eq:gprn}
	\begin{aligned}
		&f_i(\cdot) \sim \GP(0, \kappa_f(\cdot, \cdot)) \, , \quad i=1,2, \\
		&W_{j,i}(\cdot) \sim \GP(0, \kappa_W(\cdot, \cdot)) \, ,\quad j=1,\dots,3, \,\, i=1,2, \\
		&\vy_n \mid f(\MX_n), W(\MX_n) \sim \N(\vy_n \mid W(\MX_n)f(\MX_n) , \, \Sigma) \, ,
	\end{aligned}
\end{gather}
where $\vy_n \in \R^{3\times 1}$. Inference in \cref{eq:gprn} is notoriously difficult due to the nonlinear interaction of many components (there are 8 latent Gaussian processes, $\fdim=8$, $\ydim=3$). $\kappa_f$ and $\kappa_W$ are Mat\'ern-$\nicefrac{5}{2}$ kernels with unit variance. $\kappa_f$ and $\kappa_W$ have lengthscales of 10 and 70 respectively. We assume correlated observation noise, 
$$\Sigma=\left(\begin{matrix} \phantom{-}0.02\phantom{5} & -0.015 & -0.005 \\ -0.015 & \phantom{-}0.04\phantom{5} & \phantom{-}0.01\phantom{5} \\ -0.005 & \phantom{-}0.01\phantom{5} & \phantom{-}0.06\phantom{5} \end{matrix}\right),$$
and draw four samples from the model to be used as separate data sets, with 400 time steps (each with 3 outputs, giving 1200 total data points) evenly spaced in the range $[-17, 147]$. When the output streams are partially observed, the GPRN can be a powerful model for interpolation of missing data, so we remove the middle third of data for two of the three output streams and then compute the NLPD of the removed data as well as the RMSE of the posterior mean relative to the ground truth. We use a learning rate of $\rho=0.3$ and a quasi-Newton damping rate of $\xi=0.3$. The hyperparameters are fixed to their true values. Since the intractable integrals required for the VI-, EP- and PL-based methods are now 8-dimensional, we use the 5\textsuperscript{th}-order unscented transform \citep{McNamee1967} to approximate them instead of Gauss--Hermite.

\begin{figure*}[t!]
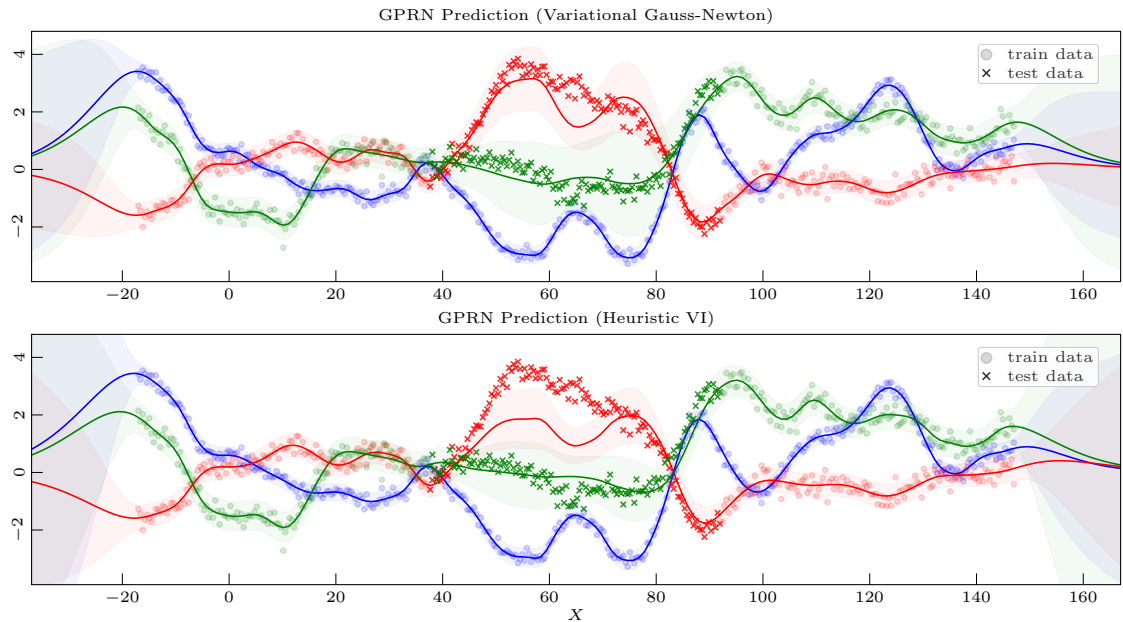

	\centering\tiny
	\pgfplotsset{yticklabel style={rotate=90}, ylabel style={yshift=0pt},scale only axis,axis on top,title style={yshift=-6pt}, xlabel style={yshift=3pt}}
	\pgfplotsset{legend style={inner xsep=1pt, inner ysep=1pt, row sep=0pt},legend style={at={(0.98,0.95)},anchor=north east},legend style={rounded corners=1pt}}
	\setlength{\figurewidth}{0.95\textwidth}
	\setlength{\figureheight}{.23\figurewidth}
	\begin{subfigure}[t]{\textwidth}
		\input{./fig/gprn-plot-vgn.tex}
	\end{subfigure}\\
	\vspace{-0.25em}
	\begin{subfigure}[t]{\textwidth}
		\input{./fig/gprn-plot-vi.tex}
	\end{subfigure}
	\vspace{-0.5em}
	\caption{Example GPRN prediction results. The variational Gauss--Newton method is capable of learning meaningful dependencies between the output streams, and hence is able to interpolate accurately. The heuristic VI approach, which assumes independence between the posterior latent processes, makes less accurate predictions.} \label{fig:gprn-plot}
\end{figure*}

\begin{figure*}[t!]
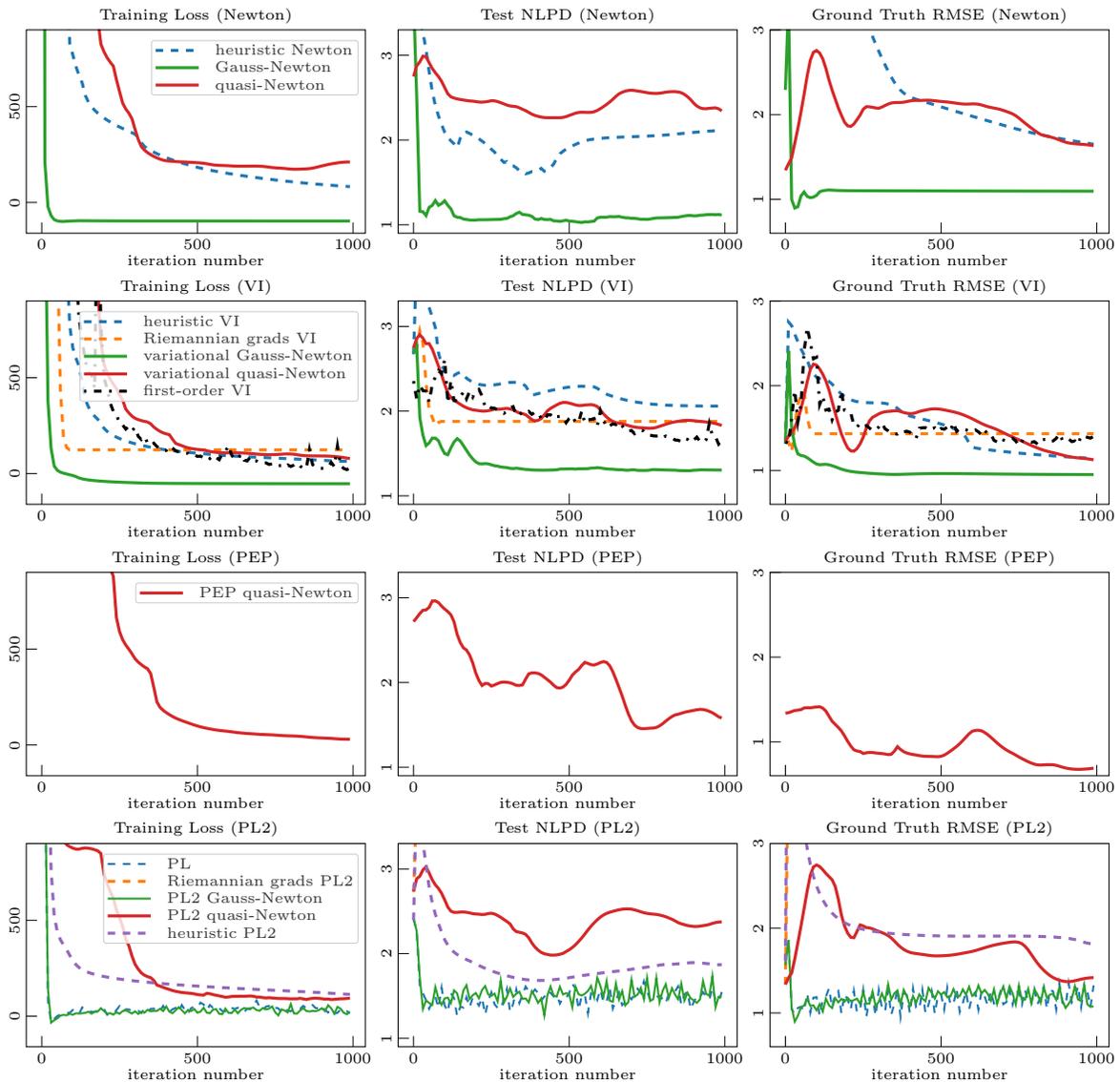

	\centering\tiny
	\pgfplotsset{yticklabel style={rotate=90}, ylabel style={yshift=0pt},scale only axis,axis on top,title style={yshift=-6pt}, xlabel style={yshift=3pt}}
	\pgfplotsset{legend style={inner xsep=1pt, inner ysep=1pt, row sep=0pt},legend style={at={(0.98,0.95)},anchor=north east},legend style={rounded corners=1pt}}
	\setlength{\figurewidth}{.31\textwidth}
	\setlength{\figureheight}{.6\figurewidth}
	\begin{subfigure}[t]{.32\textwidth}
		\raggedright
		% This file was created with tikzplotlib v0.9.14.
\begin{tikzpicture}

\definecolor{color0}{rgb}{0.12156862745098,0.466666666666667,0.705882352941177}
\definecolor{color1}{rgb}{1,0.498039215686275,0.0549019607843137}
\definecolor{color2}{rgb}{0.172549019607843,0.627450980392157,0.172549019607843}
\definecolor{color3}{rgb}{0.83921568627451,0.152941176470588,0.156862745098039}

\begin{axis}[
height=\figureheight,
legend cell align={left},
legend style={fill opacity=0.8, draw opacity=1, text opacity=1, draw=white!80!black},
tick pos=left,
title={Training Loss (Newton)},
unbounded coords=jump,
width=\figurewidth,
x grid style={white!69.0196078431373!black},
xlabel={iteration number},
xmin=-49.5, xmax=1039.5,
xtick style={color=black},
y grid style={white!69.0196078431373!black},
ymin=-160, ymax=900,
ytick style={color=black}
]
\addplot [very thick, color0, dashed]
table {%
0 26075.5218703147
10 1609.88345107299
20 1448.27154192424
30 1368.13354288839
40 1316.02953379963
50 1277.76177331795
60 1248.15137113129
70 1224.48289821011
80 1191.36037779605
90 829.308739638906
100 768.958891588788
110 708.276897751431
120 662.161591830025
130 623.380171305028
140 558.26513005845
150 523.581427488685
160 499.497493465129
170 481.253739190908
180 466.000352348414
190 452.661451062066
200 440.738920658828
210 429.913166150767
220 419.959171055061
230 410.709325593054
240 402.033135670584
250 393.825738207995
260 386.000929287956
270 378.484262798437
280 371.192855468751
290 363.919455641635
300 355.51612710402
310 339.623382719531
320 311.520281636492
330 292.571323132785
340 280.668131270663
350 271.136514026292
360 262.67595034561
370 254.846160692961
380 247.486717909157
390 240.539513083369
400 233.977967574148
410 227.783059777415
420 221.934868948328
430 216.408006913364
440 211.168023960387
450 206.168005156547
460 201.347513230922
470 196.644656956155
480 192.037422074923
490 187.587173845633
500 183.390697502268
510 179.475379191018
520 175.798343497704
530 172.308309116735
540 168.970543269283
550 165.763664285612
560 162.673412889929
570 159.689120305406
580 156.802078992793
590 154.004808274944
600 151.290706329917
610 148.653868944513
620 146.088982758896
630 143.591253853427
640 141.156354842033
650 138.780383371876
660 136.459829410371
670 134.191550833193
680 131.972757722565
690 129.801005806963
700 127.674198685544
710 125.590596863557
720 123.548829302205
730 121.547900587442
740 119.587184843859
750 117.666397389589
760 115.7855379704
770 113.944805434466
780 112.144491595828
790 110.384868978075
800 108.666090105118
810 106.98811345865
820 105.350664168149
830 103.753228951712
840 102.195078049746
850 100.675303703111
860 99.1928650540139
870 97.7466318337777
880 96.3354223201353
890 94.9580337214841
900 93.6132649556623
910 92.2999327576755
920 91.0168824158727
930 89.7629944338996
940 88.5371882449507
950 87.3384238832391
960 86.1657022960093
970 85.0180647967908
980 83.8945920081976
990 82.794402530634
};
\addlegendentry{heuristic Newton}
\addplot [very thick, color1, dashed, forget plot]
table {%
0 nan
10 nan
20 nan
30 nan
40 nan
50 nan
60 nan
70 nan
80 nan
90 nan
100 nan
110 nan
120 nan
130 nan
140 nan
150 nan
160 nan
170 nan
180 nan
190 nan
200 nan
210 nan
220 nan
230 nan
240 nan
250 nan
260 nan
270 nan
280 nan
290 nan
300 nan
310 nan
320 nan
330 nan
340 nan
350 nan
360 nan
370 nan
380 nan
390 nan
400 nan
410 nan
420 nan
430 nan
440 nan
450 nan
460 nan
470 nan
480 nan
490 nan
500 nan
510 nan
520 nan
530 nan
540 nan
550 nan
560 nan
570 nan
580 nan
590 nan
600 nan
610 nan
620 nan
630 nan
640 nan
650 nan
660 nan
670 nan
680 nan
690 nan
700 nan
710 nan
720 nan
730 nan
740 nan
750 nan
760 nan
770 nan
780 nan
790 nan
800 nan
810 nan
820 nan
830 nan
840 nan
850 nan
860 nan
870 nan
880 nan
890 nan
900 nan
910 nan
920 nan
930 nan
940 nan
950 nan
960 nan
970 nan
980 nan
990 nan
};
\addplot [very thick, color2]
table {%
0 15218.5703719251
10 213.536339989099
20 -23.0674795080865
30 -68.4242903855714
40 -89.3616761357831
50 -95.819294498671
60 -98.5020033227481
70 -99.1640374293841
80 -97.6413983442532
90 -96.7158555279042
100 -96.1485011088273
110 -95.6502345342305
120 -95.0741051302126
130 -95.187386917018
140 -95.151550113355
150 -95.2994086741398
160 -95.4213694735697
170 -95.5227494101198
180 -95.6232091261851
190 -95.7147354188489
200 -95.7982094923106
210 -95.8739957948492
220 -95.9393059479131
230 -95.9943105816231
240 -96.0413475133111
250 -96.0805245520288
260 -96.1130244006127
270 -96.1416557997054
280 -96.1638287521956
290 -96.184183702671
300 -96.2009986937958
310 -96.2164883521945
320 -96.2290508386258
330 -96.2400752968758
340 -96.251152850963
350 -96.2595639720992
360 -96.2678904709099
370 -96.2754528041078
380 -96.2818644718887
390 -96.2890548377568
400 -96.2937240436699
410 -96.2997654506386
420 -96.3056519535036
430 -96.3097552141763
440 -96.3139603551271
450 -96.3194238698583
460 -96.3237892122788
470 -96.327818395597
480 -96.3320107371849
490 -96.3350544460038
500 -96.3387355757962
510 -96.3429333459779
520 -96.3467841368393
530 -96.3502583251497
540 -96.3534584490049
550 -96.3571444709754
560 -96.3600499246398
570 -96.3635444143462
580 -96.3658304477181
590 -96.3693365850708
600 -96.3717187195732
610 -96.375256498287
620 -96.3776705263359
630 -96.3805508424598
640 -96.382968732458
650 -96.3864693226364
660 -96.3895392672266
670 -96.3918030154065
680 -96.3936936403562
690 -96.3965897166984
700 -96.3988099768231
710 -96.400648640672
720 -96.4029962919481
730 -96.4051960583527
740 -96.40813439794
750 -96.4096899480144
760 -96.4116469340197
770 -96.4133858959212
780 -96.4166531784922
790 -96.4175250304476
800 -96.419259162083
810 -96.4217238029919
820 -96.4237224523912
830 -96.4248455231241
840 -96.4258594714497
850 -96.4283786729883
860 -96.43053331391
870 -96.4321105258672
880 -96.4332568335571
890 -96.4344071237561
900 -96.4369133216828
910 -96.4377732682909
920 -96.4393261448825
930 -96.4426373551431
940 -96.4434349002607
950 -96.4432225195998
960 -96.4443471768115
970 -96.445944824042
980 -96.4480158495697
990 -96.4496416041136
};
\addlegendentry{Gauss-Newton}
\addplot [very thick, color3]
table {%
0 33479.1679644083
10 4105.63164050644
20 2662.7248056654
30 1974.97483684925
40 1676.5911075227
50 1493.00120201254
60 1369.61127529299
70 1287.19736422628
80 1243.67527570037
90 1224.51726059532
100 1214.54593598518
110 1208.58441567106
120 1205.05969789125
130 1201.6995170527
140 1199.8865116804
150 1200.78669166163
160 1202.76588611464
170 1199.8433006779
180 962.751777166319
190 826.347042738904
200 785.946883867405
210 755.114842511823
220 731.890513315694
230 711.438795259013
240 640.900188376988
250 578.369440304649
260 518.077299014936
270 487.600668270885
280 465.67112434837
290 436.300929962478
300 380.55344921023
310 329.271993476267
320 291.661199581905
330 273.549870674056
340 260.142914139803
350 249.537880270962
360 240.490941309587
370 232.840150193263
380 226.198753904364
390 220.959680218825
400 217.572144978354
410 216.009522113358
420 215.069689967644
430 213.616181198404
440 212.525780792196
450 211.879771110187
460 211.432818621868
470 210.323375779699
480 209.311884789677
490 208.4371601117
500 207.465178861002
510 206.336210677871
520 204.538259523907
530 202.054750995442
540 199.094712985322
550 196.351300650043
560 194.543728909316
570 193.664037039671
580 193.130099996678
590 192.539155093056
600 191.731350693169
610 190.812934330941
620 189.855232237252
630 189.265454563684
640 189.172023753618
650 189.191744614656
660 189.238411660556
670 189.464867385496
680 189.717562948498
690 189.626218889857
700 188.9876957288
710 188.182435626469
720 186.627830129982
730 184.478075727967
740 182.366169680927
750 180.841977331593
760 179.714602642055
770 178.862588989592
780 177.544570673865
790 175.797112828681
800 174.371841757569
810 173.582800013211
820 173.656518974865
830 174.082687973072
840 174.71909409517
850 175.536917630034
860 176.85368068787
870 179.221969171359
880 182.663985482948
890 186.717426648719
900 190.858368724426
910 194.497245199986
920 198.089460039018
930 201.480085345417
940 204.066224924909
950 206.643763209088
960 208.893409743946
970 210.429104266187
980 211.250791150642
990 211.489036613808
};
\addlegendentry{quasi-Newton}
\end{axis}

\end{tikzpicture}
	\end{subfigure}
	\hspace*{\fill}
	\begin{subfigure}[t]{.32\textwidth}
		% This file was created with tikzplotlib v0.9.14.
\begin{tikzpicture}

\definecolor{color0}{rgb}{0.12156862745098,0.466666666666667,0.705882352941177}
\definecolor{color1}{rgb}{1,0.498039215686275,0.0549019607843137}
\definecolor{color2}{rgb}{0.172549019607843,0.627450980392157,0.172549019607843}
\definecolor{color3}{rgb}{0.83921568627451,0.152941176470588,0.156862745098039}

\begin{axis}[
height=\figureheight,
tick pos=left,
title={Test NLPD (Newton)},
unbounded coords=jump,
width=\figurewidth,
x grid style={white!69.0196078431373!black},
xlabel={iteration number},
xmin=-49.5, xmax=1039.5,
xtick style={color=black},
y grid style={white!69.0196078431373!black},
ymin=0.9, ymax=3.3,
ytick style={color=black}
]
\addplot [very thick, color0, dashed]
table {%
0 4.09551913581722
10 3.55939494495643
20 3.42842083310624
30 3.27140976571051
40 3.06717873020924
50 2.80599489345042
60 2.57537228043252
70 2.40621608180308
80 2.28484348919045
90 2.18191731327814
100 2.0921124205628
110 2.03189348509784
120 1.99001517444187
130 1.95602116066105
140 1.90942105169632
150 2.02074543291784
160 2.08789140035329
170 2.09534503189837
180 2.07829334482064
190 2.05728867446449
200 2.03605946589713
210 2.01337022260675
220 1.9871958617785
230 1.95902554130258
240 1.92681610769728
250 1.89075708439387
260 1.8582220396438
270 1.81431546872204
280 1.77374906829248
290 1.73613344599029
300 1.74465556976729
310 1.69919854179903
320 1.69506726342017
330 1.67234172599396
340 1.63791624713048
350 1.63112571607728
360 1.60212428023822
370 1.61097513234547
380 1.62749523817289
390 1.63083216543269
400 1.64004796390564
410 1.66498101179532
420 1.63090352111696
430 1.6434316196067
440 1.67407002046268
450 1.71618083087572
460 1.78571077479831
470 1.83775452146274
480 1.86659323958602
490 1.89265935224157
500 1.91279628262598
510 1.93012947769209
520 1.94790658034487
530 1.96393636751744
540 1.97828979308386
550 1.99060996911872
560 1.9996690259162
570 2.00661277501177
580 2.01164555150611
590 2.01589647935053
600 2.01996689573518
610 2.02279317330416
620 2.02541727471202
630 2.02773464510355
640 2.02975578704245
650 2.0316062028027
660 2.03330931063333
670 2.0348923389821
680 2.0363792211512
690 2.03779143464528
700 2.03915145714777
710 2.04048576050438
720 2.04180234334799
730 2.0430839851291
740 2.04445688739872
750 2.04597276358168
760 2.04767815418296
770 2.04961457393917
780 2.05181307930496
790 2.05412959843504
800 2.05671483561061
810 2.05955851807153
820 2.06262791569765
830 2.06588155850351
840 2.0692748060062
850 2.07276416010743
860 2.07630985253375
870 2.07985363447894
880 2.08329568376813
890 2.08671319086229
900 2.09008526542096
910 2.09339411711349
920 2.09662466417186
930 2.0997547913288
940 2.10253295098598
950 2.10519698242322
960 2.10775238290171
970 2.1101977618903
980 2.11253387028936
990 2.11458291884126
};
\addplot [very thick, color1, dashed]
table {%
0 nan
10 nan
20 nan
30 nan
40 nan
50 nan
60 nan
70 nan
80 nan
90 nan
100 nan
110 nan
120 nan
130 nan
140 nan
150 nan
160 nan
170 nan
180 nan
190 nan
200 nan
210 nan
220 nan
230 nan
240 nan
250 nan
260 nan
270 nan
280 nan
290 nan
300 nan
310 nan
320 nan
330 nan
340 nan
350 nan
360 nan
370 nan
380 nan
390 nan
400 nan
410 nan
420 nan
430 nan
440 nan
450 nan
460 nan
470 nan
480 nan
490 nan
500 nan
510 nan
520 nan
530 nan
540 nan
550 nan
560 nan
570 nan
580 nan
590 nan
600 nan
610 nan
620 nan
630 nan
640 nan
650 nan
660 nan
670 nan
680 nan
690 nan
700 nan
710 nan
720 nan
730 nan
740 nan
750 nan
760 nan
770 nan
780 nan
790 nan
800 nan
810 nan
820 nan
830 nan
840 nan
850 nan
860 nan
870 nan
880 nan
890 nan
900 nan
910 nan
920 nan
930 nan
940 nan
950 nan
960 nan
970 nan
980 nan
990 nan
};
\addplot [very thick, color2]
table {%
0 3.67757041721638
10 2.51624061314512
20 1.15922042150292
30 1.15326039701872
40 1.10740591275768
50 1.20567929696274
60 1.22772591112053
70 1.28523419574381
80 1.22902940611008
90 1.24943872982005
100 1.28252161123983
110 1.22879064910886
120 1.18698164790911
130 1.10548240221427
140 1.0879143573952
150 1.08428784013327
160 1.06708699831937
170 1.06150351504339
180 1.05551374707019
190 1.05476463535163
200 1.05498485175582
210 1.05884988232818
220 1.05851782206425
230 1.05879261089639
240 1.05944136450908
250 1.06055898411365
260 1.06159828153955
270 1.06211034044031
280 1.06325199221873
290 1.0707682071965
300 1.07435943344663
310 1.09355334383188
320 1.10558225854663
330 1.1324708603051
340 1.14788595089418
350 1.12107454884961
360 1.11714315775796
370 1.11208008130413
380 1.09859196036854
390 1.06749941250488
400 1.06000663013946
410 1.07346437818118
420 1.06198111055755
430 1.05970122568172
440 1.06275196283918
450 1.04543990212075
460 1.04450775570904
470 1.03865602885736
480 1.04592532322994
490 1.05263997720159
500 1.04320529893272
510 1.04043505987888
520 1.03578641472702
530 1.03044764198517
540 1.02604471404339
550 1.0344820565126
560 1.0329599826788
570 1.03773997159814
580 1.04186891447776
590 1.07941210346214
600 1.08861647523194
610 1.08858581063564
620 1.08881132063729
630 1.09479099892634
640 1.09355415669835
650 1.09379936068198
660 1.09695546716929
670 1.08843818656948
680 1.0818795364883
690 1.0805051159142
700 1.07558547423783
710 1.07483967700275
720 1.0704989547292
730 1.06413955207276
740 1.06442768743927
750 1.06236052873234
760 1.06720765191339
770 1.07403476153703
780 1.07982461288377
790 1.08044321364887
800 1.0812769668872
810 1.08214769748223
820 1.08325629008121
830 1.08446139597301
840 1.08593804866218
850 1.08789442554095
860 1.09582308027338
870 1.102394885613
880 1.1077315639057
890 1.10888933129857
900 1.10404364735784
910 1.10532778963922
920 1.11205838393925
930 1.11337444225191
940 1.11802288903964
950 1.1191304177049
960 1.12037063429858
970 1.11737587400888
980 1.12064677859691
990 1.11446321397106
};
\addplot [very thick, color3]
table {%
0 2.74962412595566
10 2.87564285368357
20 2.92195064574959
30 2.98970199776321
40 2.98407737192792
50 2.91246293953809
60 2.8449138473051
70 2.79361610264327
80 2.74900769750263
90 2.69265974234798
100 2.61885475531792
110 2.54576659924877
120 2.51010850772288
130 2.495556679878
140 2.4856666523639
150 2.47828305323044
160 2.47222511905691
170 2.46678087303744
180 2.46149387594875
190 2.4589974720325
200 2.45373307354789
210 2.44816672218481
220 2.44701958860545
230 2.45031966672007
240 2.45559406094626
250 2.46438377385726
260 2.46223763784047
270 2.45389990365831
280 2.44083038498176
290 2.42681087197031
300 2.41221294509818
310 2.4022752938931
320 2.39816790830332
330 2.39401636147952
340 2.38017517928768
350 2.36266699666288
360 2.3469669778245
370 2.32605562908784
380 2.30505071129391
390 2.28999761819778
400 2.27965133872856
410 2.27056564532198
420 2.26517016016747
430 2.26410150245977
440 2.26328360632635
450 2.26387555428773
460 2.26347843747354
470 2.26349998799129
480 2.26716994549003
490 2.2762916575498
500 2.29039094295389
510 2.30371825527604
520 2.31663395812088
530 2.32551964511287
540 2.3283087211189
550 2.32711672431145
560 2.32593358753539
570 2.32668516061597
580 2.33376242817448
590 2.34922335518167
600 2.37028367071309
610 2.39433965589537
620 2.42007027307591
630 2.44885465286146
640 2.48038018751468
650 2.51588017009104
660 2.54524346314641
670 2.56377426268265
680 2.57711286418994
690 2.58496566709964
700 2.58775970912844
710 2.58392187778246
720 2.57794283584883
730 2.57217579799445
740 2.56572395670403
750 2.55917303244048
760 2.553346843204
770 2.55046808456497
780 2.5493636527489
790 2.55221748266682
800 2.55759276564366
810 2.56451796950341
820 2.56870453736129
830 2.57089425177742
840 2.5708822218175
850 2.56713057407611
860 2.55823382219185
870 2.53892029178363
880 2.51137976422046
890 2.47693452522007
900 2.43467005700363
910 2.40171129921008
920 2.38030011793315
930 2.3704999893825
940 2.36979189940596
950 2.37425721027985
960 2.37775934844623
970 2.37982571358981
980 2.36859342165372
990 2.34370686192605
};
\end{axis}

\end{tikzpicture} 
	\end{subfigure}
	\hspace*{\fill}
	\begin{subfigure}[t]{.32\textwidth}
		\raggedleft
		% This file was created with tikzplotlib v0.9.14.
\begin{tikzpicture}

\definecolor{color0}{rgb}{0.12156862745098,0.466666666666667,0.705882352941177}
\definecolor{color1}{rgb}{1,0.498039215686275,0.0549019607843137}
\definecolor{color2}{rgb}{0.172549019607843,0.627450980392157,0.172549019607843}
\definecolor{color3}{rgb}{0.83921568627451,0.152941176470588,0.156862745098039}

\begin{axis}[
height=\figureheight,
tick pos=left,
title={Ground Truth RMSE (Newton)},
unbounded coords=jump,
width=\figurewidth,
x grid style={white!69.0196078431373!black},
xlabel={iteration number},
xmin=-49.5, xmax=1039.5,
xtick style={color=black},
y grid style={white!69.0196078431373!black},
ymin=0.6, ymax=3,
ytick style={color=black}
]
\addplot [very thick, color0, dashed]
table {%
0 3.21889034006456
10 8.39508467679462
20 7.97803233210832
30 7.3964685653093
40 6.81425133389386
50 6.32877722093998
60 5.94602644331137
70 5.63697646484072
80 5.37978867258284
90 5.44360551972539
100 5.29577656815584
110 5.08236591479842
120 4.89995485607599
130 4.70812831832332
140 4.51105457235488
150 4.34743328876214
160 4.21767060703664
170 4.09473671446114
180 3.96262371183267
190 3.8352347864007
200 3.71561162661358
210 3.60289122347554
220 3.49564955765729
230 3.39266714324281
240 3.29305340235548
250 3.19622310555291
260 3.10184316361734
270 3.00977863772774
280 2.92004573638781
290 2.83277595670736
300 2.7481955827585
310 2.66661880675353
320 2.58837915045982
330 2.51381435653801
340 2.44419185922191
350 2.38192744294312
360 2.32934269227369
370 2.28718004760636
380 2.25453415468147
390 2.22951688447986
400 2.21004835396281
410 2.19433937931636
420 2.18102026117579
430 2.16910659788565
440 2.15792701133857
450 2.14705117011333
460 2.13622503367268
470 2.1253154754326
480 2.1142673491249
490 2.10307614597222
500 2.0917737920063
510 2.08041487355729
520 2.06905621358201
530 2.05774077423496
540 2.04649391265142
550 2.03532712350638
560 2.02424290499138
570 2.01323837035024
580 2.00230754986948
590 1.99144283992259
600 1.98063596034956
610 1.96987862682148
620 1.95916305017308
630 1.94848233262797
640 1.937830816968
650 1.92720444147507
660 1.91660115090663
670 1.90602140593142
680 1.89546881576461
690 1.88495088649699
700 1.87447982678143
710 1.86407328262754
720 1.85375479256719
730 1.84355368689452
740 1.83350414080453
750 1.82364317982993
760 1.81400765724359
770 1.80463055041317
780 1.79553724884383
790 1.78674266914514
800 1.77824990460657
810 1.77005070795947
820 1.76212758218607
830 1.75445684748589
840 1.74701191592942
850 1.73976613682306
860 1.73269485452354
870 1.72577660341733
880 1.71899356133465
890 1.71233147135981
900 1.70577924693659
910 1.69932843447448
920 1.69297265308881
930 1.6867070814355
940 1.6805280243454
950 1.67443256772069
960 1.66841831642134
970 1.6624832034351
980 1.6566253567003
990 1.65084301048781
};
\addplot [very thick, color1, dashed]
table {%
0 nan
10 nan
20 nan
30 nan
40 nan
50 nan
60 nan
70 nan
80 nan
90 nan
100 nan
110 nan
120 nan
130 nan
140 nan
150 nan
160 nan
170 nan
180 nan
190 nan
200 nan
210 nan
220 nan
230 nan
240 nan
250 nan
260 nan
270 nan
280 nan
290 nan
300 nan
310 nan
320 nan
330 nan
340 nan
350 nan
360 nan
370 nan
380 nan
390 nan
400 nan
410 nan
420 nan
430 nan
440 nan
450 nan
460 nan
470 nan
480 nan
490 nan
500 nan
510 nan
520 nan
530 nan
540 nan
550 nan
560 nan
570 nan
580 nan
590 nan
600 nan
610 nan
620 nan
630 nan
640 nan
650 nan
660 nan
670 nan
680 nan
690 nan
700 nan
710 nan
720 nan
730 nan
740 nan
750 nan
760 nan
770 nan
780 nan
790 nan
800 nan
810 nan
820 nan
830 nan
840 nan
850 nan
860 nan
870 nan
880 nan
890 nan
900 nan
910 nan
920 nan
930 nan
940 nan
950 nan
960 nan
970 nan
980 nan
990 nan
};
\addplot [very thick, color2]
table {%
0 2.29170768249979
10 3.30744021848066
20 0.998452599460215
30 0.896975527576538
40 0.912530198397575
50 1.0402104398769
60 1.08616490788479
70 1.03398014504709
80 1.01909002266764
90 1.02610620062015
100 1.04655713355604
110 1.08862900820857
120 1.10120420445381
130 1.106448471049
140 1.10975260021603
150 1.10767511021858
160 1.1056723853654
170 1.10437319137389
180 1.10344435009799
190 1.10280562114832
200 1.1023554961664
210 1.10202819811433
220 1.1017870106691
230 1.10159465631535
240 1.10143901281488
250 1.10130103966065
260 1.10118079706712
270 1.10106460325805
280 1.10095102293811
290 1.10083672785543
300 1.10072224215595
310 1.10061889191742
320 1.10050528156342
330 1.1003995495218
340 1.10028712476166
350 1.10018114289817
360 1.10007595702073
370 1.09996885955188
380 1.09986137777637
390 1.09976138646569
400 1.09966095760668
410 1.09955414277784
420 1.0994536688822
430 1.0993621661712
440 1.09925906270446
450 1.09916451170002
460 1.09907148561921
470 1.09897724266759
480 1.09888693636888
490 1.09878938729984
500 1.09870603166034
510 1.09861948791245
520 1.09853304137278
530 1.09844636682032
540 1.09836011903423
550 1.09826468961893
560 1.09819529856742
570 1.09811235930902
580 1.09803681906005
590 1.09795886852683
600 1.09788457587052
610 1.09780776307384
620 1.09773385468806
630 1.09766361091828
640 1.09758933944862
650 1.09751563925568
660 1.09745331941529
670 1.0973803243519
680 1.09731478027314
690 1.0972495261477
700 1.097189732831
710 1.09712122105836
720 1.09705974002616
730 1.09700228770637
740 1.09694303896035
750 1.09687784863771
760 1.09682827768687
770 1.09676782107958
780 1.09671402581947
790 1.09666197697205
800 1.09660418404127
810 1.09655654636802
820 1.09650142183107
830 1.09645151527353
840 1.09640435949681
850 1.09636052976317
860 1.0963113157367
870 1.09626455284039
880 1.09622041298666
890 1.09617526898719
900 1.09612849998064
910 1.09608312465794
920 1.09604514361852
930 1.0960004781465
940 1.09596540702005
950 1.09592775211388
960 1.09588675359662
970 1.09585141214133
980 1.09580974310889
990 1.09576951809104
};
\addplot [very thick, color3]
table {%
0 1.34289234345876
10 1.41970055324366
20 1.48822487637458
30 1.66206497511436
40 1.8360344049227
50 2.02949580539304
60 2.23834244557048
70 2.45831332349415
80 2.64197282212947
90 2.73625008439498
100 2.75646514350831
110 2.72678560065041
120 2.65821333798668
130 2.54641281104608
140 2.4085866039742
150 2.28913730389102
160 2.18402284391844
170 2.0795300358411
180 1.99209088176501
190 1.91030494237385
200 1.86783866383365
210 1.85995798054412
220 1.88900188251498
230 1.94043178663549
240 1.99459748331537
250 2.06331122620615
260 2.09390598627237
270 2.09219757580816
280 2.08562307267262
290 2.07866412452496
300 2.07510603368433
310 2.08889771263561
320 2.10733191016146
330 2.12443918262497
340 2.13695600246968
350 2.14217527323844
360 2.14297892319237
370 2.14337984971671
380 2.14657269529899
390 2.1535952733693
400 2.16182442308317
410 2.16671441379882
420 2.16739152523235
430 2.16707941628473
440 2.1680372040443
450 2.16976274319936
460 2.16988516667936
470 2.16745557116204
480 2.16344328177136
490 2.15901669913914
500 2.15500641021713
510 2.15174398690457
520 2.14866940304098
530 2.14470994533933
540 2.13957553295283
550 2.13397672806496
560 2.12889610834727
570 2.12528991501324
580 2.12390082086971
590 2.12474046951388
600 2.12637361748527
610 2.12630958207042
620 2.12235697968814
630 2.1138545175972
640 2.10279526742594
650 2.09247956086218
660 2.08423887011114
670 2.07708953413408
680 2.06979418484107
690 2.06080997430796
700 2.04955588070372
710 2.03683307829022
720 2.02345298189169
730 2.00791320371059
740 1.98678781563702
750 1.95956347585848
760 1.92875885893546
770 1.89843717854719
780 1.86972975036514
790 1.8416319793276
800 1.81409658688281
810 1.78822914935023
820 1.76538959818155
830 1.75462060775124
840 1.74586190384653
850 1.73716756063403
860 1.7273511893309
870 1.71533342148292
880 1.70116876022814
890 1.686726459619
900 1.67434843094768
910 1.66546567816887
920 1.65979511457404
930 1.65607005489847
940 1.65309732478931
950 1.65026035236875
960 1.64734134966069
970 1.64403201369475
980 1.63942058380843
990 1.63300140697096
};
\end{axis}

\end{tikzpicture} 
	\end{subfigure}\\
	\vspace{-1em}
	\begin{subfigure}[t]{.32\textwidth}
		\raggedright
		\input{./fig/gprn-vi-loss.tex}
	\end{subfigure}
	\hspace*{\fill}
	\begin{subfigure}[t]{.32\textwidth}
		\input{./fig/gprn-vi-nlpd.tex}
	\end{subfigure}
	\hspace*{\fill}
	\begin{subfigure}[t]{.32\textwidth}
		\raggedleft
		\input{./fig/gprn-vi-rmse.tex}
	\end{subfigure}\\
	\vspace{-1em}
	\begin{subfigure}[t]{.32\textwidth}
		\raggedright
		% This file was created with tikzplotlib v0.9.14.
\begin{tikzpicture}

\definecolor{color0}{rgb}{0.12156862745098,0.466666666666667,0.705882352941177}
\definecolor{color1}{rgb}{1,0.498039215686275,0.0549019607843137}
\definecolor{color2}{rgb}{0.172549019607843,0.627450980392157,0.172549019607843}
\definecolor{color3}{rgb}{0.83921568627451,0.152941176470588,0.156862745098039}

\begin{axis}[
height=\figureheight,
legend cell align={left},
legend style={fill opacity=0.8, draw opacity=1, text opacity=1, draw=white!80!black},
tick pos=left,
title={Training Loss (PEP)},
unbounded coords=jump,
width=\figurewidth,
x grid style={white!69.0196078431373!black},
xlabel={iteration number},
xmin=-49.5, xmax=1039.5,
xtick style={color=black},
y grid style={white!69.0196078431373!black},
ymin=-160, ymax=900,
ytick style={color=black}
]
\addplot [very thick, color0, dashed, forget plot]
table {%
0 nan
10 nan
20 nan
30 nan
40 nan
50 nan
60 nan
70 nan
80 nan
90 nan
100 nan
110 nan
120 nan
130 nan
140 nan
150 nan
160 nan
170 nan
180 nan
190 nan
200 nan
210 nan
220 nan
230 nan
240 nan
250 nan
260 nan
270 nan
280 nan
290 nan
300 nan
310 nan
320 nan
330 nan
340 nan
350 nan
360 nan
370 nan
380 nan
390 nan
400 nan
410 nan
420 nan
430 nan
440 nan
450 nan
460 nan
470 nan
480 nan
490 nan
500 nan
510 nan
520 nan
530 nan
540 nan
550 nan
560 nan
570 nan
580 nan
590 nan
600 nan
610 nan
620 nan
630 nan
640 nan
650 nan
660 nan
670 nan
680 nan
690 nan
700 nan
710 nan
720 nan
730 nan
740 nan
750 nan
760 nan
770 nan
780 nan
790 nan
800 nan
810 nan
820 nan
830 nan
840 nan
850 nan
860 nan
870 nan
880 nan
890 nan
900 nan
910 nan
920 nan
930 nan
940 nan
950 nan
960 nan
970 nan
980 nan
990 nan
};
\addplot [very thick, color1, dashed, forget plot]
table {%
0 nan
10 nan
20 nan
30 nan
40 nan
50 nan
60 nan
70 nan
80 nan
90 nan
100 nan
110 nan
120 nan
130 nan
140 nan
150 nan
160 nan
170 nan
180 nan
190 nan
200 nan
210 nan
220 nan
230 nan
240 nan
250 nan
260 nan
270 nan
280 nan
290 nan
300 nan
310 nan
320 nan
330 nan
340 nan
350 nan
360 nan
370 nan
380 nan
390 nan
400 nan
410 nan
420 nan
430 nan
440 nan
450 nan
460 nan
470 nan
480 nan
490 nan
500 nan
510 nan
520 nan
530 nan
540 nan
550 nan
560 nan
570 nan
580 nan
590 nan
600 nan
610 nan
620 nan
630 nan
640 nan
650 nan
660 nan
670 nan
680 nan
690 nan
700 nan
710 nan
720 nan
730 nan
740 nan
750 nan
760 nan
770 nan
780 nan
790 nan
800 nan
810 nan
820 nan
830 nan
840 nan
850 nan
860 nan
870 nan
880 nan
890 nan
900 nan
910 nan
920 nan
930 nan
940 nan
950 nan
960 nan
970 nan
980 nan
990 nan
};
\addplot [very thick, color2, forget plot]
table {%
0 nan
10 nan
20 nan
30 nan
40 nan
50 nan
60 nan
70 nan
80 nan
90 nan
100 nan
110 nan
120 nan
130 nan
140 nan
150 nan
160 nan
170 nan
180 nan
190 nan
200 nan
210 nan
220 nan
230 nan
240 nan
250 nan
260 nan
270 nan
280 nan
290 nan
300 nan
310 nan
320 nan
330 nan
340 nan
350 nan
360 nan
370 nan
380 nan
390 nan
400 nan
410 nan
420 nan
430 nan
440 nan
450 nan
460 nan
470 nan
480 nan
490 nan
500 nan
510 nan
520 nan
530 nan
540 nan
550 nan
560 nan
570 nan
580 nan
590 nan
600 nan
610 nan
620 nan
630 nan
640 nan
650 nan
660 nan
670 nan
680 nan
690 nan
700 nan
710 nan
720 nan
730 nan
740 nan
750 nan
760 nan
770 nan
780 nan
790 nan
800 nan
810 nan
820 nan
830 nan
840 nan
850 nan
860 nan
870 nan
880 nan
890 nan
900 nan
910 nan
920 nan
930 nan
940 nan
950 nan
960 nan
970 nan
980 nan
990 nan
};
\addplot [very thick, color3]
table {%
0 8165.52772321186
10 8317.83045935139
20 7763.52860478028
30 7015.93715525467
40 6181.89432936483
50 5560.98109885435
60 4883.79024803458
70 3939.87474606513
80 3391.8599629564
90 2886.39639812989
100 2524.37735665009
110 2240.62231719351
120 2045.24005187597
130 1822.43551948242
140 1617.22405958553
150 1379.05264716707
160 1332.89385346946
170 1277.66852334583
180 1142.64614274905
190 1050.95801060309
200 961.900177284202
210 938.240132762061
220 922.055501569081
230 879.494077410562
240 664.111618164159
250 591.330809297045
260 548.972574852646
270 521.439614888825
280 499.469208212918
290 475.373283614566
300 449.615394110528
310 432.469563876605
320 418.964946666246
330 407.205573033013
340 397.505269356904
350 372.333814255489
360 300.566705542081
370 224.888271785188
380 196.503467785497
390 181.05891828334
400 169.005308413633
410 158.48657561104
420 149.020025018712
430 140.579678249841
440 133.571640988739
450 127.257662836714
460 121.395548887469
470 115.620706103093
480 109.804719866632
490 104.339640319702
500 99.2154626998671
510 94.7845718305899
520 90.7351842978097
530 87.3411796702576
540 84.1508563167499
550 81.0813058489564
560 78.4758117713061
570 76.3092291763695
580 74.4705222236498
590 72.6626756654319
600 70.8536208638186
610 68.7893728906791
620 66.7601507936024
630 64.5307414534705
640 62.8843834137543
650 61.6427569803603
660 60.1990510483113
670 59.0898772489075
680 57.9213999261831
690 56.8824169964638
700 55.7899494887891
710 54.8838298194871
720 54.1613428794187
730 53.5492825344351
740 52.8951254589897
750 51.730666942288
760 50.5781372524933
770 49.4154694738811
780 48.565477398679
790 47.9051726481648
800 47.5817826048362
810 47.0849297558552
820 46.3693174839728
830 45.2956604489249
840 43.8194970396981
850 42.2981539585115
860 40.7154027685859
870 39.221067155632
880 38.2354956304434
890 37.471564236326
900 36.8606061957541
910 36.1570739788241
920 35.4533974122647
930 34.3852879980252
940 33.1616496639847
950 31.9036544217155
960 30.9944390720345
970 30.3836301166061
980 30.1050175844218
990 30.0456550781524
};
\addlegendentry{PEP quasi-Newton}
\end{axis}

\end{tikzpicture}
	\end{subfigure}
	\hspace*{\fill}
	\begin{subfigure}[t]{.32\textwidth}
		% This file was created with tikzplotlib v0.9.14.
\begin{tikzpicture}

\definecolor{color0}{rgb}{0.12156862745098,0.466666666666667,0.705882352941177}
\definecolor{color1}{rgb}{1,0.498039215686275,0.0549019607843137}
\definecolor{color2}{rgb}{0.172549019607843,0.627450980392157,0.172549019607843}
\definecolor{color3}{rgb}{0.83921568627451,0.152941176470588,0.156862745098039}

\begin{axis}[
height=\figureheight,
tick pos=left,
title={Test NLPD (PEP)},
unbounded coords=jump,
width=\figurewidth,
x grid style={white!69.0196078431373!black},
xlabel={iteration number},
xmin=-49.5, xmax=1039.5,
xtick style={color=black},
y grid style={white!69.0196078431373!black},
ymin=0.9, ymax=3.3,
ytick style={color=black}
]
\addplot [very thick, color0, dashed]
table {%
0 nan
10 nan
20 nan
30 nan
40 nan
50 nan
60 nan
70 nan
80 nan
90 nan
100 nan
110 nan
120 nan
130 nan
140 nan
150 nan
160 nan
170 nan
180 nan
190 nan
200 nan
210 nan
220 nan
230 nan
240 nan
250 nan
260 nan
270 nan
280 nan
290 nan
300 nan
310 nan
320 nan
330 nan
340 nan
350 nan
360 nan
370 nan
380 nan
390 nan
400 nan
410 nan
420 nan
430 nan
440 nan
450 nan
460 nan
470 nan
480 nan
490 nan
500 nan
510 nan
520 nan
530 nan
540 nan
550 nan
560 nan
570 nan
580 nan
590 nan
600 nan
610 nan
620 nan
630 nan
640 nan
650 nan
660 nan
670 nan
680 nan
690 nan
700 nan
710 nan
720 nan
730 nan
740 nan
750 nan
760 nan
770 nan
780 nan
790 nan
800 nan
810 nan
820 nan
830 nan
840 nan
850 nan
860 nan
870 nan
880 nan
890 nan
900 nan
910 nan
920 nan
930 nan
940 nan
950 nan
960 nan
970 nan
980 nan
990 nan
};
\addplot [very thick, color1, dashed]
table {%
0 nan
10 nan
20 nan
30 nan
40 nan
50 nan
60 nan
70 nan
80 nan
90 nan
100 nan
110 nan
120 nan
130 nan
140 nan
150 nan
160 nan
170 nan
180 nan
190 nan
200 nan
210 nan
220 nan
230 nan
240 nan
250 nan
260 nan
270 nan
280 nan
290 nan
300 nan
310 nan
320 nan
330 nan
340 nan
350 nan
360 nan
370 nan
380 nan
390 nan
400 nan
410 nan
420 nan
430 nan
440 nan
450 nan
460 nan
470 nan
480 nan
490 nan
500 nan
510 nan
520 nan
530 nan
540 nan
550 nan
560 nan
570 nan
580 nan
590 nan
600 nan
610 nan
620 nan
630 nan
640 nan
650 nan
660 nan
670 nan
680 nan
690 nan
700 nan
710 nan
720 nan
730 nan
740 nan
750 nan
760 nan
770 nan
780 nan
790 nan
800 nan
810 nan
820 nan
830 nan
840 nan
850 nan
860 nan
870 nan
880 nan
890 nan
900 nan
910 nan
920 nan
930 nan
940 nan
950 nan
960 nan
970 nan
980 nan
990 nan
};
\addplot [very thick, color2]
table {%
0 nan
10 nan
20 nan
30 nan
40 nan
50 nan
60 nan
70 nan
80 nan
90 nan
100 nan
110 nan
120 nan
130 nan
140 nan
150 nan
160 nan
170 nan
180 nan
190 nan
200 nan
210 nan
220 nan
230 nan
240 nan
250 nan
260 nan
270 nan
280 nan
290 nan
300 nan
310 nan
320 nan
330 nan
340 nan
350 nan
360 nan
370 nan
380 nan
390 nan
400 nan
410 nan
420 nan
430 nan
440 nan
450 nan
460 nan
470 nan
480 nan
490 nan
500 nan
510 nan
520 nan
530 nan
540 nan
550 nan
560 nan
570 nan
580 nan
590 nan
600 nan
610 nan
620 nan
630 nan
640 nan
650 nan
660 nan
670 nan
680 nan
690 nan
700 nan
710 nan
720 nan
730 nan
740 nan
750 nan
760 nan
770 nan
780 nan
790 nan
800 nan
810 nan
820 nan
830 nan
840 nan
850 nan
860 nan
870 nan
880 nan
890 nan
900 nan
910 nan
920 nan
930 nan
940 nan
950 nan
960 nan
970 nan
980 nan
990 nan
};
\addplot [very thick, color3]
table {%
0 2.71951148137902
10 2.76538811512015
20 2.81390804297689
30 2.8532807218466
40 2.85770280125079
50 2.88781082853641
60 2.9629279155427
70 2.96380810321597
80 2.94389437544657
90 2.90235285137645
100 2.86732418898299
110 2.83764919418576
120 2.79885877075691
130 2.71812117755312
140 2.5682412091942
150 2.47230835547251
160 2.4103539793543
170 2.38350758970109
180 2.32599537922487
190 2.23250365008466
200 2.11109080015555
210 2.02345619542941
220 1.96518171590312
230 1.98938408261836
240 1.98617148581559
250 1.95842033482083
260 1.97142316478219
270 1.98762820378415
280 2.0020780819507
290 2.00606954786349
300 2.00326494168098
310 1.99677133193669
320 1.98403258204641
330 1.96688896274379
340 1.96491916293799
350 1.97290663528816
360 2.04081619679797
370 2.10280325678251
380 2.11373658869606
390 2.11356922185913
400 2.10799480120984
410 2.09062314624861
420 2.06243863379679
430 2.03404935650386
440 2.00413378478351
450 1.96804883835316
460 1.9394839733037
470 1.93473632996716
480 1.94636222873245
490 1.97795713838986
500 2.02848481734416
510 2.08685931258564
520 2.12443028528307
530 2.15445123330251
540 2.20050297406553
550 2.23841704213758
560 2.22606385466728
570 2.2107941052951
580 2.20693827898486
590 2.21906781882422
600 2.23617919280356
610 2.24728334440464
620 2.23788620394614
630 2.19811048744455
640 2.1278391464678
650 2.03748120125062
660 1.94099074308022
670 1.84151788420803
680 1.73111241517437
690 1.63028783805164
700 1.5486400450437
710 1.49658349591948
720 1.46828960670842
730 1.45695036522937
740 1.45612970050751
750 1.45797974005623
760 1.4610182169611
770 1.46231205867506
780 1.4697642359562
790 1.48812004738498
800 1.51321376235974
810 1.544110088071
820 1.57324199132978
830 1.59828470556231
840 1.61731193419285
850 1.62541914682182
860 1.63705048423656
870 1.64837324566515
880 1.65687295608023
890 1.66388355161966
900 1.671282468239
910 1.67842226801114
920 1.68367240337566
930 1.68141980592747
940 1.67334375302578
950 1.65971774557971
960 1.64326840304755
970 1.62028085622914
980 1.5952187263521
990 1.5811012366944
};
\end{axis}

\end{tikzpicture}
	\end{subfigure}
	\hspace*{\fill}
	\begin{subfigure}[t]{.32\textwidth}
		\raggedleft
		% This file was created with tikzplotlib v0.9.14.
\begin{tikzpicture}

\definecolor{color0}{rgb}{0.12156862745098,0.466666666666667,0.705882352941177}
\definecolor{color1}{rgb}{1,0.498039215686275,0.0549019607843137}
\definecolor{color2}{rgb}{0.172549019607843,0.627450980392157,0.172549019607843}
\definecolor{color3}{rgb}{0.83921568627451,0.152941176470588,0.156862745098039}

\begin{axis}[
height=\figureheight,
tick pos=left,
title={Ground Truth RMSE (PEP)},
unbounded coords=jump,
width=\figurewidth,
x grid style={white!69.0196078431373!black},
xlabel={iteration number},
xmin=-49.5, xmax=1039.5,
xtick style={color=black},
y grid style={white!69.0196078431373!black},
ymin=0.6, ymax=3,
ytick style={color=black}
]
\addplot [very thick, color0, dashed]
table {%
0 nan
10 nan
20 nan
30 nan
40 nan
50 nan
60 nan
70 nan
80 nan
90 nan
100 nan
110 nan
120 nan
130 nan
140 nan
150 nan
160 nan
170 nan
180 nan
190 nan
200 nan
210 nan
220 nan
230 nan
240 nan
250 nan
260 nan
270 nan
280 nan
290 nan
300 nan
310 nan
320 nan
330 nan
340 nan
350 nan
360 nan
370 nan
380 nan
390 nan
400 nan
410 nan
420 nan
430 nan
440 nan
450 nan
460 nan
470 nan
480 nan
490 nan
500 nan
510 nan
520 nan
530 nan
540 nan
550 nan
560 nan
570 nan
580 nan
590 nan
600 nan
610 nan
620 nan
630 nan
640 nan
650 nan
660 nan
670 nan
680 nan
690 nan
700 nan
710 nan
720 nan
730 nan
740 nan
750 nan
760 nan
770 nan
780 nan
790 nan
800 nan
810 nan
820 nan
830 nan
840 nan
850 nan
860 nan
870 nan
880 nan
890 nan
900 nan
910 nan
920 nan
930 nan
940 nan
950 nan
960 nan
970 nan
980 nan
990 nan
};
\addplot [very thick, color1, dashed]
table {%
0 nan
10 nan
20 nan
30 nan
40 nan
50 nan
60 nan
70 nan
80 nan
90 nan
100 nan
110 nan
120 nan
130 nan
140 nan
150 nan
160 nan
170 nan
180 nan
190 nan
200 nan
210 nan
220 nan
230 nan
240 nan
250 nan
260 nan
270 nan
280 nan
290 nan
300 nan
310 nan
320 nan
330 nan
340 nan
350 nan
360 nan
370 nan
380 nan
390 nan
400 nan
410 nan
420 nan
430 nan
440 nan
450 nan
460 nan
470 nan
480 nan
490 nan
500 nan
510 nan
520 nan
530 nan
540 nan
550 nan
560 nan
570 nan
580 nan
590 nan
600 nan
610 nan
620 nan
630 nan
640 nan
650 nan
660 nan
670 nan
680 nan
690 nan
700 nan
710 nan
720 nan
730 nan
740 nan
750 nan
760 nan
770 nan
780 nan
790 nan
800 nan
810 nan
820 nan
830 nan
840 nan
850 nan
860 nan
870 nan
880 nan
890 nan
900 nan
910 nan
920 nan
930 nan
940 nan
950 nan
960 nan
970 nan
980 nan
990 nan
};
\addplot [very thick, color2]
table {%
0 nan
10 nan
20 nan
30 nan
40 nan
50 nan
60 nan
70 nan
80 nan
90 nan
100 nan
110 nan
120 nan
130 nan
140 nan
150 nan
160 nan
170 nan
180 nan
190 nan
200 nan
210 nan
220 nan
230 nan
240 nan
250 nan
260 nan
270 nan
280 nan
290 nan
300 nan
310 nan
320 nan
330 nan
340 nan
350 nan
360 nan
370 nan
380 nan
390 nan
400 nan
410 nan
420 nan
430 nan
440 nan
450 nan
460 nan
470 nan
480 nan
490 nan
500 nan
510 nan
520 nan
530 nan
540 nan
550 nan
560 nan
570 nan
580 nan
590 nan
600 nan
610 nan
620 nan
630 nan
640 nan
650 nan
660 nan
670 nan
680 nan
690 nan
700 nan
710 nan
720 nan
730 nan
740 nan
750 nan
760 nan
770 nan
780 nan
790 nan
800 nan
810 nan
820 nan
830 nan
840 nan
850 nan
860 nan
870 nan
880 nan
890 nan
900 nan
910 nan
920 nan
930 nan
940 nan
950 nan
960 nan
970 nan
980 nan
990 nan
};
\addplot [very thick, color3]
table {%
0 1.33755298562715
10 1.34357300056338
20 1.35672917872558
30 1.37000479591446
40 1.3738622903058
50 1.3802976810667
60 1.39933309690663
70 1.40336493133955
80 1.40179363504703
90 1.40707625089832
100 1.41118002466005
110 1.41402873916163
120 1.39800581770749
130 1.36289842960277
140 1.30251465225381
150 1.24298549654943
160 1.20626692092609
170 1.16956831545193
180 1.12848699278578
190 1.07014326005863
200 1.00630233861569
210 0.951016833689065
220 0.905802390971238
230 0.892015412466139
240 0.884865666107309
250 0.862663910230795
260 0.870052203646435
270 0.875005790157117
280 0.873273453872242
290 0.868413507330212
300 0.864732441219559
310 0.86153742205727
320 0.855735456339578
330 0.849269795583996
340 0.852218619882907
350 0.889091411377365
360 0.942671610943318
370 0.909520178637563
380 0.883551806697752
390 0.867946742421122
400 0.857364680953049
410 0.850318253349083
420 0.844074173748589
430 0.837392018759571
440 0.831607449673845
450 0.82851347165853
460 0.827828682335842
470 0.827146874877681
480 0.825202955789403
490 0.824208783431501
500 0.828837925990745
510 0.842731208118471
520 0.864597793713005
530 0.890700439536561
540 0.917106830515686
550 0.945579468390847
560 0.978559413953877
570 1.01620581587433
580 1.05580202909029
590 1.09500190578307
600 1.12452597641747
610 1.1347647139074
620 1.13715691980121
630 1.12937037274345
640 1.11058905108998
650 1.08434527641958
660 1.05330011104288
670 1.01996210730094
680 0.986452798320162
690 0.955909900139024
700 0.927996196601964
710 0.903445646990328
720 0.881994703765551
730 0.86373309538482
740 0.84301084930936
750 0.821246773871318
760 0.801372481019286
770 0.782460122186413
780 0.764903689277483
790 0.748852185281722
800 0.735692756380176
810 0.727882633966809
820 0.725909613892344
830 0.727860057395688
840 0.73045038792316
850 0.730091796243487
860 0.729266060978655
870 0.722319553109643
880 0.711840438935679
890 0.70046935621256
900 0.691058022347181
910 0.684144753346756
920 0.679377000286462
930 0.676315366422935
940 0.67490059304322
950 0.675470639082674
960 0.677735771530641
970 0.680405271933907
980 0.683837159549071
990 0.689079114351721
};
\end{axis}

\end{tikzpicture}
	\end{subfigure}\\
	\vspace{-1em}
	\begin{subfigure}[t]{.32\textwidth}
		\raggedright
		\input{./fig/gprn-pl2-loss.tex}
	\end{subfigure}
	\hspace*{\fill}
	\begin{subfigure}[t]{.32\textwidth}
		\input{./fig/gprn-pl2-nlpd.tex}
	\end{subfigure}
	\hspace*{\fill}
	\begin{subfigure}[t]{.32\textwidth}
		\raggedleft
		\input{./fig/gprn-pl2-rmse.tex}
	\end{subfigure}\\
	\vspace{-0.2cm}
	\caption{Gaussian process regression network (GPRN) results. Mean value across 4 synthetic data sets shown. Some methods not plotted due to divergent behaviour during training. The Gauss--Newton methods (including PL) significantly outperform all other approaches in terms of convergence rate, test NLPD and their ability to recover the ground truth.} \label{fig:gprn-results}
\end{figure*}

\cref{fig:gprn-plot} plots an example prediction result on one of the synthetic data sets, and \cref{fig:gprn-results} plots the mean prediction performance across the four data sets. These results show that the independence assumption between latent processes made in the original GPRN paper is highly detrimental to performance and convergence rate. By including the cross-covariance terms, the model improves in terms of prediction quality. However, the original approach proposed by \citet{wilson2012gaussian} is much more efficient than our methods when the number of latent functions is large. 

The Gauss--Newton methods significantly outperform all others with regards to convergence rate, test NLPD and ground truth RMSE. PL and PL2 Gauss--Newton both perform well, but exhibit oscillatory behaviour suggesting they may require a smaller learning rate than the Gauss--Newton and variational Gauss--Newton approaches. The quasi-Newton methods perform poorly on this task, since the rank-two BFGS updates are less accurate than on the previous tasks due to the higher number of latent functions.

\section{Conclusions and Discussion}

Through careful analysis of variational inference, expectation propagation and posterior linearisation, we have shown that many approaches to approximate Bayesian inference can be viewed as update rules to local approximate likelihood terms. Furthermore these update rules involve computing the Jacobian and (approximate) Hessian of a surrogate target, and can be cast under the framework of numerical optimisation.

Our work aims to draw connections between such ideas from optimisation, machine learning and signal processing, and in particular the development of the variational Gauss--Newton method provides a key link between VI, the Gauss--Newton method, and linearisation-based methods (which also turn out to make Gauss--Newton approximations). Additionally, it serves as an approximation to VI that guarantees PSD updates and which exhibits excellent performance in some very difficult inference tasks such as the GPRN multi-output model.

The quasi-Newton methods showed less promise experimentally, especially in terms of convergence rates, but for some tasks they are able to capture shared information between latent components that other methods are not. This may motivate further work in this area, since it is clear that the heuristic and Gauss--Newton methods are not optimal in terms of inference quality. Our PEP quasi-Newton approach also contributes to the search for accurate and stable EP algorithms.

The explicit connections to the optimisation literature presented here motivate avenues for further research. Perhaps the most interesting of these would be analysis of the convergence properties of the Bayes--Newton methods, and development of line-search methods for improving the stability and convergence of all schemes. Whilst convergence analysis was out of scope for this paper, we hope that the connections to Newton's method and its variants is a useful perspective in this regard. Python code for the methods and experiments is provided at \url{https://github.com/AaltoML/BayesNewton} (see \cref{sec:reproducability} for instructions on how to reproduce the results).

\vskip 0.2in
\bibliography{bibliography}

\begin{thebibliography}{64}
\providecommand{\natexlab}[1]{#1}
\providecommand{\url}[1]{\texttt{#1}}
\expandafter\ifx\csname urlstyle\endcsname\relax
  \providecommand{\doi}[1]{doi: #1}\else
  \providecommand{\doi}{doi: \begingroup \urlstyle{rm}\Url}\fi

\bibitem[Adam et~al.(2021)Adam, Chang, Khan, and Solin]{adam2021dual}
Vincent Adam, Paul~E. Chang, Mohammad~Emtiyaz Khan, and Arno Solin.
\newblock Dual parameterization of sparse variational {G}aussian processes.
\newblock In \emph{Advances in Neural Information Processing Systems 34
  (NeurIPS)}, pages 11474--11486. Curran Associates, Inc., 2021.

\bibitem[Amari(1998)]{amari1998natural}
Shun-Ichi Amari.
\newblock Natural gradient works efficiently in learning.
\newblock \emph{Neural Computation}, 10\penalty0 (2):\penalty0 251--276, 1998.

\bibitem[Bell(1994)]{bell1994iterated}
Bradley~M Bell.
\newblock The iterated {K}alman smoother as a {G}auss--{N}ewton method.
\newblock \emph{SIAM Journal on Optimization}, 4\penalty0 (3):\penalty0
  626--636, 1994.

\bibitem[Bj{\"o}rck(1996)]{bjorck1996numerical}
{\AA}ke Bj{\"o}rck.
\newblock \emph{Numerical Methods for Least Squares Problems}.
\newblock SIAM, 1996.

\bibitem[Blei et~al.(2017)Blei, Kucukelbir, and McAuliffe]{blei2017variational}
David~M. Blei, Alp Kucukelbir, and Jon~D. McAuliffe.
\newblock Variational inference: A review for statisticians.
\newblock \emph{Journal of the American Statistical Association}, 112\penalty0
  (518):\penalty0 859--877, 2017.

\bibitem[Broyden(1967)]{broyden1967quasi}
Charles~G Broyden.
\newblock Quasi-{N}ewton methods and their application to function
  minimisation.
\newblock \emph{Mathematics of Computation}, 21\penalty0 (99):\penalty0
  368--381, 1967.

\bibitem[Broyden(1969)]{broyden1969new}
Charles~G Broyden.
\newblock A new double-rank minimisation algorithm.
\newblock \emph{Notices of the American Mathematical Society}, 16\penalty0
  (4):\penalty0 670, 1969.

\bibitem[Bui et~al.(2017)Bui, Yan, and Turner]{bui2017unifying}
Thang~D Bui, Josiah Yan, and Richard~E Turner.
\newblock A unifying framework for {G}aussian process pseudo-point
  approximations using power expectation propagation.
\newblock \emph{Journal of Machine Learning Research (JMLR)}, 18\penalty0
  (1):\penalty0 3649--3720, 2017.

\bibitem[Bui et~al.(2018)Bui, Nguyen, Swaroop, and Turner]{bui2018partitioned}
Thang~D Bui, Cuong~V Nguyen, Siddharth Swaroop, and Richard~E Turner.
\newblock Partitioned variational inference: A unified framework encompassing
  federated and continual learning.
\newblock \emph{arXiv preprint arXiv:1811.11206}, 2018.

\bibitem[Challis and Barber(2013)]{challis2013gaussian}
Edward Challis and David Barber.
\newblock Gaussian {K}ullback-{L}eibler approximate inference.
\newblock \emph{Journal of Machine Learning Research}, 14\penalty0
  (8):\penalty0 2239--2286, 2013.

\bibitem[Chang et~al.(2020)Chang, Wilkinson, Khan, and Solin]{chang2020fast}
Paul~E. Chang, William~J. Wilkinson, Mohammed~Emtiyaz Khan, and Arno Solin.
\newblock Fast variational learning in state-space {G}aussian process models.
\newblock In \emph{International Workshop on Machine Learning for Signal
  Processing (MLSP)}. IEEE, 2020.

\bibitem[Csat{\'o} and Opper(2002)]{csato2002sparse}
Lehel Csat{\'o} and Manfred Opper.
\newblock Sparse on-line {G}aussian processes.
\newblock \emph{Neural Computation}, 14\penalty0 (3):\penalty0 641--668, 2002.

\bibitem[Dehaene and Barthelm{\'e}(2018)]{dehaene2018expectation}
Guillaume Dehaene and Simon Barthelm{\'e}.
\newblock Expectation propagation in the large data limit.
\newblock \emph{Journal of the Royal Statistical Society: Series B (Statistical
  Methodology)}, 80\penalty0 (1):\penalty0 199--217, 2018.

\bibitem[Fletcher(1970)]{fletcher1970new}
Roger Fletcher.
\newblock A new approach to variable metric algorithms.
\newblock \emph{The Computer Journal}, 13\penalty0 (3):\penalty0 317--322,
  1970.

\bibitem[Garc{\'\i}a-Fern{\'a}ndez et~al.(2016)Garc{\'\i}a-Fern{\'a}ndez,
  Svensson, and S{\"a}rkk{\"a}]{garcia2016iterated}
{\'A}ngel~F Garc{\'\i}a-Fern{\'a}ndez, Lennart Svensson, and Simo
  S{\"a}rkk{\"a}.
\newblock Iterated posterior linearization smoother.
\newblock \emph{IEEE Transactions on Automatic Control}, 62\penalty0
  (4):\penalty0 2056--2063, 2016.

\bibitem[Garc{\'\i}a-Fern{\'a}ndez et~al.(2019)Garc{\'\i}a-Fern{\'a}ndez,
  Tronarp, and S{\"a}rkk{\"a}]{garcia2019gaussian}
{\'A}ngel~F Garc{\'\i}a-Fern{\'a}ndez, Filip Tronarp, and Simo S{\"a}rkk{\"a}.
\newblock Gaussian process classification using posterior linearization.
\newblock \emph{IEEE Signal Processing Letters}, 26\penalty0 (5):\penalty0
  735--739, 2019.

\bibitem[Gelb(1974)]{gelb1974applied}
Arthur Gelb.
\newblock \emph{Applied Optimal Estimation}.
\newblock MIT Press, 1974.

\bibitem[Goldberg et~al.(1997)Goldberg, Williams, and
  Bishop]{goldberg1997regression}
Paul~W Goldberg, Christopher~KI Williams, and Christopher~M Bishop.
\newblock Regression with input-dependent noise: A {G}aussian process
  treatment.
\newblock \emph{Advances in Neural Information Processing Systems 10 (NIPS)},
  pages 493--499, 1997.

\bibitem[Goldfarb(1970)]{goldfarb1970family}
Donald Goldfarb.
\newblock A family of variable-metric methods derived by variational means.
\newblock \emph{Mathematics of Computation}, 24\penalty0 (109):\penalty0
  23--26, 1970.

\bibitem[Golub and Pereyra(1973)]{golub1973differentiation}
Gene~H Golub and Victor Pereyra.
\newblock The differentiation of pseudo-inverses and nonlinear least squares
  problems whose variables separate.
\newblock \emph{SIAM Journal on Numerical Analysis}, 10\penalty0 (2):\penalty0
  413--432, 1973.

\bibitem[Hamelijnck et~al.(2021)Hamelijnck, Wilkinson, Loppi, Solin, and
  Damoulas]{hamelijnck2021spatio}
Oliver Hamelijnck, William~J. Wilkinson, Niki Loppi, Arno Solin, and Theodoros
  Damoulas.
\newblock Spatio-temporal variational {G}aussian processes.
\newblock In \emph{Advances in Neural Information Processing Systems 34
  (NeurIPS)}, pages 23621--23633. Curran Associates, Inc., 2021.

\bibitem[Hartikainen(2013)]{Hartikainen:2013}
Jouni Hartikainen.
\newblock \emph{Sequential Inference for Latent Temporal {G}aussian Process
  Models}.
\newblock Doctoral dissertation, Aalto University, Finland, 2013.

\bibitem[Hennig and Kiefel(2013)]{hennig2013quasi}
Philipp Hennig and Martin Kiefel.
\newblock Quasi-{N}ewton methods: {A} new direction.
\newblock \emph{The Journal of Machine Learning Research}, 14\penalty0
  (1):\penalty0 843--865, 2013.

\bibitem[Hennig et~al.(2015)Hennig, Osborne, and
  Girolami]{hennig2015probabilistic}
Philipp Hennig, Michael~A Osborne, and Mark Girolami.
\newblock Probabilistic numerics and uncertainty in computations.
\newblock \emph{Proceedings of the Royal Society A: Mathematical, Physical and
  Engineering Sciences}, 471\penalty0 (2179):\penalty0 20150142, 2015.

\bibitem[Hensman et~al.(2015)Hensman, Matthews, and
  Ghahramani]{hensman2015scalable}
James Hensman, Alexander Matthews, and Zoubin Ghahramani.
\newblock Scalable variational {G}aussian process classification.
\newblock In \emph{Proceedings of the Eighteenth International Conference on
  Artificial Intelligence and Statistics (AISTATS)}, volume~38 of
  \emph{Proceedings of Machine Learning Research}, pages 351--360. PMLR, 2015.

\bibitem[Jyl{\"a}nki et~al.(2011)Jyl{\"a}nki, Vanhatalo, and
  Vehtari]{jylanki2011robust}
Pasi Jyl{\"a}nki, Jarno Vanhatalo, and Aki Vehtari.
\newblock Robust {G}aussian process regression with a {S}tudent-t likelihood.
\newblock \emph{Journal of Machine Learning Research}, 12\penalty0
  (99):\penalty0 3227--3257, 2011.

\bibitem[Khan and Lin(2017)]{khan2017conjugate}
Mohammad~Emtiyaz Khan and Wu~Lin.
\newblock Conjugate-computation variational inference: {C}onverting variational
  inference in non-conjugate models to inferences in conjugate models.
\newblock In \emph{Proceedings of the 20th International Conference on
  Artificial Intelligence and Statistics (AISTATS)}, volume~54 of
  \emph{Proceedings of Machine Learning Research}, pages 878--887. PMLR, 2017.

\bibitem[Khan and Rue(2021)]{khan2021bayesian}
Mohammad~Emtiyaz Khan and H{\aa}vard Rue.
\newblock The {B}ayesian {L}earning {R}ule.
\newblock \emph{arXiv preprint arXiv:2107.04562}, 2021.

\bibitem[Khan et~al.(2018)Khan, Nielsen, Tangkaratt, Lin, Gal, and
  Srivastava]{khan2018fast}
Mohammad~Emtiyaz Khan, Didrik Nielsen, Voot Tangkaratt, Wu~Lin, Yarin Gal, and
  Akash Srivastava.
\newblock Fast and scalable {B}ayesian deep learning by weight-perturbation in
  {A}dam.
\newblock In \emph{Proceedings of the 35th International Conference on Machine
  Learning (ICML)}, volume~80 of \emph{Proceedings of Machine Learning
  Research}, pages 2611--2620. PMLR, 2018.

\bibitem[Khan et~al.(2019)Khan, Immer, Abedi, and Korzepa]{khan2019approximate}
Mohammad~Emtiyaz Khan, Alexander Immer, Ehsan Abedi, and Maciej Korzepa.
\newblock Approximate inference turns deep networks into {G}aussian processes.
\newblock In \emph{Advances in Neural Information Processing Systems 32
  (NeurIPS)}, pages 3094--3104. Curran Associates, Inc., 2019.

\bibitem[Kingma and Ba(2014)]{kingma2014adam}
Diederik~P Kingma and Jimmy Ba.
\newblock Adam: A method for stochastic optimization.
\newblock \emph{arXiv preprint arXiv:1412.6980}, 2014.

\bibitem[L{\'a}zaro-Gredilla and Titsias(2011)]{lazaro2011variational}
Miguel L{\'a}zaro-Gredilla and Michalis~K Titsias.
\newblock Variational heteroscedastic {G}aussian process regression.
\newblock In \emph{Proceedings of the 28th International Conference on Machine
  Learning (ICML)}. Omnipress, 2011.

\bibitem[Leithead and Zhang(2007)]{leithead2007n}
William~E Leithead and Yunong Zhang.
\newblock {O(${N^{2}}$)}-operation approximation of covariance matrix inverse
  in {G}aussian process regression based on quasi-{N}ewton {BFGS} method.
\newblock \emph{Communications in Statistics—Simulation and Computation},
  36\penalty0 (2):\penalty0 367--380, 2007.

\bibitem[Li et~al.(2015)Li, Hern{\'a}ndez-Lobato, and Turner]{li2015stochastic}
Yingzhen Li, Jos{\'e}~Miguel Hern{\'a}ndez-Lobato, and Richard~E Turner.
\newblock Stochastic expectation propagation.
\newblock In \emph{Advances in Neural Information Processing Systems 28
  (NIPS)}, pages 2323--2331. Curran Associates, Inc., 2015.

\bibitem[Lin et~al.(2019)Lin, Khan, and Schmidt]{lin2019stein}
Wu~Lin, Mohammad~Emtiyaz Khan, and Mark Schmidt.
\newblock Stein's lemma for the reparameterization trick with exponential
  family mixtures.
\newblock In \emph{ICML Workshop on Stein's Method in Machine Learning and
  Statistics}, 2019.

\bibitem[Lin et~al.(2020)Lin, Schmidt, and Khan]{lin2020handling}
Wu~Lin, Mark Schmidt, and Mohammad~Emtiyaz Khan.
\newblock Handling the positive-definite constraint in the {B}ayesian
  {L}earning {R}ule.
\newblock In \emph{Proceedings of the 37th International Conference on Machine
  Learning (ICML)}, volume 119 of \emph{Proceedings of Machine Learning
  Research}. PMLR, 2020.

\bibitem[McNamee and Stenger(1967)]{McNamee1967}
John McNamee and Frank Stenger.
\newblock Construction of fully symmetric numerical integration formulas.
\newblock \emph{Numerische Mathematik}, 10\penalty0 (4):\penalty0 327--344,
  1967.

\bibitem[Minka(2004)]{minka2004power}
Thomas~P. Minka.
\newblock Power {EP}.
\newblock Technical report, Microsoft Research, 2004.
\newblock MSR-TR-2005-173.

\bibitem[Minka(2001)]{minka2001family}
Thomas~Peter Minka.
\newblock \emph{A family of algorithms for approximate {B}ayesian inference}.
\newblock PhD thesis, Massachusetts Institute of Technology, 2001.

\bibitem[Minka(2005)]{minka2005divergence}
Tom Minka.
\newblock Divergence measures and message passing.
\newblock Technical report, Microsoft Research, 2005.

\bibitem[Nickisch and Rasmussen(2008)]{nickisch2008approximations}
Hannes Nickisch and Carl~Edward Rasmussen.
\newblock Approximations for binary {G}aussian process classification.
\newblock \emph{Journal of Machine Learning Research}, 9\penalty0
  (Oct):\penalty0 2035--2078, 2008.

\bibitem[Nickisch et~al.(2018)Nickisch, Solin, and
  Grigorievskiy]{nickisch2018state}
Hannes Nickisch, Arno Solin, and Alexander Grigorievskiy.
\newblock State space {G}aussian processes with non-{G}aussian likelihood.
\newblock In \emph{Proceedings of the 35th International Conference on Machine
  Learning (ICML)}, volume~80 of \emph{Proceedings of Machine Learning
  Research}, pages 3789--3798. PMLR, 2018.

\bibitem[Nocedal and Wright(2006)]{nocedal2006numerical}
Jorge Nocedal and Stephen~J. Wright.
\newblock \emph{Numerical Optimization}.
\newblock Springer Series in Operations Research and Financial Engineering,
  2006.

\bibitem[Opper and Archambeau(2009)]{opper2009variational}
Manfred Opper and C{\'e}dric Archambeau.
\newblock The variational {G}aussian approximation revisited.
\newblock \emph{Neural Computation}, 21\penalty0 (3):\penalty0 786--792, 2009.

\bibitem[Opper and Winther(2005)]{opper2005expectation}
Manfred Opper and Ole Winther.
\newblock Expectation consistent approximate inference.
\newblock \emph{Journal of Machine Learning Research}, 6\penalty0
  (Dec):\penalty0 2177--2204, 2005.

\bibitem[Rasmussen and Williams(2006)]{rasmussen2003gaussian}
Carl~Edward Rasmussen and Christopher~KI Williams.
\newblock \emph{Gaussian Processes for Machine Learning}.
\newblock MIT Press, Cambridge, MA, USA, 2006.

\bibitem[Salimbeni et~al.(2018)Salimbeni, Eleftheriadis, and
  Hensman]{salimbeni2018natural}
Hugh Salimbeni, Stefanos Eleftheriadis, and James Hensman.
\newblock Natural gradients in practice: {N}on-conjugate variational inference
  in {G}aussian process models.
\newblock In \emph{Proceedings of the Twenty-First International Conference on
  Artificial Intelligence and Statistics (AISTATS)}, volume~84 of
  \emph{Proceedings of Machine Learning Research}, pages 689--697. PMLR, 2018.

\bibitem[S{\"a}rkk{\"a}(2013)]{sarkka2013bayesian}
Simo S{\"a}rkk{\"a}.
\newblock \emph{Bayesian Filtering and Smoothing}.
\newblock Cambridge University Press, 2013.

\bibitem[S{\"a}rkk{\"a} and Solin(2019)]{sarkka2019applied}
Simo S{\"a}rkk{\"a} and Arno Solin.
\newblock \emph{Applied Stochastic Differential Equations}.
\newblock Cambridge University Press, 2019.

\bibitem[S\"arkk\"a et~al.(2013)S\"arkk\"a, Solin, and
  Hartikainen]{Sarkka+Solin+Hartikainen:2013}
Simo S\"arkk\"a, Arno Solin, and Jouni Hartikainen.
\newblock Spatiotemporal learning via infinite-dimensional {B}ayesian filtering
  and smoothing.
\newblock \emph{IEEE Signal Processing Magazine}, 30\penalty0 (4):\penalty0
  51--61, 2013.

\bibitem[Sato(2001)]{sato2001online}
Masa-Aki Sato.
\newblock Online model selection based on the variational {B}ayes.
\newblock \emph{Neural Computation}, 13\penalty0 (7):\penalty0 1649--1681,
  2001.

\bibitem[Saul et~al.(2016)Saul, Hensman, Vehtari, and
  Lawrence]{saul2016chained}
Alan~D. Saul, James Hensman, Aki Vehtari, and Neil~D. Lawrence.
\newblock Chained {G}aussian processes.
\newblock In \emph{Proceedings of the 19th International Conference on
  Artificial Intelligence and Statistics (AISTATS)}, volume~51 of
  \emph{Proceedings of Machine Learning Research}, pages 1431--1440. PMLR,
  2016.

\bibitem[Seeger(2005)]{seeger2005expectation}
Matthias Seeger.
\newblock Expectation propagation for exponential families.
\newblock Technical report, University of California at Berkeley, 2005.

\bibitem[Seeger and Nickisch(2011)]{seeger2011fast}
Matthias Seeger and Hannes Nickisch.
\newblock Fast convergent algorithms for expectation propagation approximate
  {B}ayesian inference.
\newblock In \emph{Proceedings of the Fourteenth International Conference on
  Artificial Intelligence and Statistics (AISTATS)}, volume~15 of
  \emph{Proceedings of Machine Learning Research}, pages 652--660. PMLR, 2011.

\bibitem[Shanno(1970)]{shanno1970conditioning}
David~F Shanno.
\newblock Conditioning of quasi-{N}ewton methods for function minimization.
\newblock \emph{Mathematics of computation}, 24\penalty0 (111):\penalty0
  647--656, 1970.

\bibitem[Silverman(1985)]{silverman1985some}
Bernhard~W Silverman.
\newblock Some aspects of the spline smoothing approach to non-parametric
  regression curve fitting.
\newblock \emph{Journal of the Royal Statistical Society: Series B
  (Methodological)}, 47\penalty0 (1):\penalty0 1--21, 1985.

\bibitem[Tebbutt et~al.(2021)Tebbutt, Solin, and Turner]{tebbutt2021combining}
Will Tebbutt, Arno Solin, and Richard~E. Turner.
\newblock Combining pseudo-point and state space approximations for
  sum-separable {G}aussian processes.
\newblock In \emph{Proceedings of the 37th Conference on Uncertainty in
  Artificial Intelligence (UAI)}, Proceedings of Machine Learning Research.
  PMLR, 2021.

\bibitem[Tierney and Kadane(1986)]{tierney1986accurate}
Luke Tierney and Joseph~B Kadane.
\newblock Accurate approximations for posterior moments and marginal densities.
\newblock \emph{Journal of the American Statistical Association}, 81\penalty0
  (393):\penalty0 82--86, 1986.

\bibitem[Tolvanen et~al.(2014)Tolvanen, Jyl{\"a}nki, and
  Vehtari]{tolvanen2014expectation}
Ville Tolvanen, Pasi Jyl{\"a}nki, and Aki Vehtari.
\newblock Expectation propagation for nonstationary heteroscedastic {G}aussian
  process regression.
\newblock In \emph{2014 IEEE International Workshop on Machine Learning for
  Signal Processing (MLSP)}, pages 1--6. IEEE, 2014.

\bibitem[Tran et~al.(2019)Tran, Nguyen, and Nguyen]{tran2019variational}
Minh-Ngoc Tran, Dang~H Nguyen, and Duy Nguyen.
\newblock Variational {B}ayes on manifolds.
\newblock \emph{arXiv preprint arXiv:1908.03097}, 2019.

\bibitem[Turner and Sahani(2011)]{turner2011demodulation}
Richard~E Turner and Maneesh Sahani.
\newblock Demodulation as probabilistic inference.
\newblock \emph{IEEE Transactions on Audio, Speech, and Language Processing},
  19\penalty0 (8):\penalty0 2398--2411, 2011.

\bibitem[Wilkinson et~al.(2021)Wilkinson, Solin, and Adam]{wilkinson2021sparse}
William Wilkinson, Arno Solin, and Vincent Adam.
\newblock Sparse algorithms for {M}arkovian {G}aussian processes.
\newblock In \emph{Proceedings of The 24th International Conference on
  Artificial Intelligence and Statistics (AISTATS)}, volume 130 of
  \emph{Proceedings of Machine Learning Research}, pages 1747--1755. PMLR,
  2021.

\bibitem[Wilkinson et~al.(2020)Wilkinson, Chang, Andersen, and
  Solin]{wilkinson2020state}
William~J. Wilkinson, Paul~E. Chang, Michael~Riis Andersen, and Arno Solin.
\newblock State space expectation propagation: {E}fficient inference schemes
  for temporal {G}aussian processes.
\newblock In \emph{Proceedings of the 37th International Conference on Machine
  Learning (ICML)}, volume 119 of \emph{Proceedings of Machine Learning
  Research}, pages 10270--10281. PMLR, 2020.

\bibitem[Wilson et~al.(2012)Wilson, Knowles, and
  Ghahramani]{wilson2012gaussian}
Andrew~G. Wilson, David~A. Knowles, and Zoubin Ghahramani.
\newblock {G}aussian processes regression networks.
\newblock In \emph{Proceedings of the 29th International Conference on Machine
  Learning (ICML)}. Omnipress, 2012.

\end{thebibliography}

\newpage

\appendix

\section{Derivation of the Online Newton Updates} \label{app:newton-updates}

Newton's method corresponds to approximating
\begin{equation}
	\begin{split}
		\mathcal{L}(\mathbf{f}) \approx \mathcal{L}(\mathbf{m}_k)
		+ \nabla \mathcal{L}(\mathbf{m}_k)^\top \,  (\mathbf{f} - \mathbf{m}_k)
		+ \frac{1}{2} (\mathbf{f} - \mathbf{m}_k)^\top\,  \nabla^2 \mathcal{L}(\mathbf{m}_k) \, (\mathbf{f} - \mathbf{m}_k),
	\end{split}
\end{equation}
whose minimum is given by setting the derivative of the right hand side to zero:
\begin{equation}
	\begin{split}
		\nabla \mathcal{L}(\mathbf{m}_k) + \nabla^2 \mathcal{L}(\mathbf{m}_k) \, (\mathbf{f} - \mathbf{m}_k) = 0,
	\end{split}
\end{equation}
which then gives the next iterate as
\begin{equation}
	\begin{split}
		\mathbf{m}_{k+1} = \mathbf{m}_k - (\nabla^2 \mathcal{L}(\mathbf{m}_k))^{-1} \, \nabla \mathcal{L}(\mathbf{m}_k).
	\end{split}
\end{equation}
We can now set $\mathbf{C}_{k+1} = -(\nabla^2 \mathcal{L}(\mathbf{m}_k))^{-1}$ which allows us to write
\begin{equation}
	\begin{split}
		\mathbf{C}_{k+1}^{-1} &= -(\nabla^2 \mathcal{L}(\mathbf{m}_k))^{-1}, \\
		\mathbf{m}_{k+1} &= \mathbf{m}_k + \mathbf{C}_{k+1} \, \nabla \mathcal{L}(\mathbf{m}_k).
	\end{split}
\end{equation}
However, instead of taking full step of Newton's we can also take a partial step (as is often done via line-search in the optimization literature) which replaces the update by
\begin{equation}
	\begin{split}
		\mathbf{C}_{k+1}^{-1} &= (1 - \rho) \, \mathbf{C}_k - (\nabla^2 \mathcal{L}(\mathbf{m}_k))^{-1}, \\
		\mathbf{m}_{k+1} &= \mathbf{m}_k + \rho \, \mathbf{C}_{k+1} \, \nabla \mathcal{L}(\mathbf{m}_k),
	\end{split}
\end{equation}
as given in \cref{eq:Newton}. The iterates $\mathbf{C}_k$ indeed converge to $\mathbf{C} = -(\nabla^2 \mathcal{L}(\mathbf{m}^*))^{-1}$ which is the Laplace approximation to the posterior covariance.

The following is a more detailed derivation of the result given in \cref{eq:Newton-natural},
\begin{gather}
	\begin{aligned}
		\postnattwo_{k+1} := -\frac{1}{2} \postcov_{k+1}^{-1} &= -(1-\rho)\frac{1}{2} \postcov_{k}^{-1} - \rho \, \frac{1}{2} \left(\priorcov^{-1} - \nabla_\vfunc^2\log p(\vy\mid \postmean_k) \right) \\
		&= -(1-\rho)\frac{1}{2} (\priorcov^{-1} + \likcov_k^{-1})- \rho \, \frac{1}{2} \left(\priorcov^{-1} - \nabla_\vfunc^2\log p(\vy\mid \postmean_k) \right) \\
		&= \priornattwo + (1-\rho) \liknattwo_k + \rho \, \frac{1}{2}\nabla_\vfunc^2\log p(\vy\mid \postmean_k) \,, \\
		\postnatone_{k+1} := \postcov^{-1}_{k+1} \postmean_{k+1} &= \postcov^{-1}_{k+1} \postmean_k + \rho \, \nabla_\vfunc \log p(\vy \mid \postmean_k) - \rho \, \MK^{-1}(\postmean_k-\priormean) \\
		&=  (1-\rho) \, \postcov^{-1}_{k} \postmean_k + \rho \, \priorcov^{-1}\priormean + \rho \, \left( \nabla_\vfunc \log p(\vy \mid \postmean_k) - \nabla_\vfunc^2\log p(\vy\mid \postmean_k) \, \postmean_k \right) \\
		&= \priornattwo \postmean_k -\rho\, \priorcov^{-1} \postmean_{k} + \rho \, \priorcov^{-1}\priormean + \rho \, \left( \nabla_\vfunc \log p(\vy \mid \postmean_k) - \nabla_\vfunc^2\log p(\vy\mid \postmean_k) \, \postmean_k \right) \\
		&= \priornatone + (1-\rho)\liknatone_k + \rho \, \big(\nabla_\vfunc \log p(\vy \mid \postmean_k) - \nabla_\vfunc^2\log p(\vy\mid \postmean_k) \, \postmean_k \big) \,.
	\end{aligned}
\end{gather}

\section{Derivation of the VI Updates} \label{app:VI}

The following is a more detailed derivation of the result given in \cref{eq:vi-natural},
\begin{align}
	\natparams_{k+1} &= \natparams_k - \rho \, \nabla_{\meanparams} \text{VFE}(q(\vfunc)) \nonumber \\
	&= \natparams_k - \rho \, (-\nabla_{\meanparams}\E_{q(\vfunc)}[\log p(\vy, \vfunc)] + \nabla_{\meanparams}\E_{q(\vfunc)}[\log q(\vfunc)] ) \nonumber \\
	&= \natparams_k - \rho \, (-\nabla_{\meanparams}\E_{q(\vfunc)}[\log p(\vy, \vfunc)] + \natparams_k ) \nonumber \\
	&= (1 - \rho)\natparams_k + \rho \, \nabla_{\meanparams}\E_{q(\vfunc)}[\log p(\vy, \vfunc)] .
\end{align}
By application of the chain rule \citep[see][for a detailed explanation of this step]{khan2021bayesian}, the individual posterior parameter updates then become
\begin{gather}
	\begin{aligned}
		\postnattwo_{k+1} &= (1-\rho) \postnattwo_k + \rho \, \frac{1}{2} \nabla^2_{\vm} \E_{q(\vfunc)}[ \log p(\vy, \vfunc) ] \\
		&= (1-\rho) \postnattwo_k + \rho \left( \frac{1}{2}\nabla^2_{\vm} \E_{q(\vfunc)}[ \log p(\vy \mid  \vfunc) ] + \priornattwo \right) \\
		&= \priornattwo + (1-\rho) \liknattwo_k + \rho \, \frac{1}{2} \nabla^2_{\vm} \E_{q(\vfunc)}[ \log p(\vy \mid  \vfunc) ] ,  \\
		\postnatone_{k+1} &= (1-\rho) \postnatone_k + \rho \left( \nabla_{\vm} \E_{q(\vfunc)}[ \log p(\vy , \vfunc)] - \nabla^2_{\vm} \E_{q(\vfunc)}[ \log p(\vy, \vfunc) ] \, \postmean_k \right) \\
		&= (1-\rho) \postnatone_k + \rho \left( \nabla_{\vm} \E_{q(\vfunc)}[ \log p(\vy \mid \vfunc)] - \priorcov^{-1} (\postmean_k - \priormean) - (\nabla^2_{\postmean} \E_{q(\vfunc)}[ \log p(\vy \mid \vfunc) ] - \priorcov^{-1}) \, \postmean_k \right) \\
		&= \priornatone + (1-\rho) \liknatone_k + \rho \left( \nabla_{\postmean} \E_{q(\vfunc)}[ \log p(\vy \mid \vfunc)] - \nabla^2_{\postmean} \E_{q(\vfunc)}[ \log p(\vy \mid \vfunc) ] \, \postmean_k \right) .
	\end{aligned}
\end{gather} 

\section{Derivation of the PEP Updates} \label{app:EP}

It is tempting to view PEP as minimising an alternative free energy approximation, and to apply natural gradient descent to the approximate energy as in the VI case. However, this approach would be flawed since this energy is not a bound for the true energy, and stationary points obtained via moment matching may not even be local minima \citep{opper2005expectation}. 

The PEP algorithm proceeds by minimising local KL divergences,
\begin{gather}
	\begin{aligned}
		t(\vfunc_n) &\leftarrow \mathrm{arg}\,\min_{t_*(\vfunc_n)} \KLbig{\frac{1}{Z_n}\frac{p^\alpha(\vy_n \mid \vfunc_n)}{t^\alpha(\vfunc_n)} q(\vfunc_n)} {\frac{1}{W_n}\frac{t_*^\alpha(\vfunc_n)}{t^\alpha(\vfunc_n)} q(\vfunc_n)},
	\end{aligned}
\end{gather}
where $Z_n=\int \frac{p^\alpha(\vy_n \mid \vfunc_n)}{t^\alpha(\vfunc_n)} q(\vfunc_n) \mathrm{d}\vfunc_n$ and $W_n= \int \frac{t_*^\alpha(\vfunc_n)}{t^\alpha(\vfunc_n)} q(\vfunc_n) \mathrm{d}\vfunc_n$. Since the right hand side is Gaussian, this minimisation amounts to moment matching. Hence we must compute the first two moments of $\frac{1}{Z_n}\frac{p^\alpha(\vy_n \mid \vfunc_n)}{t^\alpha(\vfunc_n)} q(\vfunc_n)$. To derive the moment matching equations we take the derivatives of $Z_n$ w.r.t.\ the cavity mean (we let $\N(\vfunc_n \mid \cavmean, \cavcov)$ be the marginal cavity mean),
	\begin{align}
		\frac{\partial Z_n}{\partial \cavmean} &= {\cavcovfull}_{n,n}^{-1} \int  (\vfunc_n - \cavmean) \frac{p^\alpha(\vy_n \mid \vfunc_n)}{t^\alpha(\vfunc_n)}q(\vfunc_n) \mathrm{d}\vfunc_n \nonumber \\
		&= {\cavcovfull}_{n,n}^{-1} \int \vfunc_n \frac{p^\alpha(\vy_n \mid \vfunc_n)}{t^\alpha(\vfunc_n)}  q(\vfunc_n) \mathrm{d}\vfunc_n - {\cavcovfull}_{n,n}^{-1} \cavmean \int \frac{p^\alpha(\vy_n \mid \vfunc_n)}{t^\alpha(\vfunc_n)}  q(\vfunc_n) \mathrm{d}\vfunc_n \nonumber \\
		&= {\cavcovfull}_{n,n}^{-1} Z_n \int \vfunc_n Z_n^{-1} \frac{p^\alpha(\vy_n \mid \vfunc_n)}{t^\alpha(\vfunc_n)}  q(\vfunc_n) \mathrm{d}\vfunc_n - {\cavcovfull}_{n,n}^{-1} \cavmean Z_n \nonumber \\
		&= {\cavcovfull}_{n,n}^{-1} Z_n \E_{\tilde{q}} [\vfunc_n] - {\cavcovfull}_{n,n}^{-1} \cavmean Z_n,
	\end{align}
and rearranging the terms gives
	\begin{align}
		\E_{\tilde{q}} [\vfunc_n] &= \cavmean + {\cavcov} \frac{\partial Z_n}{\partial \cavmean} Z_n^{-1} \nonumber \\
		&= \cavmean + {\cavcov} \frac{\partial \log Z_n}{\partial \cavmean}.
	\end{align}
Differentiating again we get,
	\begin{align}
		\frac{\partial^2 Z_n}{\partial \cavmean \partial {\cavmean}^\T} &= {\cavcovfull}_{n,n}^{-1} \int (\vfunc_n - \cavmean)(\vfunc_n - \cavmean)^\T  \frac{p^\alpha(\vy_n \mid \vfunc_n)}{t^\alpha(\vfunc_n)} q(\vfunc_n) \mathrm{d}\vfunc_n {\cavcovfull}_{n,n}^{-1} \nonumber \\
		&\quad\quad - {\cavcovfull}_{n,n}^{-1} \int \frac{p^\alpha(\vy_n \mid \vfunc_n)}{t^\alpha(\vfunc_n)}  q(\vfunc_n) \mathrm{d}\vfunc_n \nonumber \\
		&= {\cavcovfull}_{n,n}^{-1} \int \vfunc_n \vfunc_n^\T \frac{p^\alpha(\vy_n \mid \vfunc_n)}{t^\alpha(\vfunc_n)}  q(\vfunc_n) \mathrm{d}\vfunc_n {\cavcovfull}_{n,n}^{-1} - 2 {\cavcovfull}_{n,n}^{-1} \cavmean \int \vfunc_n \frac{p^\alpha(\vy_n \mid \vfunc_n)}{t^\alpha(\vfunc_n)}  q(\vfunc_n) \mathrm{d}\vfunc_n {\cavcovfull}_{n,n}^{-1} \nonumber \\
		& \quad \quad + {\cavcovfull}_{n,n}^{-1} \cavmean {\cavmean}^\T \int \frac{p^\alpha(\vy_n \mid \vfunc_n)}{t^\alpha(\vfunc_n)}  q(\vfunc_n) \mathrm{d}\vfunc_n {\cavcovfull}_{n,n}^{-1} - {\cavcovfull}_{n,n}^{-1} \int \frac{p^\alpha(\vy_n \mid \vfunc_n)}{t^\alpha(\vfunc_n)}  q(\vfunc_n) \mathrm{d}\vfunc_n \nonumber \\
		&= {\cavcovfull}_{n,n}^{-1} Z_n \int \vfunc_n \vfunc_n^\T Z_n^{-1} \frac{p^\alpha(\vy_n \mid \vfunc_n)}{t^\alpha(\vfunc_n)}  q(\vfunc_n) \mathrm{d}\vfunc_n {\cavcovfull}_{n,n}^{-1} \nonumber \\
		& \quad\quad - 2 {\cavcovfull}_{n,n}^{-1} \cavmean Z_n \int \vfunc_n Z_n^{-1} \frac{p^\alpha(\vy_n \mid \vfunc_n)}{t^\alpha(\vfunc_n)}  q(\vfunc_n) \mathrm{d}\vfunc_n {\cavcovfull}_{n,n}^{-1} \nonumber \\
		& \quad \quad + {\cavcovfull}_{n,n}^{-1} \cavmean {\cavmean}^\T Z_n {\cavcovfull}_{n,n}^{-1} - {\cavcovfull}_{n,n}^{-1} Z_n \nonumber \\
		&= {\cavcovfull}_{n,n}^{-1} Z_n \E_{\tilde{q}} [\vfunc_n \vfunc_n^\T] {\cavcovfull}_{n,n}^{-1} - 2 {\cavcovfull}_{n,n}^{-1} \cavmean Z_n \E_{\tilde{q}} [\vfunc_n] {\cavcovfull}_{n,n}^{-1} \nonumber \\
		& \quad\quad + {\cavcovfull}_{n,n}^{-1} \cavmean {\cavmean}^\T Z_n {\cavcovfull}_{n,n}^{-1} - {\cavcovfull}_{n,n}^{-1} Z_n
	\end{align}
which gives
	\begin{equation}
		\E_{\tilde{q}} [\vfunc_n \vfunc_n^\T] = 2 \cavmean \E_{\tilde{q}} [\vfunc_n]  -\cavmean {\cavmean}^\T + {\cavcov} + {\cavcov} \frac{\partial^2 Z_n}{\partial \cavmean \partial {\cavmean}^\T} Z_n^{-1} {\cavcov}, \\
	\end{equation}
	\begin{align}
		\text{Cov}_{\tilde{q}}[\vfunc_n] &= \E_{\tilde{q}} [\vfunc_n \vfunc_n^\T] - \E_{\tilde{q}} [\vfunc_n] \E_{\tilde{q}} [\vfunc_n]^\T \nonumber \\
		&= 2 \cavmean {\cavmean}^\T + 2 \cavmean {\cavcov} \frac{\partial Z_n}{\partial \cavmean} Z_n^{-1}  -\cavmean {\cavmean}^\T + {\cavcov} + {\cavcov} \frac{\partial^2 Z_n}{\partial \cavmean \partial {\cavmean}^\T} Z_n^{-1} {\cavcov} \nonumber \\
		& \quad - \cavmean {\cavmean}^\T - 2  \cavmean {\cavcov} \frac{\partial Z_n}{\partial \cavmean} Z_n^{-1} - {\cavcov} \frac{\partial Z_n}{\partial \cavmean} \frac{\partial Z_n}{\partial \cavmean}^\T Z_n^{-2} {\cavcov} \nonumber \\
		&= {\cavcov} + {\cavcov} \left( \frac{\partial^2 Z_n}{\partial \cavmean \partial {\cavmean}^\T} Z_n^{-1} - \frac{\partial Z_n}{\partial \cavmean} \frac{\partial Z_n}{\partial \cavmean}^\T Z_n^{-2}\right) {\cavcov} \nonumber \\
		&= {\cavcov} + {\cavcov} \frac{\partial^2 \log Z_n}{\partial \cavmean \partial {\cavmean}^\T} {\cavcov}.
	\end{align}
Now removing the cavity contribution from these posterior moments and scaling by the inverse power (since substituting the cavity results in a fraction $\alpha$ of the full likelihood) gives the new approximate likelihood parameter updates,
	\begin{align}
		\likcov_{n,n}^{-1} & = \frac{1}{\alpha}\left( {\cavcov} + {\cavcov} \frac{\partial^2 \log Z_n}{\partial \cavmean \partial {\cavmean}^\T} {\cavcov} \right)^{-1} - \frac{1}{\alpha}{\cavcovfull}_{n,n}^{-1} \nonumber \\
		&= \frac{1}{\alpha}\left( \MI + \frac{\partial^2 \log Z_n}{\partial \cavmean \partial {\cavmean}^\T} {\cavcov} \right)^{-1} {\cavcovfull}_{n,n}^{-1} - \frac{1}{\alpha} {\cavcovfull}_{n,n}^{-1} \nonumber \\
		&= \frac{1}{\alpha}\left( \MI + \frac{\partial^2 \log Z_n}{\partial \cavmean \partial {\cavmean}^\T} {\cavcov} \right)^{-1} {\cavcovfull}_{n,n}^{-1} - \frac{1}{\alpha} \left( \MI + \frac{\partial^2 \log Z_n}{\partial \cavmean \partial {\cavmean}^\T} {\cavcov} \right)^{-1} \left( \MI + \frac{\partial^2 \log Z_n}{\partial \cavmean \partial {\cavmean}^\T} {\cavcov} \right) {\cavcovfull}_{n,n}^{-1} \nonumber \\
		&= \frac{1}{\alpha} \left(  \MI + \frac{\partial^2 \log Z_n}{\partial \cavmean \partial {\cavmean}^\T} {\cavcov}\right)^{-1} \left( - \frac{\partial^2 \log Z_n}{\partial \cavmean \partial {\cavmean}^\T} \right) \nonumber \\
		&= \frac{1}{\alpha}{\cavcovfull}_{n,n}^{-1} \left( {\cavcovfull}_{n,n}^{-1} + \frac{\partial^2 \log Z_n}{\partial \cavmean \partial {\cavmean}^\T} \right)^{-1} \left( - \frac{\partial^2 \log Z_n}{\partial \cavmean \partial {\cavmean}^\T} \right),
	\end{align}
where the re-arrangement on the last line is to ensure we only invert a symmetric matrix (note that whilst the quantity being inverted is guaranteed to be symmetric, it is not guaranteed to be PSD).
	\begin{align}
		\liknatone_n & = \frac{1}{\alpha}\left( {\cavcov} + {\cavcov} \frac{\partial^2 \log Z_n}{\partial \cavmean \partial {\cavmean}^\T} {\cavcov} \right)^{-1} \left( \cavmean + {\cavcov} \frac{\partial \log Z_n}{\partial \cavmean} \right) - \frac{1}{\alpha} {\cavcovfull}_{n,n}^{-1} \cavmean \nonumber \\
		&= \frac{1}{\alpha} \left( \MI +  \frac{\partial^2 \log Z_n}{\partial \cavmean \partial {\cavmean}^\T} {\cavcov} \right)^{-1}  \left( {\cavcovfull}_{n,n}^{-1} \cavmean + \frac{\partial \log Z_n}{\partial \cavmean} \right) - \frac{1}{\alpha} {\cavcovfull}_{n,n}^{-1} \cavmean \nonumber \\
		&= \frac{1}{\alpha} \left( \MI + \frac{\partial^2 \log Z_n}{\partial \cavmean \partial {\cavmean}^\T} {\cavcovfull}_{n,n}^{-1} \right)^{-1} \left(\frac{\partial \log Z_n}{\partial \cavmean} - \frac{\partial^2 \log Z_n}{\partial \cavmean \partial {\cavmean}^\T} \cavmean \right) \nonumber \\
		&= \frac{1}{\alpha} {\cavcovfull}_{n,n}^{-1} \left( {\cavcovfull}_{n,n}^{-1} + \frac{\partial^2 \log Z_n}{\partial \cavmean \partial {\cavmean}^\T} \right)^{-1} \left(\frac{\partial \log Z_n}{\partial \cavmean} - \frac{\partial^2 \log Z_n}{\partial \cavmean \partial {\cavmean}^\T} \cavmean \right).
	\end{align}

\section{Equivalence of PEP $\alpha \rightarrow 0$ and Natural Gradient VI} \label{app:PEP-VI}

Following \citet{bui2017unifying}, we utilise the Maclaurin series, $\exp(x)=1 + x + \frac{x^2}{2!} + \frac{x^3}{3!} + \dots$, to write
	\begin{align}
		p^\alpha(\vy_n \mid \vfunc_n) &= \exp(\alpha \log p(\vy_n \mid \vfunc_n)) \nonumber \\
		&= 1 + \alpha \log p(\vy_n \mid \vfunc_n) + \underbrace{\frac{1}{2!}(\alpha \log p(\vy_n \mid \vfunc_n))^2+ \frac{1}{3!}(\alpha \log p(\vy_n \mid \vfunc_n))^3+ \dots} _{\alpha^2\psi(\cdot)} ,
	\end{align}
which leads to (using the series $\log(1+x)=x-\frac{x^2}{2!}+\frac{x^3}{3!}-\dots$),
	\begin{align}
		\frac{1}{\alpha}\log \E_{q^{\cav }(\vfunc_n)} [ p^\alpha(\vy_n \mid \vfunc_n) ] & = \frac{1}{\alpha}\log \int q^{\cav }(\vfunc_n) [1 + \alpha \log p(\vy_n \mid \vfunc_n) + \alpha^2\psi(\cdot)] \mathrm{d}\vfunc_n \nonumber \\
		& = \frac{1}{\alpha}\log \left[ 1 + \alpha\int q^{\cav }(\vfunc_n) \log p(\vy_n \mid \vfunc_n)  \mathrm{d}\vfunc_n + \alpha^2\psi(\cdot) \right] \nonumber \\
		& = \int q^{\cav }(\vfunc_n) \log p(\vy_n \mid \vfunc_n)  \mathrm{d}\vfunc_n + \alpha\psi(\cdot) , 
	\end{align}
therefore,
\begin{equation}
	\lim_{\alpha\rightarrow 0} \frac{1}{\alpha} \log \E_{q^{\cav }(\vfunc_n)} [ p^\alpha(\vy_n \mid \vfunc_n) ] = \E_{q(\vfunc_n)} [ \log p(\vy_n \mid \vfunc_n) ] .
\end{equation}
It is also the case that the PEP scaling factor reverts to the identity in the limit,
\begin{gather}
	\begin{aligned}
		\MR_n &= {\cavcovfull_n}^{-1} \left( \alpha \nabla_{\cavmeanfull_n}^2\LLbar(\cavmeanfull_{k,n}) + {\cavcovfull_n}^{-1} \, \right)^{-1}  \\
		&= \left( \alpha \nabla_{\cavmeanfull_n}^2\LLbar(\cavmeanfull_{k,n}) {\cavcovfull_n} + \MI \, \right)^{-1} \\
		\implies \lim_{\alpha\rightarrow 0} \MR_n &= \MI
	\end{aligned}
\end{gather}
which shows that the PEP updates given by \cref{eq:pep-target} when $\alpha \rightarrow 0$ are equivalent to the natural gradient VI updates given by \cref{eq:vi-target}.

\section{Derivation of the PL Updates} \label{app:PL}

After performing SLR on a single likelihood term, we obtain the approximation
\begin{equation}
	p(\vy_n \mid \vfunc_n) \approx \N(\vy_n \mid \MA_n \vfunc_n + \vb_n, \MOmega_{n,n}),
\end{equation}
where $\MA_n=\MQ_n^\T \postcov_{n,n}^{-1}$, $\vb_n=\E_{q(\vfunc_n)} [ \E[\vy_n\mid \vfunc_n] ]-\MQ_n^\T \postcov_{n,n}^{-1}\postmean_n$, and $\MOmega_{n,n}=\MS_n-\MQ_n^\T \postcov_{n,n}^{-1} \MQ_n$ for,
\begin{gather} \label{eq:gauss-filt-components}
	\begin{aligned}
		\MS_n = & \E_{q(\vfunc_n)} \left[ (\E[\vy_n\mid \vfunc_n] - \E_{q(\vfunc_n)}\left[\E[\vy_n\mid \vfunc_n] \right]) (\E[\vy_n\mid \vfunc_n] - \E_{q(\vfunc_n)}\left[\E[\vy_n\mid \vfunc_n] \right])^\T+\Cov[\vy_n\mid \vfunc_n]\right] , \\
		\MQ_n = & \E_{q(\vfunc_n)} \left[ (\vfunc_n - \postmean_n) (\E[\vy_n\mid \vfunc_n] - \E_{q(\vfunc_n)}\left[\E[\vy_n\mid \vfunc_n] \right])^\T \right].
	\end{aligned}
\end{gather}
Given this Gaussian term, the question remains what its corresponding factor $t(\vfunc_n)$ is, such that PL can be presented in the same light as the other approximate inference methods. If $\MA_n$ is square and invertible, then finding $t(\vfunc_n)$ is trivial, but we are interested in the more general case where the dimension of $\vfunc_n$ and $\vy_n$ may not be the same.

Fortunately, the EP and VI updates provide a means by which to compute a factor $t(\vfunc_n)$ given a general likelihood model, and furthermore we know that these methods are exact in the Gaussian case (which PL provides after linearisation). Since the VI updates are slightly simpler than the EP ones, we proceed by treating $ \N(\vy_n \mid \MA_n \vfunc_n + \vb_n, \MOmega_{n,n})$ as the target distribution in \cref{eq:vi-target}, which gives
	\begin{align}
		\LL_n &= \E_{q(\vfunc_n)} [ \log \N(\vy_n \mid \MA_n \vfunc_n + \vb_n, \MOmega_{n,n}) ] \nonumber \\
		&= -\frac{1}{2}\log 2\pi - \frac{1}{2} \log \MOmega_{n,n} - \frac{1}{2} ( (\vy_n - \MA_n \postmean_n - \vb_n)\MOmega_{n,n}^{-1} (\vy_n - \MA_n \postmean_n - \vb_n)^\T + \postcov_{n,n}) , \nonumber \\
		&= -\frac{1}{2}\log 2\pi - \frac{1}{2} \log \MOmega_{n,n} - \frac{1}{2} ( (\vy_n - \E_{q(\vfunc_n)} [ \E[\vy_n\mid \vfunc_n] ]) \MOmega_{n,n}^{-1} (\vy_n - \E_{q(\vfunc_n)} [ \E[\vy_n\mid \vfunc_n] ])^\T + \postcov_{n,n}) , \nonumber \\
		&= c - \frac{1}{2}  (\vy_n - \E_{q(\vfunc_n)} [ \E[\vy_n\mid \vfunc_n] ]) \MOmega_{n,n}^{-1} (\vy_n - \E_{q(\vfunc_n)} [ \E[\vy_n\mid \vfunc_n] ])^\T , 
	\end{align}
where $c$ is constant and contains the terms that do not depend on $\postmean_n$ (and hence do not effect the updates). Differentiating with respect to $\postmean_n$ gives
\begin{gather}
	\begin{aligned}
		\frac{\partial \LL_n}{\partial \postmean_n^\T} &= {\frac{\partial \E_{q(\vfunc_n)} [ \E[\vy_n\mid \vfunc_n] ] }{\partial \postmean_n}}^\T \MOmega_{n,n}^{-1} (\vy_n - \E_{q(\vfunc_n)} [ \E[\vy_n\mid \vfunc_n] ] ) , \\
		\frac{\partial^2 \LL_n}{\partial \postmean_n^\T \partial \postmean_n} &= -{\frac{\partial \E_{q(\vfunc_n)} [ \E[\vy_n\mid \vfunc_n] ] }{\partial \postmean_n}}^\T \MOmega_{n,n}^{-1} \frac{\partial \E_{q(\vfunc_n)} [ \E[\vy_n\mid \vfunc_n] ] }{\partial \postmean_n} , \\
	\end{aligned}
\end{gather}
which gives the desired updates in \cref{eq:pl-target}.

\section{Posterior Linearisation as a Bayes--Gauss--Newton Method} \label{sec:pl-gauss-newton}

Posterior linearisation can be characterised as a Gauss--Newton algorithm. Whilst it is less clear what optimisation target PL minimises, the updates can be re-derived as the result of a least-squares problem by defining the surrogate target $\LLbar(\postmean_n, \postcov_{n,n}) = \log \N(\vy_n \mid \E_{q(\vfunc_n)} [ \E[\vy_n\mid \vfunc_n] ], \, \MOmega_{n,n})$, and the least-squares residuals
\begin{equation}
	\MV(\postmean, \postcov)=\left[ 
	\begin{matrix}
		\MOmega_{1,1}^{-\frac{1}{2}} (\vy_1 - \E_{q(\vfunc_1)} [\E[\vy_1\mid \vfunc_1]  ]) \\
		\vdots \\
		\MOmega_{N,N}^{-\frac{1}{2}} (\vy_N - \E_{q(\vfunc_N)} [ \E[\vy_N\mid \vfunc_N]  ]) \\
	\end{matrix}
	\right] ,
\end{equation}
such that $\LLbar(\postmean, \postcov)=\MV(\postmean, \postcov)^\T \MV(\postmean, \postcov)$. The residual Jacobian is then defined as
\begin{equation}
	\nabla_\postmean \MV(\postmean, \postcov) =\left[
	\begin{matrix}
		-\MOmega_{1,1}^{-\frac{1}{2}} \nabla_{\postmean^\T} \E_{q(\vfunc_1)} [\E[\vy_1\mid \vfunc_1]  ] \\
		\vdots \\
		-\MOmega_{N,N}^{-\frac{1}{2}} \nabla_{\postmean^\T} \E_{q(\vfunc_N)} [\E[\vy_N\mid \vfunc_N]  ] \\
	\end{matrix}
	\right] ,
\end{equation}
where
\begin{equation}
	-\MOmega_{n,n}^{-\frac{1}{2}} \nabla_{\postmean^\T} \E_{q(\vfunc_n)} [\E[\vy_n\mid \vfunc_n]  ]  = \Big[ \bm{0}, \dots, -\MOmega_{n,n}^{-\frac{1}{2}} \nabla_{\postmean_n} \E_{q(\vfunc_n)} [\E[\vy_n\mid \vfunc_n]  ] , \dots, \bm{0} \Big] .
\end{equation}
Use of the Gauss--Newton approximation to the Hessian results in
\begin{align}
	\nabla^2_\postmean \LLbar(\postmean, \postcov) &\approx-\nabla_\postmean \MV(\postmean, \postcov)^\T \nabla_\postmean \MV(\postmean, \postcov) \nonumber \\
	&= -\nabla_{\postmean}\E_{q(\vfunc)} [\E[\vy\mid \vfunc]  ]^\T \MOmega^{-1} \nabla_{\postmean} \E_{q(\vfunc)} [\E[\vy\mid \vfunc]  ] ,
\end{align} 
which matches the PL updates. Note that it has been assumed that the gradient of $\MOmega_k$ is zero (see \cref{sec:improved-pl} for discussion). Since these terms take into account the full Bayesian posterior via the expectation with respect to $q(\vfunc)$, rather than a point estimate, we refer to PL as a \emph{Bayes--Gauss--Newton} method.

\section{Taylor Expansion / Extended Kalman Smoother as a Gauss--Newton Method} \label{sec:eks-gauss-newton}

It is well known that the Taylor/EKS approach given in \cref{sec:eks} is equivalent to a Gauss--Newton method \citep{bell1994iterated}. This can be seen by modifying the PL approach in \cref{sec:pl-gauss-newton} by setting $\MOmega=\Cov[\vy\mid \vfunc]$ and replacing the expectations with respect to $q(\vfunc)$ with point estimates at the mean. This gives
\begin{equation}
	\MV(\vfunc)=\left[ 
	\begin{matrix}
		\Cov[\vy_1\mid \vfunc_1]^{-\frac{1}{2}} (\vy_1 - \E[\vy_1\mid \vfunc_1]  ) \\
		\vdots \\
		\Cov[\vy_N\mid \vfunc_N]^{-\frac{1}{2}} (\vy_N - \E[\vy_N\mid \vfunc_N]  ) \\
	\end{matrix}
	\right] ,
\end{equation}
such that $\LLbar(\vfunc)=\log \N(\vy\mid \E[\vy\mid \vfunc],\Cov[\vy\mid \vfunc] )=\MV(\vfunc)^\T \MV(\vfunc)+c$. The residual Jacobian is then defined as
\begin{equation}
	\nabla_\vfunc \MV(\vfunc) =\left[
	\begin{matrix}
		-\Cov[\vy_1\mid \vfunc_1]^{-\frac{1}{2}} \nabla_{\vfunc^\T} \E[\vy_1\mid \vfunc_1]   \\
		\vdots \\
		-\Cov[\vy_N\mid \vfunc_N]^{-\frac{1}{2}} \nabla_{\vfunc^\T} \E[\vy_N\mid \vfunc_N]   \\
	\end{matrix}
	\right] ,
\end{equation}
where
\begin{equation}
	-\Cov[\vy_n\mid \vfunc_n]^{-\frac{1}{2}} \nabla_{\vfunc^\T} \E[\vy_n\mid \vfunc_n]  = \Big[ \bm{0}, \dots, -\Cov[\vy_n\mid \vfunc_n]^{-\frac{1}{2}} \nabla_{\vfunc_n} \E[\vy_n\mid \vfunc_n]  ] , \dots, \bm{0} \Big] .
\end{equation}
Use of the Gauss--Newton approximation to the Hessian results in,
\begin{align}
	\nabla^2_\vfunc \LLbar(\postmean) &\approx-\nabla_\vfunc \MV(\postmean)^\T \nabla_\vfunc \MV(\postmean) \nonumber \\
	&= -\nabla_{\vfunc}\E[\vy\mid \postmean]^\T \Cov[\vy\mid \vfunc]^{-1} \nabla_{\vfunc} \E[\vy\mid \postmean]   ,
\end{align} 
where $\nabla_{\vfunc} \E[\vy\mid \postmean]:=\nabla_{\vfunc} \E[\vy\mid \vfunc]_{\vfunc=\postmean}$, which matches the EKS updates. Again note that it has been assumed that the gradient of $\Cov[\vy\mid\vfunc]$ is zero.

\section{Full Algorithm Descriptions}\label{sec:algos}

Here we outline the exact algorithms used to perform inference in GPs, sparse GPs and Markovian GPs. The `update rule' $\Lambda(\cdot)$ can be chosen to be any of the local likelihood updates listed in the main paper. Newton: \cref{eq:newton-target}, variational inference: \cref{eq:vi-target}, power expectation propagation: \cref{eq:pep-target} (update rule acts on cavity marginals rather than posterior marginals), posterior linearisation: \cref{eq:pl-target}, extended Kalman smoother: \cref{eq:taylor-target}, second-order posterior linearisation: \cref{eq:pl-newton-target}, quasi-Newton: \cref{eq:quasi-newton-target}, variational quasi-Newton: \cref{eq:variational-quasi-newton-target}, power expectation propagation quasi-Newton: \cref{eq:quasi-newton-pep-target}, posterior linearisation quasi-Newton: \cref{eq:pl-quasi-newton-target}. A Gauss--Newton approximation to the Hessian can be obtained by applying \cref{eq:gauss-newton}, \cref{eq:pgn}, \cref{eq:ggn}, \cref{eq:variational-gn}, or \cref{eq:vggn}. PSD constraints via Riemannian gradients can be applied using \cref{eq:vi-target-psd} or \cref{eq:pep-target-psd}.

\begin{center}
\begin{minipage}[t]{0.75\textwidth}
		\begin{algorithm}[H]
			\footnotesize
			\caption{Gaussian process inference}
			\label{alg:gp}
			\begin{algorithmic}
				\STATE {\bfseries Input:} data: $\{\MX, \vy\}$, kernel: $\kappa(\cdot)$, update rule: $\Lambda(\cdot)$, initial approx. likelihood: $\{\liknatone, \liknattwo\}$, learning rate: $\rho$
				\STATE $\priorcov=\kappa(\MX,\MX)$
				\WHILE{energy not converged}
				\STATE Convert approximate likelihood to mean/cov:
				\STATE $\likcov = -\frac{1}{2}(\liknattwo)^{-1}$
				\STATE $\likmean=\likcov\liknatone$
				\STATE Compute the approximate posterior:
				\STATE $\postmean = \priorcov (\priorcov + \likcov)^{-1} \likmean$
				\STATE $\postcov =	\priorcov - \priorcov (\priorcov + \likcov)^{-1} \priorcov$
				\STATE Update the approximate likelihood:
				\STATE $\jacobian_n, \hessian_{n,n} =\Lambda(\postmean_n, \postcov_{n,n}, \vy_n) \quad \forall n$
				\STATE $\liknattwo = (1-\rho) \, \liknattwo + \rho \, \frac{1}{2} \hessian$
				\STATE $\liknatone = (1-\rho) \, \liknatone + \rho \, \left( \jacobian -\hessian \, \postmean \right)$
				\ENDWHILE
			\end{algorithmic}
		\end{algorithm}
\end{minipage}
\end{center}

\begin{center}
\begin{minipage}[t]{0.75\textwidth}
		\begin{algorithm}[H]
			\footnotesize
			\caption{Stochastic sparse GP inference}
			\label{alg:sgp}
			\begin{algorithmic}
				\STATE {\bfseries Input:} data: $\{\MX, \vy\}$, inducing inputs: $\MZ$, kernel: $\kappa(\cdot)$, update rule: $\Lambda(\cdot)$, initial approx. likelihood: $\{\liknatone, \liknattwo\}$, learning rate: $\rho$, batch size: $N_b$\\
				\STATE $\MK_{\vu\vu}=\kappa(\MZ,\MZ)$
				\STATE $\MK_{\vfunc_n\!\vu}=\kappa(\MX_n,\MZ) \quad \forall n$
				\STATE $\MW_{\vfunc_n\!\vu}=\MK_{\vfunc_n\!\vu}\MK_{\vu\vu}^{-1} \quad \forall n$
				\WHILE{energy not converged}
				\STATE Convert approximate likelihood to mean/cov:
				\STATE $\likcov_\vu = -\frac{1}{2}(\MW_{\vfunc\vu}^\T \liknattwo \MW_{\vfunc\vu})^{-1}$
				\STATE $\likmean_\vu=\likcov_\vu\MW_{\vfunc\vu}^\T \liknatone$
				\STATE Compute the approximate posterior:
				\STATE $\postmean_\vu = \priorcov_{\vu\vu} (\priorcov_{\vu\vu} + \likcov_\vu)^{-1} \likmean_\vu$
				\STATE $\postcov_\vu =	\priorcov_{\vu\vu} - \priorcov_{\vu\vu} (\priorcov_{\vu\vu} + \likcov_\vu)^{-1} \priorcov_{\vu\vu}$
				\STATE Compute the posterior marginals $\forall n$ in batch:
				\STATE $\postmean_n = \MW_{\vfunc_n\vu} \postmean_\vu$
				\STATE $\postcov_{n,n} = \kappa(\MX_n,\MX_n) - \MW_{\vfunc_n\!\vu}\MK_{\vfunc_n\!\vu}^\T + \MW_{\vfunc_n\!\vu} \postcov_\vu \MW_{\vfunc_n\!\vu}^\T$ \hspace{-1em}
				\STATE Update the approximate likelihood:
				\STATE $\jacobian_n, \hessian_{n,n} =\Lambda(\postmean_n, \postcov_{n,n}, \vy_n) \quad \forall n$ in batch
				\STATE $\liknattwo = (1-\rho) \, \liknattwo + \rho \, \frac{1}{2} \hessian$
				\STATE $\liknatone = (1-\rho) \, \liknatone + \rho \, \left( \jacobian -\hessian \, \postmean \right)$
				\ENDWHILE
			\end{algorithmic}
		\end{algorithm}
\end{minipage}
\end{center}

\begin{center}
\begin{minipage}[t]{0.75\textwidth}
		\begin{algorithm}[H]
			\footnotesize
			\caption{State space model / Markovian GP inference}
			\label{alg:mgp}
			\begin{algorithmic}
				\STATE {\bfseries Input:} data: $\{\MX, \vy\}$, update rule: $\Lambda(\cdot)$, model matrices: $\{\MA_n, \MQ_n, \bar{\MH}\}_{n=1}^N$, initial state: $\{\bar{\vm}_0, \bar{\MP}_0\}$, initial approx. likelihood: $\{\liknatone, \liknattwo\}$, learning rate: $\rho$
				\WHILE{energy not converged}
				\STATE Convert approximate likelihood to mean/cov:
				\STATE $\likcov = -\frac{1}{2}(\liknattwo)^{-1}$
				\STATE $\likmean=\likcov\liknatone$
				\STATE Compute the approximate posterior:
				\FOR{$n=1:N$}
				\STATE $\bar{\vm}_n = \MA_n \bar{\vm}_{n-1}, \quad \bar{\MP}_n = \MA_n \bar{\MP}_{n-1} \MA_n^\T + \MQ_n$
				\STATE $\MV_n = \bar{\MH} \bar{\MP}_n \bar{\MH}^\T + \likcov_{n,n}, \quad \MW_n = \bar{\MP}_n \bar{\MH}^\T \MV_n^{-1}$
				\STATE $\bar{\vm}_n = \bar{\vm}_n + \MW_n (\likmean_n - \bar{\MH} \bar{\vm}_n)$
				\STATE $\bar{\MP}_n = \bar{\MP}_n - \MW_n \MV_n \MW_n^\T$
				\ENDFOR
				\FOR{$n=N-1:1$}
				\STATE $\MG_n = \bar{\MP}_n \MA_{n+1} \bar{\MP}_n^{-1}$
				\STATE $\MR_{n+1} = \MA_{n+1} \bar{\MP}_n \MA_{n+1}^\T + \MQ_{n+1}$
				\STATE $\bar{\vm}_n = \bar{\vm}_n + \MG_n(\bar{\vm}_{n+1} - \MA_{n+1} \bar{\vm}_n)$
				\STATE $\bar{\MP}_n = \bar{\MP}_n + \MG_n (\bar{\MP}_{n+1} - \MR_{n+1})\MG_n^\T$
				\STATE $\postmean_n=\bar{\MH}\bar{\vm}_n$
				\STATE $\postcov_{n,n}=\bar{\MH} \bar{\MP}_n \bar{\MH}^\T$
				\ENDFOR
				\STATE Update the approximate likelihood:
				\STATE $\jacobian_n, \hessian_{n,n} =\Lambda(\postmean_n, \postcov_{n,n}, \vy_n) \quad \forall n$
				\STATE $\liknattwo = (1-\rho) \, \liknattwo + \rho \, \frac{1}{2} \hessian$
				\STATE $\liknatone = (1-\rho) \, \liknatone + \rho \, \left( \jacobian -\hessian \, \postmean \right)$
				\ENDWHILE
			\end{algorithmic}
		\end{algorithm}
	\end{minipage}
\end{center}

\clearpage

\section{Reproducibility}\label{sec:reproducability}

Scripts to reproduce the results in this paper can be found in the experiments folder in the code repository: \url{https://github.com/AaltoML/BayesNewton/tree/main/experiments}. The individual experiments can be found in the \texttt{motorcycle}, \texttt{product} and \texttt{gprn} folders respectively. Each folder contains a main Python script, plus bash scripts to produce the results for each inference method class: \texttt{bn-newton.sh}, \texttt{bn-vi.sh}, \texttt{bn-ep.sh}, \texttt{bn-pl.sh}. After these have finished running, the \texttt{results\_bn.py} script can then be run to produce the plots.

\end{document}